%% file: conference.tex
\newif\ifarxiv
\arxivtrue

\ifarxiv
\documentclass[11pt]{article}
\else
\documentclass{article}
\fi

\usepackage{microtype}
\usepackage{graphicx}
\usepackage{subcaption}
\usepackage{booktabs} %

\usepackage{hyperref}

\ifarxiv
\else
\usepackage{icml2021}
\fi

\ifarxiv
\RequirePackage{fullpage}
\usepackage[utf8]{inputenc} %
\usepackage[T1]{fontenc}    %
\usepackage{hyperref}       %
\usepackage{url}            %
\usepackage{booktabs}       %
\usepackage{amsfonts}       %
\usepackage{nicefrac}       %
\usepackage{microtype}      %

\usepackage{xcolor,fixltx2e}
\usepackage[numbers]{natbib}
\usepackage{lmodern} %

\usepackage{makecell}
\usepackage{enumitem}
\usepackage{wrapfig}
\usepackage{bm}
\usepackage{lipsum}
\usepackage{authblk}
\usepackage{natbib}
\else
\fi
\definecolor{forestgreen}{rgb}{0.13, 0.55, 0.13}
\definecolor{green}{rgb}{0.4, 0.69, 0.2}
\definecolor{kellygreen}{rgb}{0.3, 0.73, 0.09}

\input{math_commands.tex}

\usepackage{cleveref}
\usepackage{enumitem}
\usepackage{algorithm}
\usepackage{wrapfig}
\usepackage{algpseudocode}
\usepackage{bold-extra}
\usepackage{pifont}%

\usepackage{graphicx}
\graphicspath{ {./figs/} }
\usepackage{caption}

\usepackage{booktabs}
\usepackage{multirow}
\usepackage{footmisc}
\usepackage{xurl}

\newcommand\rebuttal[1]{#1}
\newcommand\icmldraft[1]{#1}
\newcommand{\cmark}{\ding{51}}%
\newcommand{\xmark}{\ding{55}}%

\ifarxiv
\else
\def\setstretch#1{\renewcommand{\baselinestretch}{#1}}
\usepackage[compact]{titlesec}
\titlespacing{\subsection}{0pt}{*1}{*0}

\icmltitlerunning{Revisiting BFloat16 Training}
\fi

\begin{document}

\ifarxiv
\title{Revisiting BFloat16 Training}
\else
\twocolumn[
\icmltitle{Revisiting BFloat16 Training}

\icmlsetsymbol{equal}{*}

\icmlkeywords{Machine Learning, ICML}

\vskip 0.3in
]
\fi

\ifarxiv
\newcommand\CoAuthorMark{\footnotemark[\arabic{footnote}]} %
\author[1]{Pedram Zamirai\thanks{Work done at SambaNova Systems.}\footnote{Equal contributions.}}
\author[2]{Jian Zhang\protect\CoAuthorMark}
\author[2]{Christopher R. Aberger}
\author[3]{Christopher De Sa}
\affil[1]{Department of Computer Science and Engineering, University of Michigan}
\affil[2]{SambaNova Systems}
\affil[3]{Department of Computer Science, Cornell University}
\affil[ ]{\texttt{pedramz@umich.edu}, \texttt{jian.zhang@sambanovasystems.com},}
\affil[ ]{\texttt{christopher.aberger@sambanovasystems.com}, \texttt{cdesa@cs.cornell.edu}}
\maketitle
\fi

\begin{abstract}
State-of-the-art generic low-precision training algorithms use a mix of 16-bit and 32-bit precision, creating the folklore that 16-bit hardware compute units alone are not enough to maximize model accuracy.
As a result, deep learning accelerators are forced to support both 16-bit and 32-bit floating-point units (FPUs), which is more costly than only using 16-bit FPUs for hardware design.
We ask: \emph{can we train deep learning models only with 16-bit floating-point units, while still matching the model accuracy attained by 32-bit training}? 
Towards this end, we study \emph{16-bit-FPU training} on the widely adopted \BFHS unit. 
While these units conventionally use nearest rounding to cast output to 16-bit precision, we show that nearest rounding for model weight updates often cancels small updates, which degrades the convergence and model accuracy.  
Motivated by this, we study two simple techniques well-established in numerical analysis, stochastic rounding and Kahan summation, to remedy the model accuracy degradation in 16-bit-FPU training.
We demonstrate that these two techniques can enable up to $7\%$ absolute validation accuracy gain in 16-bit-FPU training. This leads to $0.1\%$ lower to $0.2\%$ higher validation accuracy compared to 32-bit training across seven deep learning applications.
\end{abstract}

\input{intro}

\input{prelim}

\input{accurate}

\input{experiment}

\input{related}

\input{conclusion}

\ifarxiv
\bibliographystyle{unsrtnat}
\bibliography{conference}
\else
\bibliography{conference}
\bibliographystyle{icml2021}
\fi

\clearpage
\appendix
\input{app_theory}

\input{algo}
\input{app_experiment}

\end{document}

%% file: math_commands.tex
\usepackage{amsmath,amsfonts,bm,amsthm}

\def\eqref#1{equation~\ref{#1}}

\def\1{\bm{1}}

\newcommand{\BFH}{BFloat16}

\newcommand{\BFHS}{BFloat16\ }
\newcommand{\FPHS}{Float16\ }

\def\mQ{{\mathcal{Q}}}

\newtheorem{theorem}{Theorem}

\def\vc{{\bm{c}}}

\def\vg{{\bm{g}}}

\def\vm{{\bm{m}}}

\def\vs{{\bm{s}}}

\def\vu{{\bm{u}}}
\def\vv{{\bm{v}}}
\def\vw{{\bm{w}}}
\def\vx{{\bm{x}}}
\def\vy{{\bm{y}}}

\def\mI{{\bm{I}}}

\def\mQ{{\bm{Q}}}

\def\mX{{\bm{X}}}

\DeclareMathAlphabet{\mathsfit}{\encodingdefault}{\sfdefault}{m}{sl}
\SetMathAlphabet{\mathsfit}{bold}{\encodingdefault}{\sfdefault}{bx}{n}

\def\sS{{\mathbb{S}}}

\newcommand{\R}{\mathbb{R}}

\DeclareMathOperator*{\argmin}{arg\,min}

%% file: intro.tex
\section{Introduction}
\label{sec:intro}

Recently there has been an explosion in the compute resources required for 
training deep learning models~\citep{shoeybi2019megatron,rajbhandari2019zero,real2019regularized}. 
As a result, there has been broad interest in leveraging low-precision ($<$ 32-bit) training
algorithms to reduce the required compute resources~\citep{de2017understanding,hubara2017quantized,gupta2015deep}. Among these algorithms, 
mixed-precision training---in which model activations and gradients are stored using a 16-bit floating point format while model weights and optimizer states use 32-bit precision---is commonly used when training generic deep learning models~\citep{micikevicius2017mixed,kalamkar2019study}. 
While there is a wide body of literature showing that low-precision training can minimally impact accuracy on specific models~\citep{wang2018training,zhang2017zipml},
conventional wisdom suggests that at least some traditional 32-bit computation is required as a fail-safe in generic deep learning training.
As such, new accelerator architectures for deep learning are forced to support both 32-bit and 16-bit floating point units (FPUs). This is much more costly in terms of area, power, and speed when compared to hardware with only 16-bit FPUs~\citep{horowitz20141,galal2013fpu}.

In this paper we question if 32-bit FPUs are truly needed for new deep learning hardware accelerators. Namely, can we match the model accuracy of 32-bit-precision
algorithms while leveraging \emph{only} 16-bit FPUs?
	To answer this question, we study \emph{16-bit-FPU training algorithms, ones which requires only 16-bit FPUs and which store activations, gradients, model weights, and optimizer states all in a 16-bit precision (unlike mixed-precision training algorithms which require 32-bit FPUs).}
Specifically, we focus on training with the \BFHS FPUs which are widely adopted in emerging deep learning accelerators~\citep{jouppi2017datacenter,burgess2019bfloat16}. 
\rebuttal{These FPUs take 16-bit inputs, internally uses 16-bit multiply and 32-bit accumulation via a fused multiply-accumulation unit, and then round the results to a 16-bit output. 
By replacing expensive 32-bit multipliers with 16-bit counterparts, \BFHS compute units can provide $3\times$ higher power efficiency, $1.5\times$ lower latency, and $1.5\times$ less chip area than 32-bit units~\citep{horowitz20141,galal2013fpu}.}
In addition, 16-bit-FPU training can reduce the memory footprint and bandwidth consumption of model weights and optimizers by $2\times$ compared to mixed precision or 32-bit precision training, especially for large models with billions of weights~\citep{shoeybi2019megatron,rajbhandari2019zero}.
Developing reliable 16-bit-FPU training algorithms will enable hardware designers to realize these advantages.

The simplest approach to 16-bit-FPU training is to take a 32-bit baseline and ``make it low-precision'' by replacing all the 32-bit numbers with 16-bit numbers and replacing each 32-bit floating-point operation with its analog in 16-bit-FPU, using nearest rounding\footnote{This nearest rounding is the standard rounding mode for compute unit output commonly supported across hardware platforms~\citep{intelround,nvround}.} to quantize as necessary: we call this approach the \emph{standard} algorithm.
\emph{Unfortunately, we show empirically that standard 16-bit-FPU training does not match 32-bit training on model accuracy across deep learning models.} 
For example, the standard 16-bit-FPU training algorithm one would run on conventional hardware attains $16\%$ and $7\%$ lower training and validation accuracies than a 32-bit baseline; this motivates us to study what factors limit the accuracy in the standard 16-bit-FPU algorithm.

The goal of our study is to first understand the model accuracy bottleneck in the standard 16-bit-FPU training, and then use the insights to suggest a clean, minimal set of simple techniques that allow 16-bit-FPU training to attain strong model accuracy across state-of-the-art models. Achieving these goals could inform hardware designers of necessary software and hardware supports to ensure strong accuracy for future training accelerators requiring only 16-bit FPUs. 

\rebuttal{To understand the accuracy degradation in the standard 16-bit-FPU algorithm, we derive insights from models with Lipschitz continuous gradients.
We reveal that nearest rounding of floating-point compute unit outputs can significantly degrade convergence and consequent model accuracy. 
Specifically, when running stochastic gradient descent, nearest rounding while updating model weights ignores small updates.
We show that such a phenomenon imposes a lower bound on the convergence limit of stochastic gradient descent on models with Lipschitz continuous gradients. 
This captures the insight that nearest rounding for updating model weights can lead to \emph{inevitable} yet significant convergence degradation for a plethora of models no matter how to tune learning rates.
This lower bound is complementary to existing \emph{worst-case} upper bounds in lower precision training~\citep{li2017training,hou2018analysis}.  
In comparison, nearest rounding in the forward and backward pass of backpropagation has a negligible impact on convergence in models such as least-squares regression.
These insights align with what we observe when training deep learning models.}

\rebuttal{Guided by our insights on the degradation, we apply
two simple techniques well-established in the numerical analysis literatures to achieve high-accuracy 16-bit-FPU training.
First, we can use \emph{stochastic rounding} instead of nearest rounding for the model weight updates. 
Here, the rounded weights become an unbiased estimate of the precise weights without rounding; thus, regardless of the magnitude of updates, the expectation of rounded weights converges at the same speed as the precise weights.
Second, we can use the Kahan summation algorithm~\citep{kahan1965further} to accumulate model updates while still keeping nearest rounding for all operations. This method tracks and compensates weight rounding errors across iterations with auxiliary 16-bit values, which avoids cancellation of many small weight updates. }

\rebuttal{Empirically, we first validate that, as suggested by our theory, nearest rounding for model weight updates is the sole convergence and model accuracy bottleneck on several deep learning models. We then demonstrate that 16-bit-FPU training using stochastic rounding or Kahan summation on model weight updates can match 32-bit training in model accuracy across applications~\citep{he2016deep,amodei2016deep,devlin2018bert,DLRM19}. 
To validate that nearest rounding for model weight updates is the cause of accuracy degradation, we show that if we store model weights in 32-bit precision without rounding during weight updates, and we keep using 16-bits and nearest rounding for all other operations, the attained model accuracy matches 32-bit training. 
Next, we demonstrate that 16-bit-FPU training with stochastic rounding for weight updates attains model accuracy matching 32-bit training for five out of seven applications in our study.
Note that while it works most of the time, this is not a silver bullet, as using stochastic rounding alone could not fully match 32-bit training on all models.
To address this, we show that Kahan summation for model weight updates closes remaining gaps on all the models we consider; this Kahan summation comes with a trade off, as it requires $2\times$ weight memory, but achieves up to $0.2\%$ higher validation accuracy than stochastic rounding. }

\emph{Our study suggests that deep learning accelerators using only 16-bit compute units are feasible if stochastic rounding and Kahan summation are supported respectively by the hardware and the software stack.} More concretely, our contributions and the paper outline are as follows.

\ifarxiv
\begin{itemize}[leftmargin=*]
\else
\vspace{-0.5em}
\begin{itemize}[leftmargin=*, itemsep=1pt]
\fi
	\item \icmldraft{In~\Cref{sec:accurate:subsec:understand}, we show that nearest rounding when updating model weights inevitably degrades convergence and consequent model accuracy. This suggests that only supporting nearest rounding is not enough to ensure strong model accuracy on emerging deep learning training accelerators requiring only 16-bit FPUs.}
	\item \icmldraft{In~\Cref{sec:accurate:subsec:algo}, guided by our insights, we improve convergence by applying \emph{stochastic rounding} and \emph{Kahan summation}, two well-known techniques in the numerical analysis literatures, during model weight updates.} 
	\item \icmldraft{In~\Cref{sec:exp}, we validate that 16-bit-FPU training with stochastic rounding or Kahan summation for model weight updates matches 32-bit training in validation accuracy on state-of-the-art models across seven applications.}
\end{itemize}

%% file: prelim.tex
\section{Preliminary}
\label{sec:prelim}

In this section we establish the background and notation for our study and present the preliminary observations that motivate our work.
We focus on the case of stochastic gradient descent (SGD), which is the primary workhorse used to train deep learning models. SGD computes gradients from a subset of training samples, and uses them to update the model weights so as to decrease the loss in expectation. In the classic supervised learning setting, let $\left(\mX, \vy \right)$ be a dataset where $\mX = \left[\vx_1, \vx_2, ..., \vx_n \right] \in \R^{n\times d}$ and $\vy = \left(y_1, y_2, ..., y_n \right) \in \R^{n}$. 
On this dataset, we use stochastic gradient descent to optimize a loss function $f(\vw) = 1/n\sum_{i = 1}^n f_i(\vw, \vx_i, y_i)$ defined by the model. 
At the $t\text{-th}$ iteration, we sample an index subset $\sigma(t)\subset \{1,2,.., n \}$  and compute a sample gradient $\nabla f_{\sigma (t)} (\vw_t)$ as an unbiased estimate of the full gradient $\nabla f(\vw)$. 
In deep learning, model training can be described as a compute graph where the {compute graph operators} such as addition and matrix multiplication are the nodes. For example, the model weight update operator is defined as the subtraction in the operation $\vw_{t + 1} = \vw_{t} - \alpha \nabla f_{\sigma (t)} (\vw_t)$ which updates the model weight $\vw$.

\begin{table}
\caption{\rebuttal{\textbf{Hardware costs such as chip area, energy and latency of the fused multiply-and-accumulation (FMAC) unit.} 16-bit FMAC reduce primary hardware cost with lower precision multiply; the 32-bit accumulation is inexpensive but critical to numerical precision, which is standard in machine learning accelerators.}}
\centering
\ifarxiv
\else
\small
\fi
\begin{tabular}{c  c  c c  c c c c}
\toprule
 \multirow{2}{*}{Compute unit} & \multicolumn{2}{c} {Multiply} & \multicolumn{2}{c} {Accumulation} \\ 
 \cmidrule(lr){2-3}
 \cmidrule(lr){4-5}
  & Precision & Cost & Precision & Cost \\
\midrule
32-bit FMAC & 32-bit & \color{red}{High} & 32-bit & \color{forestgreen}{Low}\\
16-bit FMAC & 16-bit & \color{forestgreen}{Low} & 32-bit & \color{forestgreen}{Low}\\
\bottomrule
\end{tabular}
\vspace{-0.5em}
\label{tab:comparison}
\end{table}
\paragraph{16-bit Floating-point Units} On modern hardware, operators in the compute graph are supported by {fused multiply-and-accumulation (FMAC) compute units}.
These units work by computing $a \leftarrow a + (x \times y)$, where $x$ and $y$ are input floating point numbers, and $a$ is an accumulator that is part of the FMAC unit. %
Importantly, for a 16-bit FMAC unit shown in~\Cref{tab:comparison}, the accumulator $a$ has higher-than-16-bit precision. 
\rebuttal{This higher precision accumulator is inexpensive compared to the multiplier in terms of chip area and energy consumption in FMAC units, but is critical to the numerical accuracy of operations such as matrix multiplication and convolution.
Thus, 16-bit FMAC units with higher precision accumulator are standard for modern hardware including TPUs and GPUs~\citep{chao2019cloud,markidis2018nvidia,stephens2019bfloat16}, and will likely continue to be the standard.} 
Because of the higher precision accumulator, the result in the accumulator then needs to be \emph{rounded} to 16-bits before it is output from the FMAC unit (e.g. before writing to memory). 
FMAC units use the same hardware implementation to support all operators from simple additions to computationally intensive convolutions, so this output rounding happens for all the operators in a compute graph.

\paragraph{Nearest Rounding} FMAC output rounding is widely implemented with \emph{nearest rounding} as the standard mode, due to its efficient support across hardware platforms. 
``Nearest rounding'' means rounding a higher precision floating point number to the closest low-precision representable value. 
\emph{Given that the add step already uses accurate higher precision accumulators, this nearest rounding is the primary source of numerical errors for training using 16-bit FMAC.}

\begin{table}
\caption{\rebuttal{\textbf{Training with different precisions.} Mixed-precision training uses 16-bit activations and gradients while requiring  32-bit optimizer states and a 32-bit master copy of model weights. Thus it requires both 16-bit and 32-bit FPUs. 16-bit-FPU training uses fully 16-bit weights, optimizer states, activation and gradients. Thus it requires only 16-bit FPU on accelerators.}}
\centering
\ifarxiv
\begin{tabular}{c c c c}
\else
\small
\begin{tabular}{@{\hskip 0em}c c c c@{\hskip 0em}}
\fi
\toprule
Algorithm & 32-bit & Mixed-precision & 16-bit-FPU\\
\midrule
Weight & \color{red}{32-bit} & \color{red}{32-bit (master copy)} & \color{forestgreen}{16-bit}\\
Optimizer state & \color{red}{32-bit} & \color{red}{32-bit} & \color{forestgreen}{16-bit}\\
 Activation \& grad. & \color{red}{32-bit} & \color{forestgreen}{16-bit} & \color{forestgreen}{16-bit} \\
No 32-bit FPU & \color{red}{\xmark} & \color{red}{\xmark} & \color{forestgreen}{\cmark}\\
\bottomrule
\end{tabular}
\vspace{-0.5em}
\label{tab:algo_compare}
\end{table}
\paragraph{Training with Different Precisions} In this context of 16-bit FMAC units and nearest rounding, we discuss the following training-precision approaches.
In \textbf{32-bit precision training}, all the compute graph operators read and write memory using a 32-bit precision. These operators require 32-bit FMAC units, which constrains the compute and memory efficiency. 
\rebuttal{In \textbf{mixed precision training}, optimizer states and a master copy of model weights are stored in 32-bit precision for model accuracy considerations while activations and gradients use 16-bit precision~\citep{micikevicius2017mixed,kalamkar2019study}. 
Thus, new accelerators customized to maximize efficiency for mixed precision training still require 32-bit floating-point units for operators involving 32-bit weights and optimizer states as the input; this has lower efficiency in power, speed and chip area than only using 16-bit FPU. }
\rebuttal{In \textbf{16-bit-FPU training}, activation, gradient, weight and optimizer state are all stored in 16-bit precision.
All operators use pure 16-bit input and write out 16-bit output after rounding. 
Thus, aside from just saving memory, 
16-bit-FPU training can eliminate the requirement for 32-bit FPU as shown in~\Cref{tab:algo_compare}.
In spite of the consequent favorable efficiency, we now show that it can be surprisingly challenging for standard 16-bit-FPU training to match 32-bit training on model accuracy.}

\ifarxiv
\begin{wrapfigure}{r}{0.425\textwidth}
	\begin{minipage}{0.425\textwidth}
		\vspace{-1.75em}
		\begin{figure}[H]
			\centering
			\includegraphics[width=\textwidth]{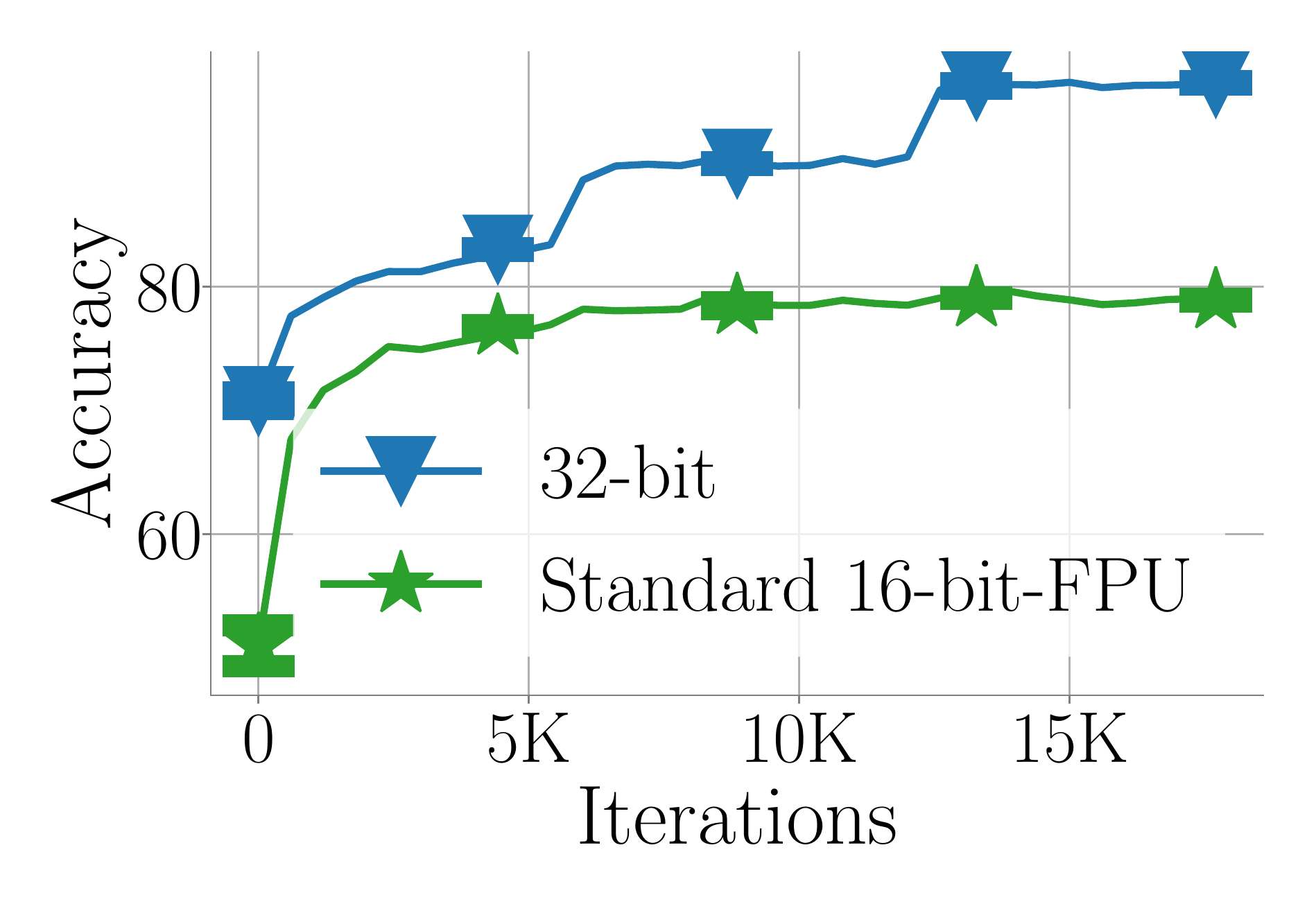}
					\vspace{-2em}
			\caption{Standard 16-bit-FPU training shows lower training accuracy compared to 32-bit training on a BERT model.\protect\footnotemark}
			\label{fig:teaser}
		\end{figure}
		\vspace{-1.5em}
	\end{minipage}
\end{wrapfigure}
\else
\begin{figure}
	\centering
	\includegraphics[width=0.675\linewidth]{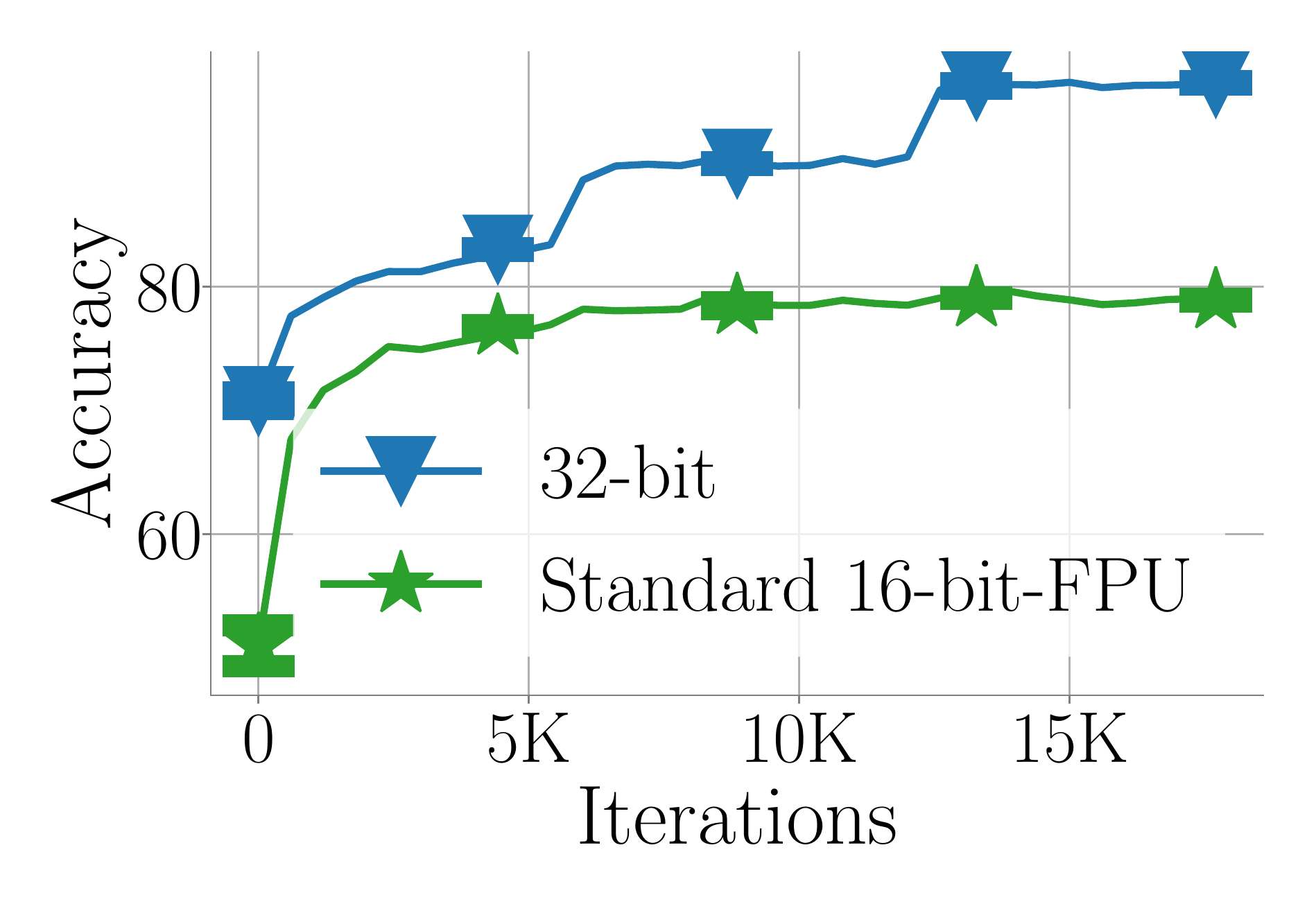}
			\vspace{-1em}
	\caption{Standard 16-bit-FPU training shows lower training accuracy compared to 32-bit training on a BERT model.\protect\footnotemark
}
	\vspace{-0.25em}
	\label{fig:teaser}
\end{figure}
\fi
\footnotetext{\rebuttal{We note that 32-bit and standard 16-bit-FPU training in~\Cref{fig:teaser} start from the same initialization with the same accuracy. The gap at the beginning of the two curves is due to smoothing for visual clarity. We refer to~\Cref{app:exp:bottleneck} for less smoothed curves.}}
\paragraph{Motivating Observations}
Although recent works have shown that certain models are robust to numerical error during training~\citep{wang2018training,de2015taming,zhang2017zipml}, surprisingly, we observe that it is challenging for 16-bit-FPU training to attain the same accuracy as 32-bit precision training on several state-of-the-art deep learning models. To demonstrate this, we compare 32-bit precision training and standard 16-bit-FPU training (using nearest rounding for all its operator outputs). 
For example, \Cref{fig:teaser} illustrates that for a BERT model for natural language inference, the standard 16-bit-unit training algorithm demonstrates $16\%$ and $7\%$ lower training and validation accuracies than 32-bit training. 
This gap suggests that nearest rounding is the primary source of numerical error in 16-bit-FPU algorithms, significantly degrading the convergence and model accuracy.  
To alleviate this problem, in~\Cref{sec:accurate}, we study how nearest rounding impacts convergence, and we expose insights which lead to simple techniques to improve the model accuracy in 16-bit-FPU training.

%% file: accurate.tex
\section{Precise Model Updates Are All You Need}
\label{sec:accurate}
To understand how to improve the model accuracy attained by 16-bit-FPU training, in this section we analyze the impact of nearest rounding, the primary source of numerical error in the standard 16-bit-FPU training algorithm. 
\rebuttal{In~\Cref{sec:accurate:subsec:understand}, we show that when model updates are small relative to the model weights, nearest rounding for model weight updates ignores small updates and \emph{inevitably} impedes the convergence limits of stochastic gradient descent no matter how the learning rate is tuned.
In contrast, we show that nearest rounding in the forward and backward compute can have a much weaker impact on the convergence throughout training. 
These insights emphasize the importance of more precise model weight updates for 16-bit-FPU training. To achieve such precise updates, in~\Cref{sec:accurate:subsec:algo} we apply two simple numerical techniques well-establish in the numerical analysis literatures, stochastic rounding and Kahan summation, for model weight updates. 
Although we use models with Lipschitz continuous gradients for the simplicity in exposing insights, we will show in~\Cref{sec:exp} that our insights transfer empirically to representative deep learning models. }

\vspace{-0.5em}
\subsection{Understanding the Impact of Nearest Rounding}
\label{sec:accurate:subsec:understand}

\rebuttal{In this section, we use training with batch size $1$ as a proxy to expose the impact of numerical errors due to nearest rounding. 
First, we show the impact of nearest rounding for weight updates on models which define loss functions with Lipschitz continuous gradients. %
Then, using least-squares regression as an example, we show that nearest rounding for forward and backward compute has a comparatively much weaker influence on convergence. %
More concretely, we focus on a loss function $f(\vw) = 1/n\sum_{i = 1}^n f_i(\vw, \vx_i, y_i)$ which can perfectly fit the dataset; this is intended to reflect the overparameterized setting in modern deep learning models~\citep{li2018algorithmic,jacot2018neural}.
In this setting, the model can fit every data sample at the same minimizer $\vw^{*}$ which  implies that the sample gradient $\nabla f_{\sigma (t)}(\vw^{*})=\bm{0}$ always holds true at this minimizer (\textbf{A1}). 
To ensure convergence with a bounded gradient variance, we assume $L$-Lipschitz continuity on the gradient of sample loss $f_i(\vw, \vx_i, y_i)$ which implies $\|\nabla f_{\sigma (t)}(\vw) - \nabla f_{\sigma (t)}(\vv) \| \leq L \|\vw - \vv\|, \forall \vv, \vw$ (\textbf{A2}).}
We let $\epsilon$ denote the machine epsilon of our floating point format, such that if $u$ and $v$ are adjacent representable numbers in our floating point format, $\epsilon |u| \le |u - v| \le 2 \epsilon |u|$. Under this standard assumption for numerical analysis on floating point numbers~\citep{stoer2013introduction}, nearest rounding $\mQ\left(\cdot\right)$ will have a bounded error of $| \mQ(u) - u | \le \epsilon |u|$ for any $u$ in range. To simplify the presentation, we ignore overflow and underflow in our analysis here, and disregard factors of $\epsilon^2$ (as is standard in analyses of floating point error).

\paragraph{Nearest Rounding for Model Weight Updates}

When stochastic gradient descent updates model weights with nearest rounding, the model weights evolve as $\vw_{t + 1} = \mQ\left(\vw_{t} - \alpha \nabla f_{\sigma (t)} (\vw_t) \right)$. For a weight dimension $j$, if the model update $\left[\alpha \nabla f_{\sigma (t)} (\vw_t)\right]_j$ is smaller than half of the difference between $[\vw_t]_j$ and its neighboring representable value in a certain precision format, nearest rounding cancels this model update. This often emerges in the late training stage when the magnitude of gradient becomes small or learning rate decays small. We show that nearest rounding cancels updates across all dimensions for models with Lipschitz continuous gradients when approaching the optimal weights in~\Cref{prop:high_prob}; we defer the proof to~\Cref{app:theory}.
\begin{theorem} 
Consider running one step of SGD on a loss function under assumptions \textbf{A1} and \textbf{A2}.
The model weight update will be entirely canceled by nearest rounding if 
\begin{equation}
    \| \vw - \vw^* \|
    \le
    \frac{\epsilon}{ \alpha L + \epsilon} \cdot \min_j \left| 
    w^*_j \right|,
    \label{equ:convergence_main0}
\end{equation}
where $w_j^*$ denotes the $j$-th dimension of the optimal solution $\vw^{\star} =\argmin_{\vw\in \R^d} f(\vw)$.
Additionally, if we run multiple steps of SGD using nearest-rounded weight updates and fixed learning rate $\alpha \le 1/L$, then the distance of the weights $\vw_t$ at any timestep $t$ from the optimum is bounded by
\[
   \| \vw_t - \vw^* \|
   \ge
   \min\left(
       \frac{\epsilon\left(1 - \alpha L \right)}{ \alpha L + \epsilon} \cdot \min_j \left| w^*_j \right|
   ,
       \| \vw_0 - \vw^* \|
   \right).
\]
\label{prop:high_prob}%
\end{theorem}\vspace{-1em}%
\Cref{prop:high_prob} reveals that for loss functions with Lipschitz continuous gradient, nearest rounding cancels the entire model updates when the distance towards the optimal solution $\vw^*$ is small relative to the magnitude of $\vw^*$. 
Thus in the late stage of training, the model weights can halt in a region with radius $\frac{\epsilon}{ \alpha L + \epsilon}\min_j \left| w^*_j \right|$ around $\vw^*$. %
Our lower bound shows that this region limits the convergence of SGD with nearest rounding on model weight updates. Because the lower bound on $\min_j \left| w^*_j \right|$ is also in the order of $\mathcal{O}\left(\epsilon\right)$, this convergence limit is worse for lower precision formats with a large $\epsilon$ value. In this bound, one key property is the dependency on the step size: as the step size becomes small, this error lower bound becomes \emph{worse}, which is the opposite of the usual effect of diminishing the step size in SGD. 
\rebuttal{More importantly, \Cref{prop:high_prob} shows that the convergence limit depends on the magnitude of the optimal weight $\vw^*$. Given that $\vw^*$ can be arbitrarily far from the zero vector, our lower bound reveals that this substantial convergence degradation can be \emph{inevitable} for stochastic gradient descent using low precision floating point numbers no matter how the learning rates are tuned.
This lower bound is complementary to existing upper bounds in worst-case convergence analysis. We use this lower bound together with the previous upper bounds to inform future accelerator designers that only supporting nearest rounding is not enough to maximize model accuracy.
} 
 In~\Cref{sec:exp}, we will empirically show that these insights from models with Lipschitz continuous gradients can also generalize to deep learning models, which explains the convergence and model accuracy degradation due to the small updates cancellation in model weight updates.

\paragraph{Nearest Rounding for Forward and Backward Compute}
In contrast to the significant convergence degradation imposed by nearest rounding on model weight updates, we show that the nearest rounding in the gradient computation (in the forward and backward passes of backpropagation) can impact convergence minimally.
To show this, we use a least-squares regression model $\frac{1}{2n}\sum_{i = 1}^n \|\vx_i^T \vw - y_i \|^2$ under the assumption \textbf{A1} and \textbf{A2} as an example. In this example, we consider SGD with nearest rounding only for compute operations which generate activations and gradients. 
Here to compute the gradient for least-squares regression models, the linear layer passes the rounded activation $a_i = \mQ \left(  \vx_i^T \vw - y_i \right)$ to the loss layer. (We see no quantization error within the dot product $\vx_i^T \vw$ itself, as all accumulation here is done with the higher-precision accumulator of the FMAC.) In the backward stage, the loss layer feeds the rounded activation gradients $g_{a, i} = \mQ\left(a_i \right)$ back to the linear layer. The weight gradient is then computed as $\nabla_{\mQ} f_i(\vw) := \mQ\left( g_{a, i} \vx_i \right) = \mQ \left( \mQ\left( \mQ\left(\vx_i^T \vw_t - y_i  \right) \right) \vx_i \right)$.  
\rebuttal{To isolate the impact of nearest rounding for activations and gradients, we do not round model weights and use exact arithmetic for weight updates in this example.
Formally in~\Cref{theo:covergence_other} in~\Cref{app:subsec:A2}, we show that stochastic gradient descent 
with activation and gradient rounding allows for an upper bound on $\|\vw_t - \vw^{\star}\|$ which can be arbitrarily close to $0$ with a large enough number of iterations. This upper bound can be substantially smaller than the lower bound in~\Cref{prop:high_prob}. 
It \emph{shows that rounding for the weight updates is the primary source of error}, as even with all other operations rounded, the algorithm is guaranteed to converge closer to the optimum than is even possible with just the weight updates rounded with nearest rounding. Note that the bound in~\Cref{theo:covergence_other} in~\Cref{app:subsec:A2} is able to guarantee arbitrarily accurate solutions because we ignore underflow. In practice, precision would eventually be limited by underflow even in the setting of~\Cref{theo:covergence_other}; however, the underflow threshold for \BFHS is small enough that this represents a level of error that deep learning applications can tolerate in general. We refer to~\Cref{app:subsec:A2} for detailed discussions.}

\ifarxiv
\begin{wrapfigure}[19]{r}{0.43\textwidth}
	\begin{minipage}{\linewidth}
	\vspace{-2em}
	\begin{figure}[H]
	\includegraphics[width=1\linewidth]{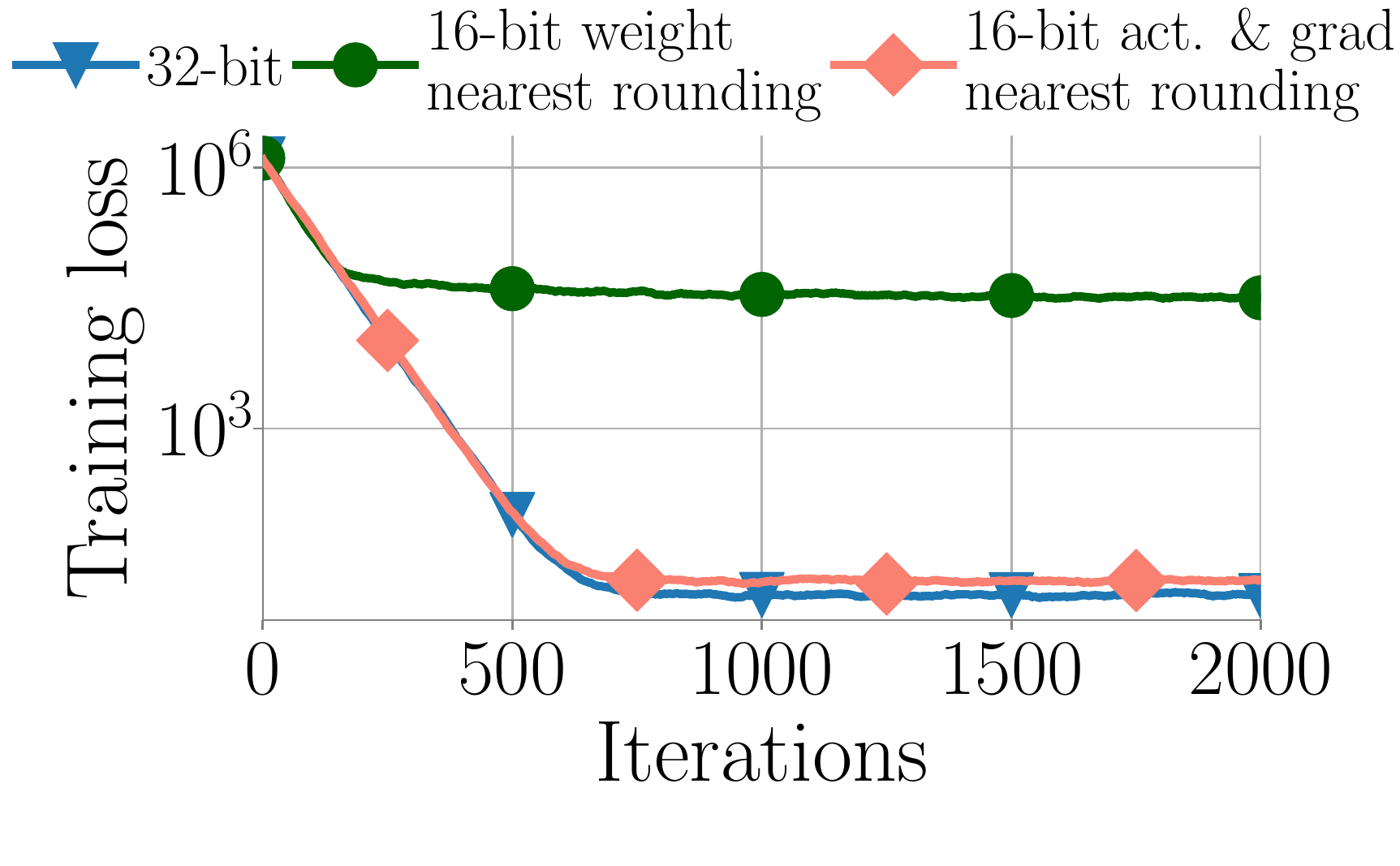}	
\else
\begin{figure}
\centering
\includegraphics[width=0.85\linewidth]{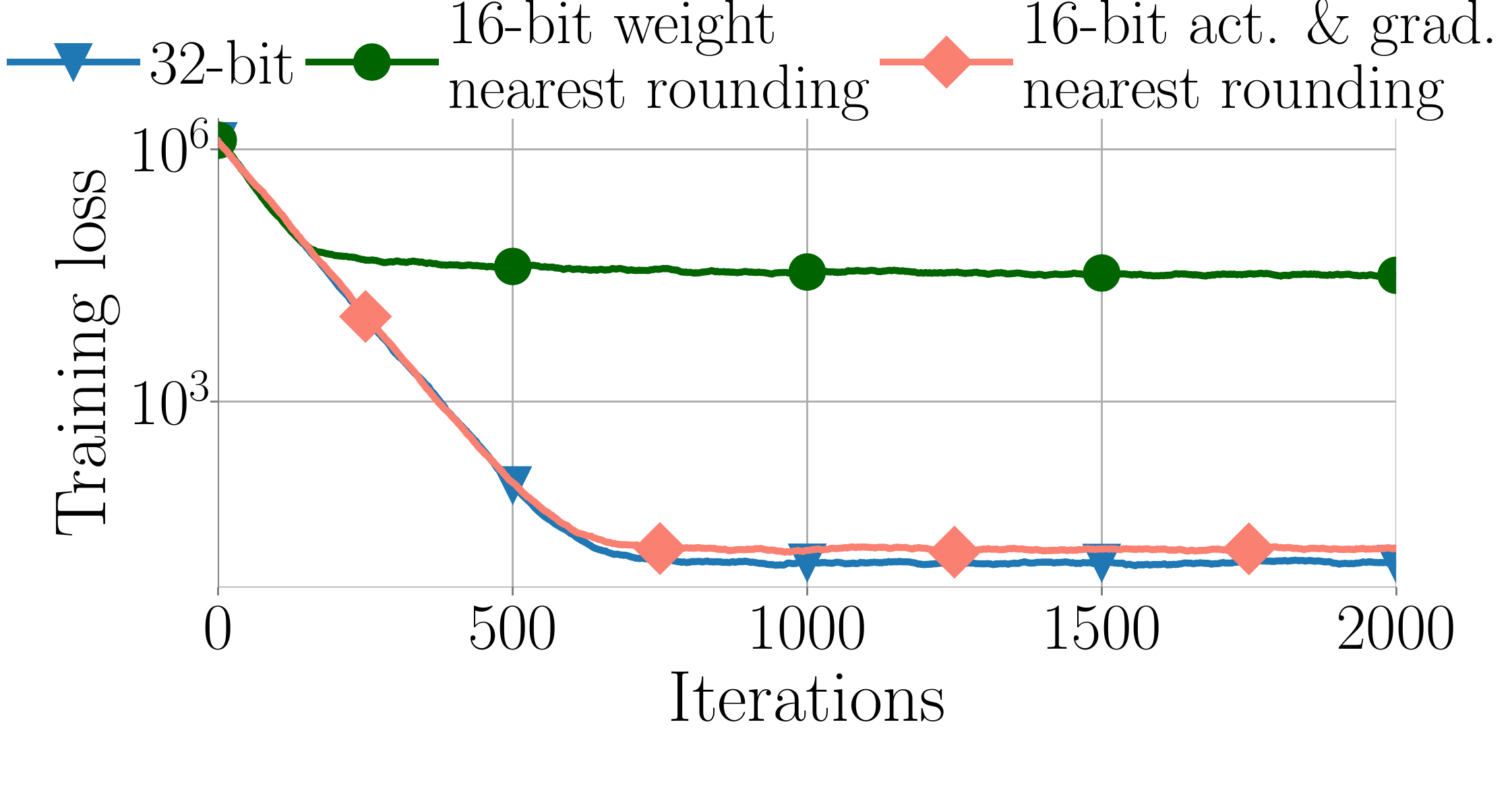}
\fi
\vspace{-1em}
\caption{\textbf{Theory validation}. On a least square regression model, (smoothed) training losses with 16-bit nearest rounding for weight updates saturate at a higher level than 32-bit training. With only using nearest rounding for forward and backward compute, the losses saturate much closer to 32-bit training.}
\label{fig:rnd_weight_inf}
\end{figure}
\ifarxiv
\end{minipage}
\end{wrapfigure}
\else
\fi

\paragraph{Theory Validation} To validate our insights on models with Lipschitz continuous gradient, we compare the impact of nearest rounding for model weight updates against that of nearest rounding in forward and backward compute on a synthetic 10-dimensional least-squares regression problem. 
Specifically, the input data are sampled from a zero-mean unit-variance normal distribution while the model weight is generated uniformly in the range of $[0, 100)$. We perturb the label with a zero-mean normal distribution with standard deviation $0.5$. 
As shown in~\Cref{fig:rnd_weight_inf}, when using a learning rate $0.01$ and 16-bit nearest rounding for model weight updates, the training loss saturates at a magnitudes higher level than stochastic gradient descent without rounding because of updates cancellation. 
Meanwhile, when using nearest rounding only for forward and backward compute, the loss saturates at a level close to that attained by training without rounding. These observations align with our insights on the relative impact of nearest rounding for model weight updates and for forward and backward compute.

\subsection{High-accuracy 16-bit-FPU Training}
\label{sec:accurate:subsec:algo}
In~\Cref{sec:accurate:subsec:understand}, we showed that nearest rounding for model weight updates is the bottleneck for convergence in the standard pure 16-bit-FPU training algorithm; this is because it cancels small model updates which degrades the model weight precision. 
This motivates us to consider two existing techniques, \emph{stochastic rounding and Kahan summation}~\citep{kahan1965further} for improving weight updates. These techniques have been reliably applied in different numerical domains~\citep{hopkins2020stochastic,antonana2017reducing} and can hypothetically enable high-accuracy 16-bit-FPU training. 
\rebuttal{We present details on how to integrate these techniques into SGD and AdamW optimizers with 16-bit model weights and optimizer states such as momentum in \Cref{app:algo}.} 
\ifarxiv
\vspace{-0.25em}
\else
\fi
\paragraph{Stochastic Rounding}
Stochastic rounding for floating point numbers has been used in training certain model components ~\citep{zhang2018training} and can potentially improve the model accuracy of 16-bit-FPU training for general models. Specifically, let $\sS$ be the set of all values representable by a limited precision format: the upper and lower neighboring values for $a \in \R$ are $a_u = \min_{x \geq a, x\in \sS}x$ and $a_l = \max_{x \leq a, x\in \sS} x$. Stochastic rounding randomly rounds $a$ up to $a_u$ with probability $(a - a_l)/(a_u - a_l)$ and otherwise rounds down to $a_l$. We consider 16-bit-FPU training using stochastic rounding only for the subtraction output in the model update $\vw_t - \alpha \nabla f_{\sigma (i)}(\vw_t)$. We keep nearest rounding for all the other compute operations. Here, the rounded model weight is an unbiased estimate of the precise value, so it will still make progress in expectation; this prevents the halting effect from nearest rounding on model weight updates. 
\rebuttal{We note that in modern hardware, stochastic rounding can be implemented without any expensive multiply or division arithmetic~\citep{de2017understanding}. Thus using stochastic rounding for model weight updates adds minimal overhead when training modern deep learning models; we discuss explicitly how to achieve this in~\Cref{app:algo:subsect:efficiency_stoc}.}
\ifarxiv
\else
\vspace{-0.5em}
\fi
\begin{algorithm}
    \caption{SGD updates with Kahan summation}
    \label{alg:kahan_sgd}
    \begin{algorithmic}[1]
         \State Auxiliary value $\vc_0 \leftarrow 0$ at initialization
         \State \textbf{Input:} Model updates $ - \alpha \nabla f_{\sigma(i)} (\vw_t)$ at iter. t
        \State $\vu_{t+1} \leftarrow - \alpha \nabla f_{\sigma(i)} (\vw_t)$
        \State $\vy_{t+1} \leftarrow \vu_{t+1} - \vc_t$ \hfill \Comment{Compensate updates}
        \State $\vs_{t+1} \leftarrow \vw_{t} + \vy_{t+1}$ \hfill \Comment{Accumulate updates}
        \State $\vc_{t+1} \leftarrow \left(\vs_{t+1} - \vw_t\right) - \vy_{t+1}$ \hfill \Comment{Measure errors}
        \State $ \vw_{t+1} \leftarrow \vs_{t+1}$ 
        \State \textbf{Return:} $\vw_{t+1}$
    \end{algorithmic}
\end{algorithm}
    \vspace{-1em}
\paragraph{Kahan Summation}
The Kahan summation algorithm uses an auxiliary variable to track numerical errors and to compensate the accumulation results. In the context of 16-bit-FPU training, we use a 16-bit auxiliary variable $\vc_t \in \R^d$ to track the error in model weights. To ensure that it still only requires 16-bit FPUs, we keep nearest rounding for all operators in compute graphs, including those during Kahan accumulation in~\Cref{alg:kahan_sgd}. At iteration $t$, we first compensate the current model update $\vu_{t + 1}$ by subtracting the previous error $\vc_t$. We then compute the new model weights by adding the compensated updates $\vy_{t+1}$ to the current weights $\vw_t$. We reversely subtract previous model weights $\vw_t$ and the compensated updates $\vy_{t+1}$ to acquire the new numerical error $\vc_{t + 1}$ in the updated weights $\vw_{t+1}$. 
For small updates $\vu_t$ which cause no change in the weights after nearest rounding, this reverse subtraction records the canceled updates in the error $\vc_{t+1}$. 
Across iterations, small updates can be accumulated in $\vc_{t+1}$ until $\vc_{t+1}$ grow large enough to affect the model weights; this allows convergence to continue when it would otherwise halt due to nearest-rounding effects. 
\rebuttal{In spite of the additional auxiliary value, 16-bit-FPU training with Kahan summation for model weight updates can still have advantages in terms of throughput and memory consumption compared to 32-bit and mixed precision training; we refer to~\Cref{app:algo:subsect:efficiency_kahan} for details.}

%% file: experiment.tex
\section{Experiments in Deep Learning}
\label{sec:exp}
\ifarxiv
\begin{table}
\else
\begin{table*}
\fi
\caption{\textbf{Model accuracy bottleneck for the standard 16-bit-FPU training algorithm.} This
algorithm shows validation accuracy gap compared to 32-bit training. In an ablation of this algorithm, we use 32-bit model weights and turn off nearest rounding only on model weight updates. This eliminates the gap, suggesting that nearest rounding on model weight updates is the accuracy bottleneck.
}
\ifarxiv
\else
\small
\fi
\centering
\begin{tabular}{c  c  c  c  c }
\toprule
\ifarxiv
\multirow{2}{*}{Model} & \multirow{2}{*}{Dateset (Metric)} & \multirow{2}{*}{32-bit} & \multirow{2}{*}{Standard 16-bit-FPU} & Standard 16-bit-FPU  \\
   & & & & \& 32-bit weights\\
\else
Model & Dateset (Metric) & 32-bit & Standard 16-bit-FPU & Standard 16-bit-FPU \& 32-bit weights \\
\fi
\midrule
 ResNet-18 &  CIFAR10 (Acc\%) & $95.45 \pm 0.07$ & $94.23 \pm 0.12$ & $95.40 \pm 0.05$ \\
  DLRM &  Kaggle (AUC\%) & $80.27 \pm 0.01$ & $78.49 \pm 0.08$ & $80.26 \pm 0.01$ \\ 
  BERT-Base &  MNLI (Acc\%) & $84.26 \pm 0.08$ & $77.53 \pm 0.07$ & $84.34 \pm 0.04$ \\
\bottomrule
\end{tabular}
\label{tab:ablation:test}
\ifarxiv
\end{table}
\else
\end{table*}
\fi

\ifarxiv
\begin{figure}[t]
\else
\begin{figure*}
\fi
\ifarxiv
\includegraphics[width=\textwidth]{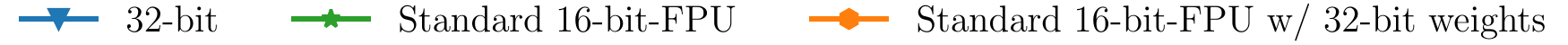}
\else
\includegraphics[width=0.95\textwidth]{accurate/legend}
\fi
\centering
\ifarxiv
\begin{tabular}{@{\hskip 0em}c@{\hskip 0em}c@{\hskip 0em}c@{\hskip 0em}}
    \includegraphics[width=0.335\textwidth]{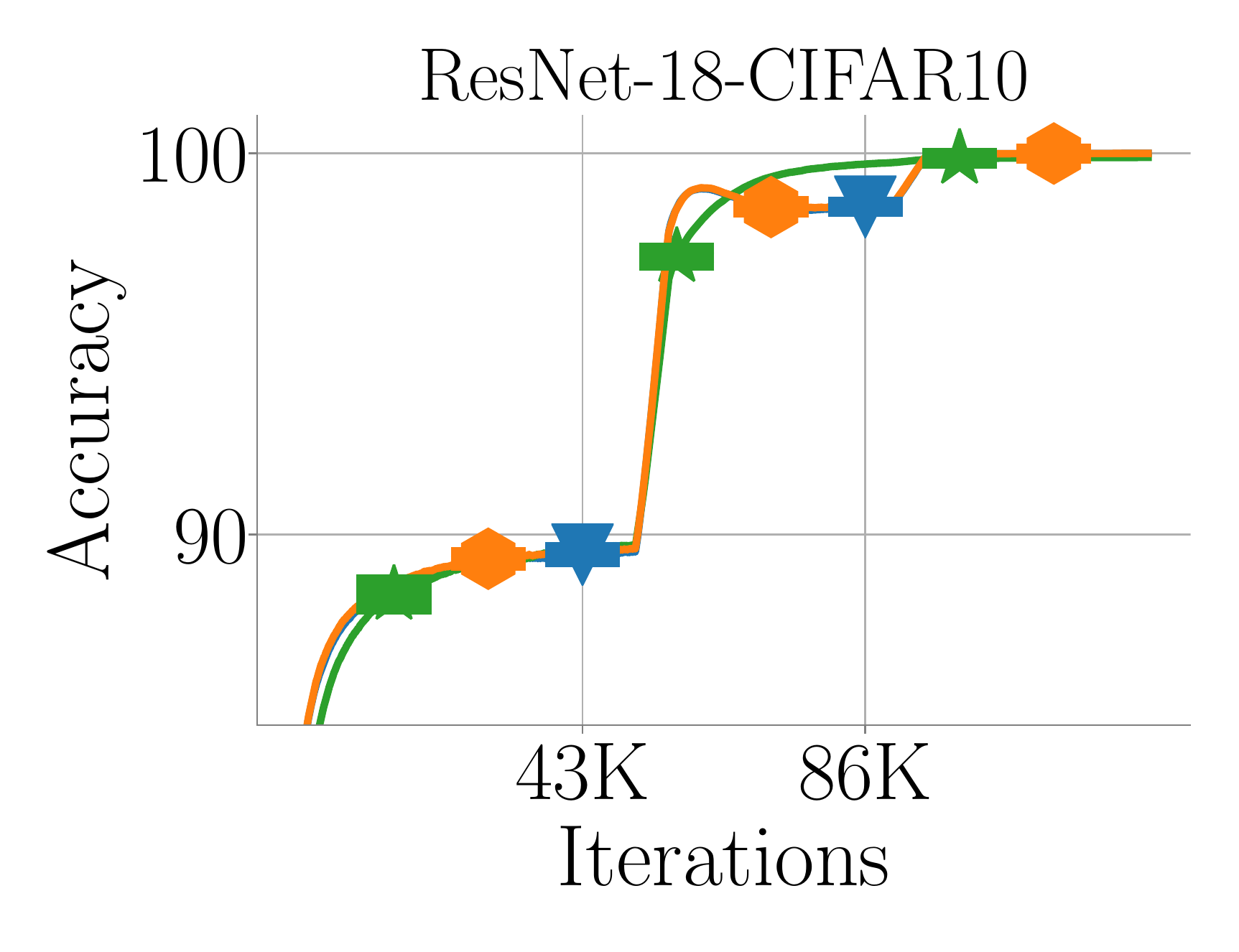} &
    \includegraphics[width=0.335\textwidth]{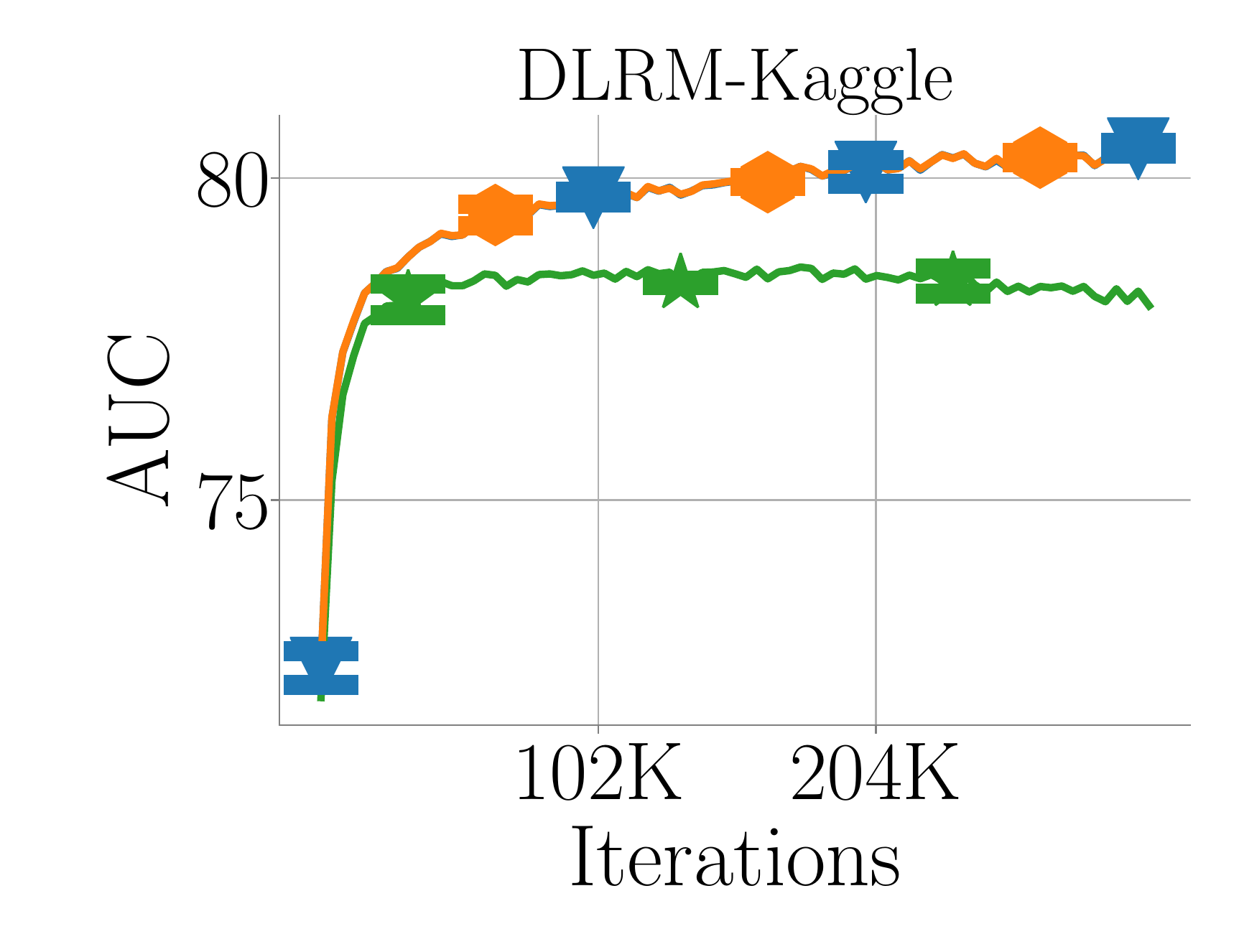} &
    \includegraphics[ width=0.335\textwidth]{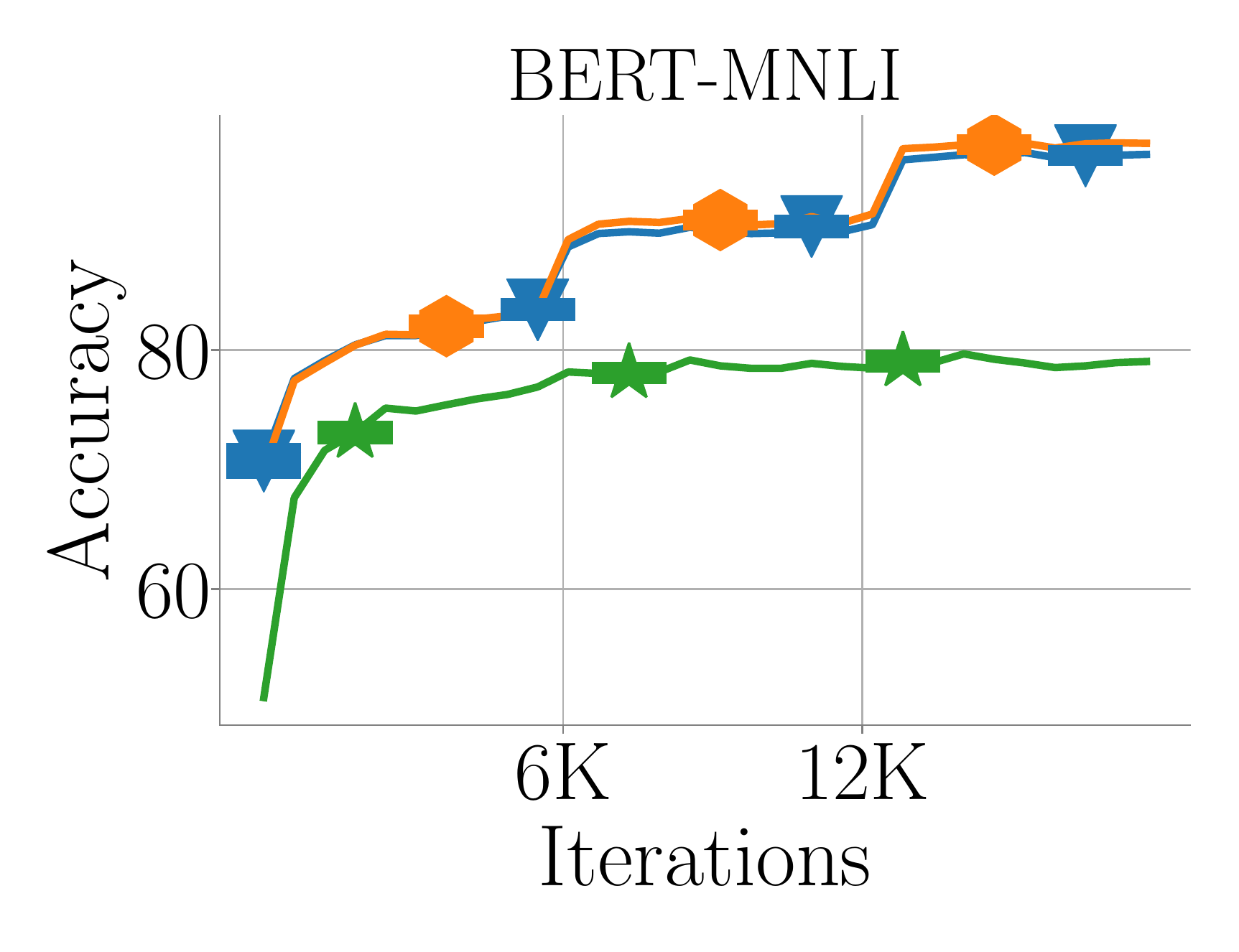}
\end{tabular}
\vspace{-0.5em}
\else
\begin{tabular}{c c c}
    \includegraphics[width=0.28\textwidth]{accurate/resnet-18-cifar10_train} &
    \includegraphics[width=0.28\textwidth]{accurate/dlrm-kaggle_train} &
    \includegraphics[ width=0.28\textwidth]{accurate/bert-mnli_train}
\end{tabular}
\fi
  \caption{\textbf{Training accuracy gap imposed by the standard 16-bit-FPU training algorithm.} The standard algorithm fails to match the training accuracy of 32-bit training, especially in the middle-to-late stage. We close this accuracy gap by ablating nearest rounding for weight updates from the standard algorithm. This indicates that nearest rounding for model weight update is the accuracy bottleneck.}
  \label{fig:acc:ablation}
 \ifarxiv
\end{figure}
\else
\end{figure*}
\fi

Our theory in~\Cref{sec:accurate} reveals that nearest rounding on model weight updates is the primary source of numerical error during training. This motivates us to suggest using stochastic rounding and Kahan summation in 16-bit-FPU training for improved model accuracy. 
To first validate our theory, in this section we start by demonstrating that by ablating nearest rounding on model weight updates from the standard 16-bit-FPU training algorithm, the model accuracy gap compared to 32-bit precision training can be closed on deep learning models.
Next, we show empirically that with stochastic rounding or Kahan summation on model weight updates, 16-bit-FPU training can match the accuracy of 32-bit training across representative deep learning applications. 

    \vspace{-0.5em}
\paragraph{Experiment Setup}
To validate the accuracy bottleneck, we consider three representative models: ResNet-18~\citep{he2016deep} on the CIFAR10 image classification~\citep{cifar10}, BERT-Base~\citep{devlin2018bert} on the MNLI natural language inference~\citep{wang2018glue}, and DLRM model~\citep{DLRM19} on the Kaggle Advertising Challenge~\citep{kaggle}. To extensively evaluate 16-bit-FPU training with stochastic rounding and Kahan summation, we additionally consider larger datasets and more applications: ResNet-50 on the ImageNet~\citep{deng2009imagenet}, BERT-Base on the Wiki103 language model\footnote{We subsample $25\%$ of the Wiki103 and $100$ hours of Librispeech training set because of the training time.}~\citep{merity2016pointer}, DLRM model on the Criteo Terabyte dataset~\citep{criteoterabyte}, and Deepspeech2~\citep{amodei2016deep} on the LibriSpeech datasets~\citep{panayotov2015librispeech}. 
As there is no publicly available accelerator with the software and hardware to flexibly support the various techniques in our study, we simulate 16-bit-FPU training using the QPyTorch simulator~\citep{zhang2019qpytorch}.
QPyTorch models PyTorch kernels such as matrix multiplication as compute graph operators, and can effectively simulate \BFHS FMAC units with 32-bit accumulators\footnote{Following the convention in mixed precision training~\citep{micikevicius2017mixed}, our simulator uses fused operators for computationally inexpensive activation and normalization layers.}.
For all training algorithms, we use the same hyperparameters as their original papers or repositories. We report results with averaged metrics and standard deviations across runs with 3 random seeds.
We refer to~\Cref{app:experiment,app:exp:results} for experiment details and extended results.

    \vspace{-0.5em}
\paragraph{The Model Accuracy Bottleneck}
To validate our insights from~\Cref{sec:accurate}, 
we first show empirically that nearest rounding on the model weights is the primary model accuracy bottleneck on several deep learning models. 
To do this, we keep the model weights in 32-bit precision and turn off nearest rounding on the model weight updates while keeping nearest rounding for all other operators in the compute graph. 
\Cref{fig:acc:ablation} shows that the standard 16-bit-FPU training algorithm with nearest rounding on all operators has up to $16\%$ training accuracy gap compared to 32-bit training.
Although this gap can be small in the early training phase, it grows larger in later stages.  
In contrast, by ablating nearest rounding on model weight updates, the standard algorithm can fully match the training accuracy attained by 32-bit training. 
We notice in~\Cref{tab:ablation:test} that this ablation can also close the $1.2\%$ to $6.7\%$ validation accuracy gap when comparing the standard 16-bit-unit training to 32-bit training.
These observations validate our insights from~\Cref{sec:accurate:subsec:understand} and motivate the use of stochastic rounding and Kahan summation on model weight updates.

\begin{table*}
\caption{\textbf{16-bit-FPU training can match 32-bit training on model accuracy.}
With stochastic rounding or Kahan summation for model weight updates, 16-bit-FPU training attains $0.1\%$ lower to $0.2\%$ higher absolute value for validation accuracy metrics across applications. 
}
\ifarxiv
\centering
\begin{tabular}{c  c  c  c  c c}
\toprule
 \multirow{2}{*}{Model} &  \multirow{2}{*}{Dateset (Metric)} &  \multirow{2}{*}{32-bit} & \multicolumn{3}{c}{16-bit-FPU}   \\
  \cmidrule(lr){4-6}
 & & & Stochastic & Kahan & Standard\\
\midrule
 ResNet-18 & CIFAR10 (Acc\%) & $95.45 \pm 0.07$ & $95.33 \pm 0.08$ & $95.36 \pm 0.07$ & $94.23 \pm 0.12$ \\
 ResNet-50 & ImageNet (Acc\%) & $75.70 \pm 0.05$ & $75.45 \pm 0.03$ & $75.61 \pm 0.14$ & $67.10 \pm 0.24$\\
 \multirow{2}{*}{DLRM} & Kaggle (AUC\%) & $80.27 \pm 0.01$ & $80.18 \pm 0.02$ & $80.26 \pm 0.01$ & $78.49 \pm 0.08$ \\ 
 & Terabyte (AUC\%) & $80.32 \pm 0.00$ & $80.25 \pm 0.00$ & $80.32 \pm 0.00$ & $78.79 \pm 0.02$ \\
 \multirow{2}{*}{BERT} &MNLI (Acc\%) & $84.26 \pm 0.08$ & $84.35 \pm 0.12$ & $84.45 \pm 0.03$ & $77.53 \pm 0.07$ \\
 &Wiki103 (PPL) & $5.50 \pm 0.50$ & $5.84 \pm 0.53$ & $5.45 \pm 0.51$ & $56.88 \pm 1.77$ \\
 DeepSpeech2 & Librispeech (WER)& $62.71 \pm 0.07$ & $62.85 \pm 0.07$ & $62.87 \pm 0.18$ & $69.42 \pm 0.22$\\
\bottomrule
\end{tabular}

\else
\small
\centering
\begin{tabular}{c  c  c  c  c c}
\toprule
Model & Dateset (Metric) & 32-bit & 16-bit-FPU Stochastic & 16-bit-FPU Kahan & Standard 16-bit-FPU \\
\midrule
 ResNet-18 & CIFAR10 (Acc\%) & $95.45 \pm 0.07$ & $95.33 \pm 0.08$ & $95.36 \pm 0.07$ & $94.23 \pm 0.12$ \\
 ResNet-50 & ImageNet (Acc\%) & $75.70 \pm 0.05$ & $75.45 \pm 0.03$ & $75.61 \pm 0.14$ & $67.10 \pm 0.24$\\
 \multirow{2}{*}{DLRM} & Kaggle (AUC\%) & $80.27 \pm 0.01$ & $80.18 \pm 0.02$ & $80.26 \pm 0.01$ & $78.49 \pm 0.08$ \\ 
 & Terabyte (AUC\%) & $80.32 \pm 0.00$ & $80.25 \pm 0.00$ & $80.32 \pm 0.00$ & $78.79 \pm 0.02$ \\
 \multirow{2}{*}{BERT} &MNLI (Acc\%) & $84.26 \pm 0.08$ & $84.35 \pm 0.12$ & $84.45 \pm 0.03$ & $77.53 \pm 0.07$ \\
 &Wiki103 (PPL) & $5.50 \pm 0.50$ & $5.84 \pm 0.53$ & $5.45 \pm 0.51$ & $56.88 \pm 1.77$ \\
 DeepSpeech2 & Librispeech (WER)& $62.71 \pm 0.07$ & $62.85 \pm 0.07$ & $62.87 \pm 0.18$ & $69.42 \pm 0.22$\\
\bottomrule
\end{tabular}
\fi
\label{tab:experiment:test}
\end{table*}

\ifarxiv
\begin{figure}
\else
\begin{figure*}
\fi
\centering
\ifarxiv
\includegraphics[width=\textwidth]{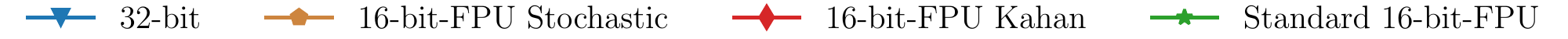} 
\begin{tabular}{c@{\hskip 3em}c}
    \includegraphics[width=0.31\textwidth]{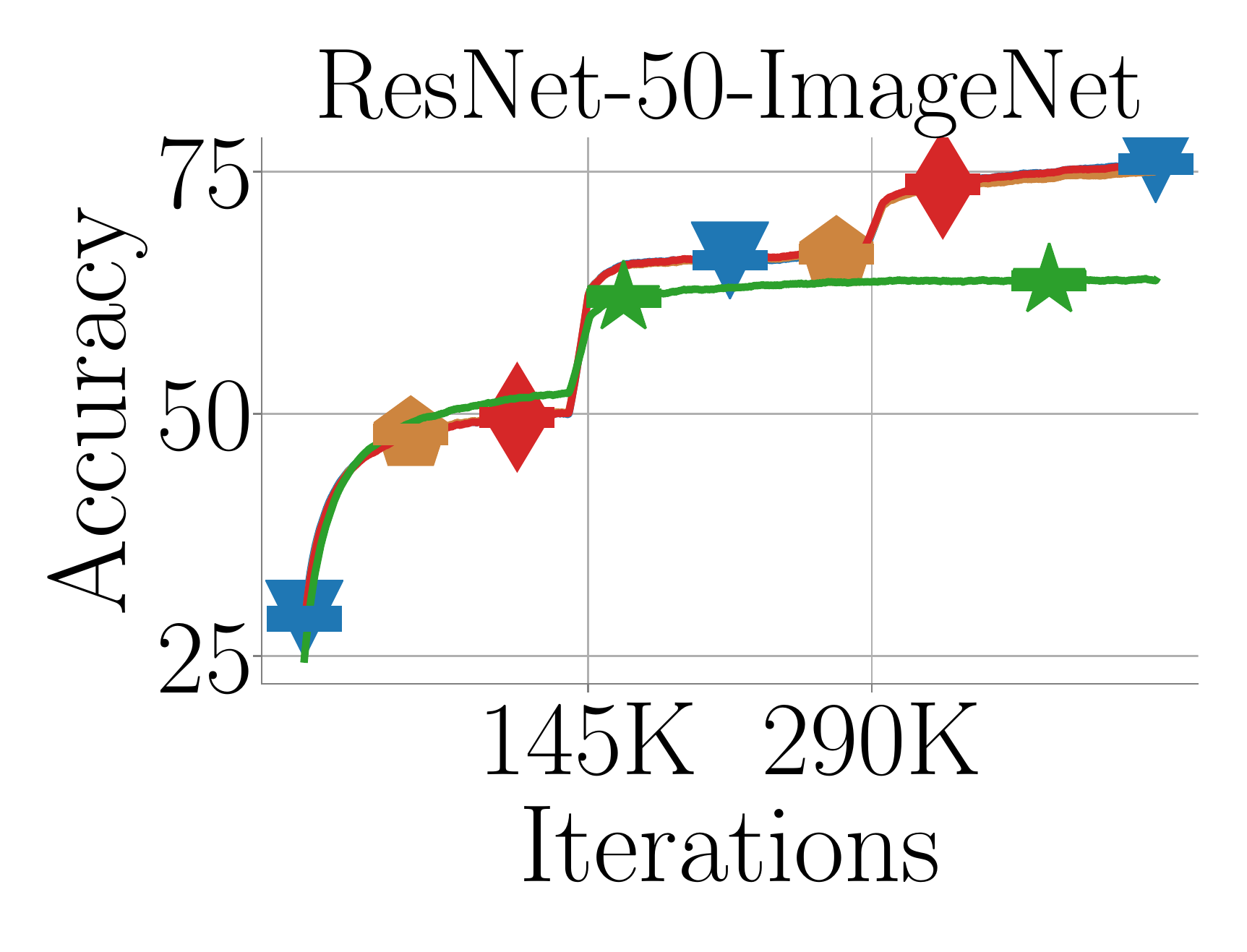} &
    \includegraphics[width=0.31\textwidth]{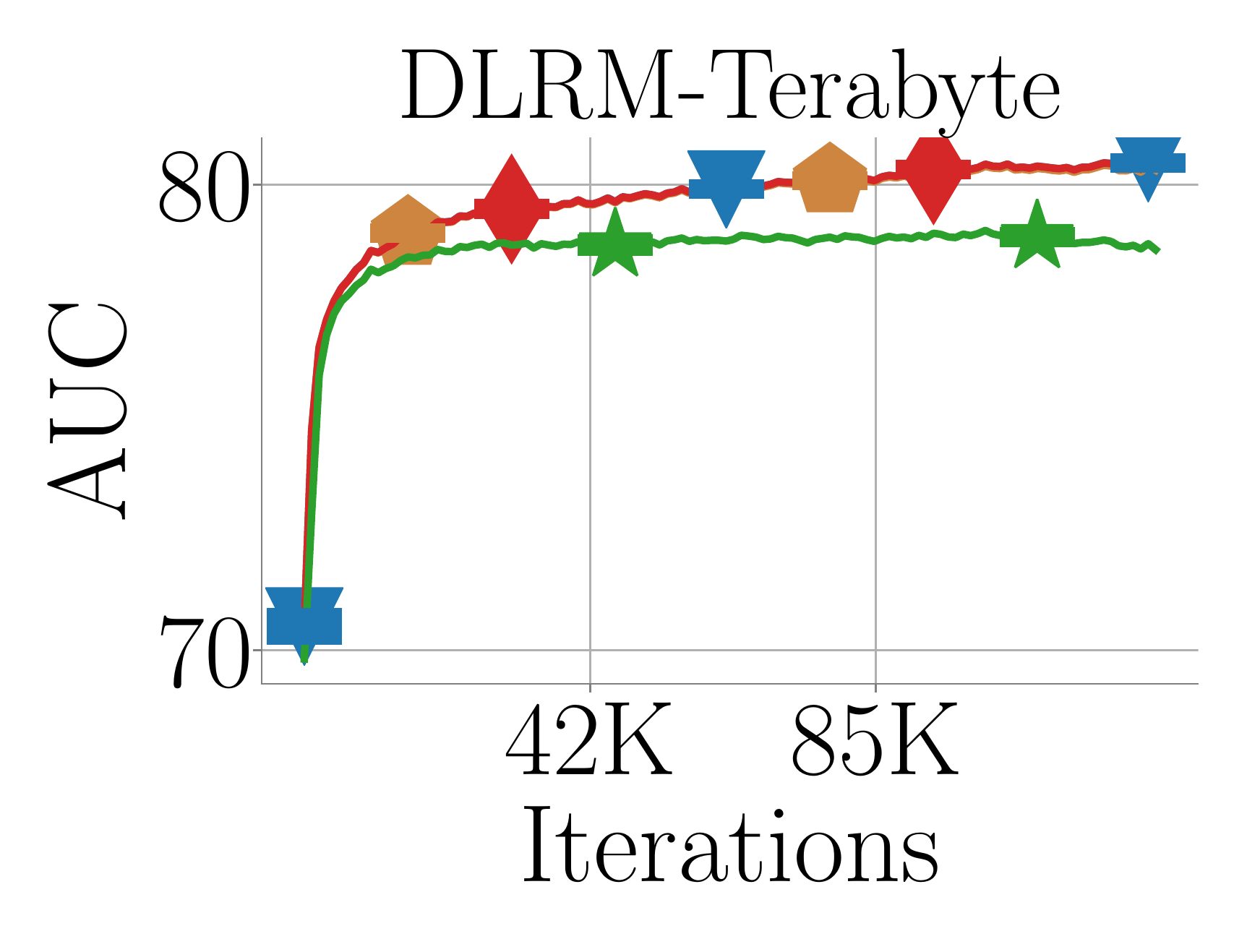} \\
    \includegraphics[width=0.31\textwidth]{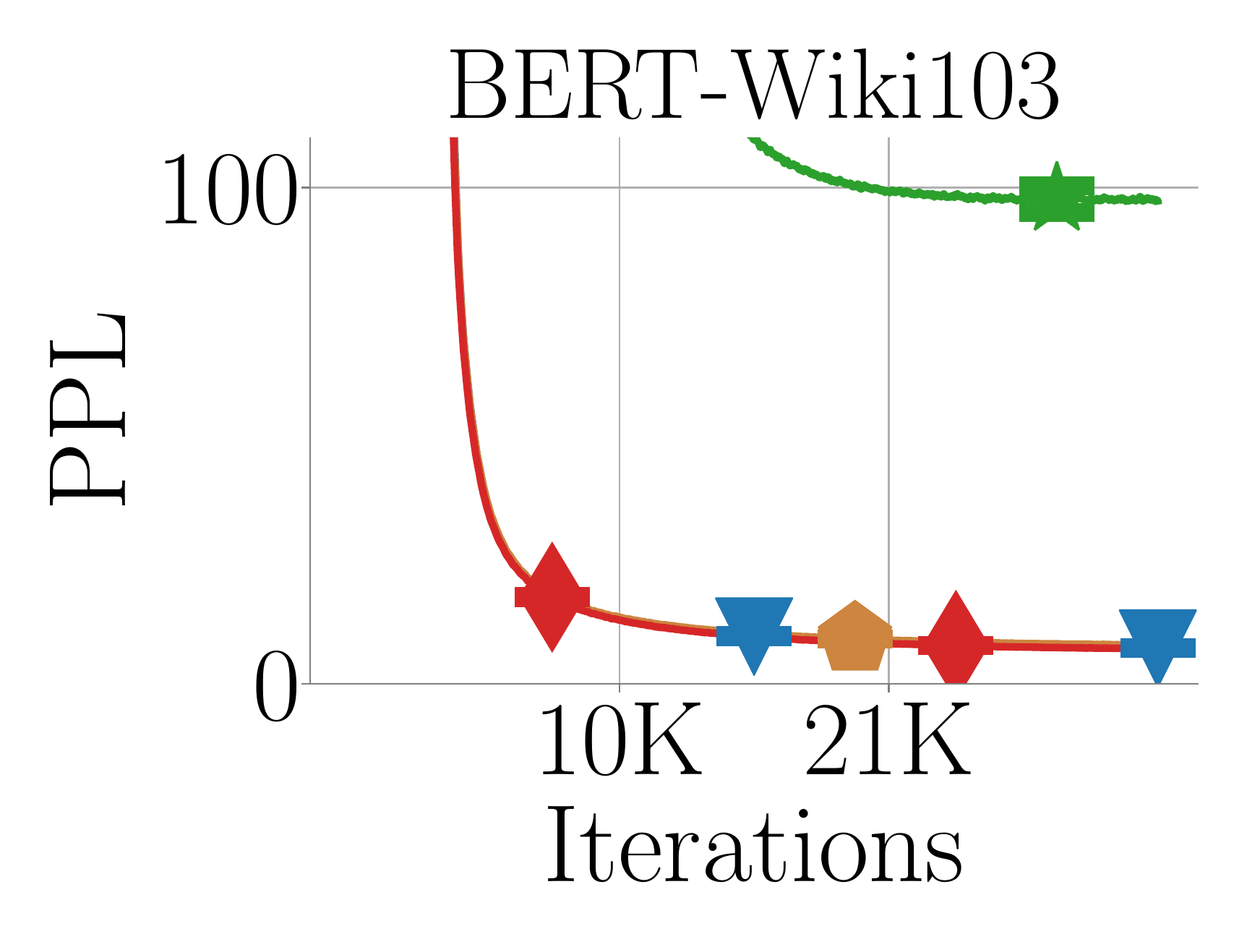} &
    \includegraphics[width=0.31\textwidth]{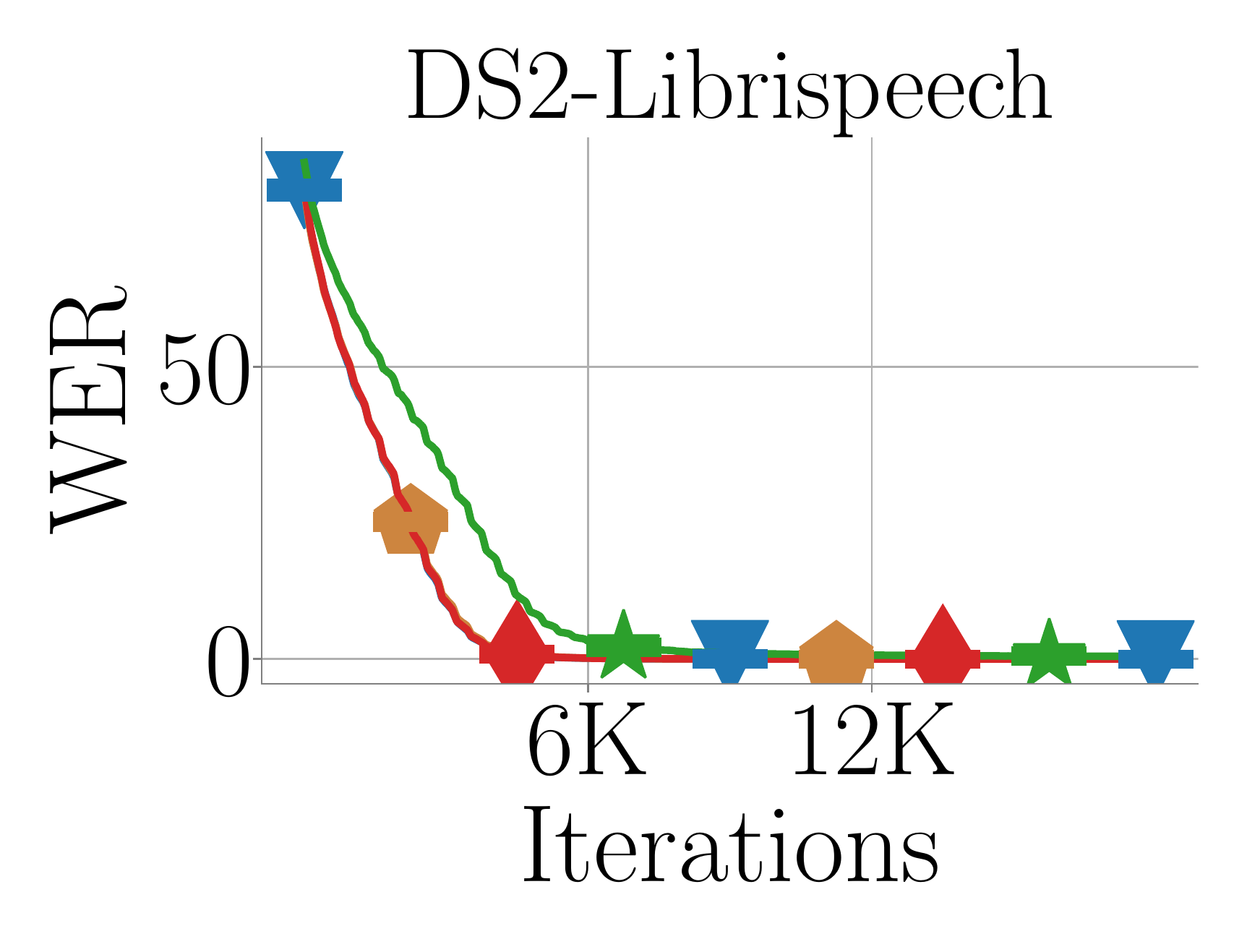}
\end{tabular}
\else
\includegraphics[width=0.975\textwidth]{stochastic_and_kahan/legend} 
\begin{tabular}{c c c c}
    \includegraphics[width=0.2275\textwidth]{stochastic_and_kahan/resnet-50-imagenet_train} &
    \includegraphics[width=0.2275\textwidth]{stochastic_and_kahan/dlrm-terabyte_train} &
    \includegraphics[width=0.2275\textwidth]{stochastic_and_kahan/bert-wiki103_train} &
    \includegraphics[width=0.2275\textwidth]{stochastic_and_kahan/ds2-librispeech_train}
\end{tabular}
\vspace{-0.5em}
\fi
  \caption{\textbf{Training accuracy for 16-bit-FPU training.} With stochastic rounding or Kahan summation enabled for model weight updates, 16-bit-FPU training matches 32-bit training in terms of training accuracy with negligible differences on the applications in our experiments.}
  \label{fig:exp}
\ifarxiv
\end{figure}
\else
\end{figure*}
\fi

\paragraph{High-accuracy 16-bit-FPU Training}
Next, we validate empirically that enabling stochastic rounding or Kahan summation for model weight updates allows 16-bit-FPU training to attain matching model accuracy as 32-bit training.
In~\Cref{tab:experiment:test}, we first show that by using stochastic rounding for model weight updates, 16-bit-FPU training matches the validation accuracy of 32-bit training with at most $0.1\%$ difference on the CIFAR10, Kaggle, Terabyte, MNLI and Librispeech datasets, a majority of the applications in our experiments. 
For applications where stochastic rounding still shows a non-negiglable accuracy gap with more than $0.1\%$ discrepancy, we show that Kahan summation for model weight updates can enable 16-bit-FPU training to match the model accuracy of 32-bit training algorithms.
We show that Kahan summation for model weight updates can boost 16-bit-FPU training to higher validation accuracy than using stochastic rounding.
More concretely, the Kahan summation for model weight updates shows $0.2\%$ higher top-1 accuracy and $0.1\%$ higher AUC respectively for ResNet-50 on ImageNet and for recommendation on Terabyte than using stochastic rounding. Consequently as shown in~\Cref{tab:experiment:test}, by using Kahan summation for model weight updates, 16-bit-FPU training match the model accuracy attained by 32-bit precision training across all the applications in our experiments. 
This validates that stochastic rounding and Kahan summation can enable 16-bit-FPU training algorithms to match the model accuracy of 32-bit training.

\paragraph{Memory efficiency and model accuracy trade-off}
Additionally, we show that stochastic rounding and Kahan summation can be combined for 16-bit-FPU training, which exposes a memory efficiency and model accuracy trade-off for practitioners to exploit. 
In \Cref{fig:tradeoff} we demonstrate this trade-off by incrementally replacing stochastic rounding with Kahan summation on various model weights in the DLRM model on the Kaggle dataset. As we apply Kahan summation to more model weights, the weight memory cost increases by up to $2\times$. As this cost increases we also observe up to $0.04\%$ improvement in AUC.  This exploits a memory efficiency and model accuracy trade-off that users should consider when deciding which technique to leverage.

%% file: related.tex
\section{Related Work}
\label{sec:related}

\ifarxiv
\else
\begin{wrapfigure}[11]{r}{0.2575\textwidth}
			\vspace{-3.95em}
			\begin{minipage}{0.2575\textwidth}
				\vspace{-1.25em}
\begin{figure}[H]
	\centering
	\includegraphics[width=\linewidth]{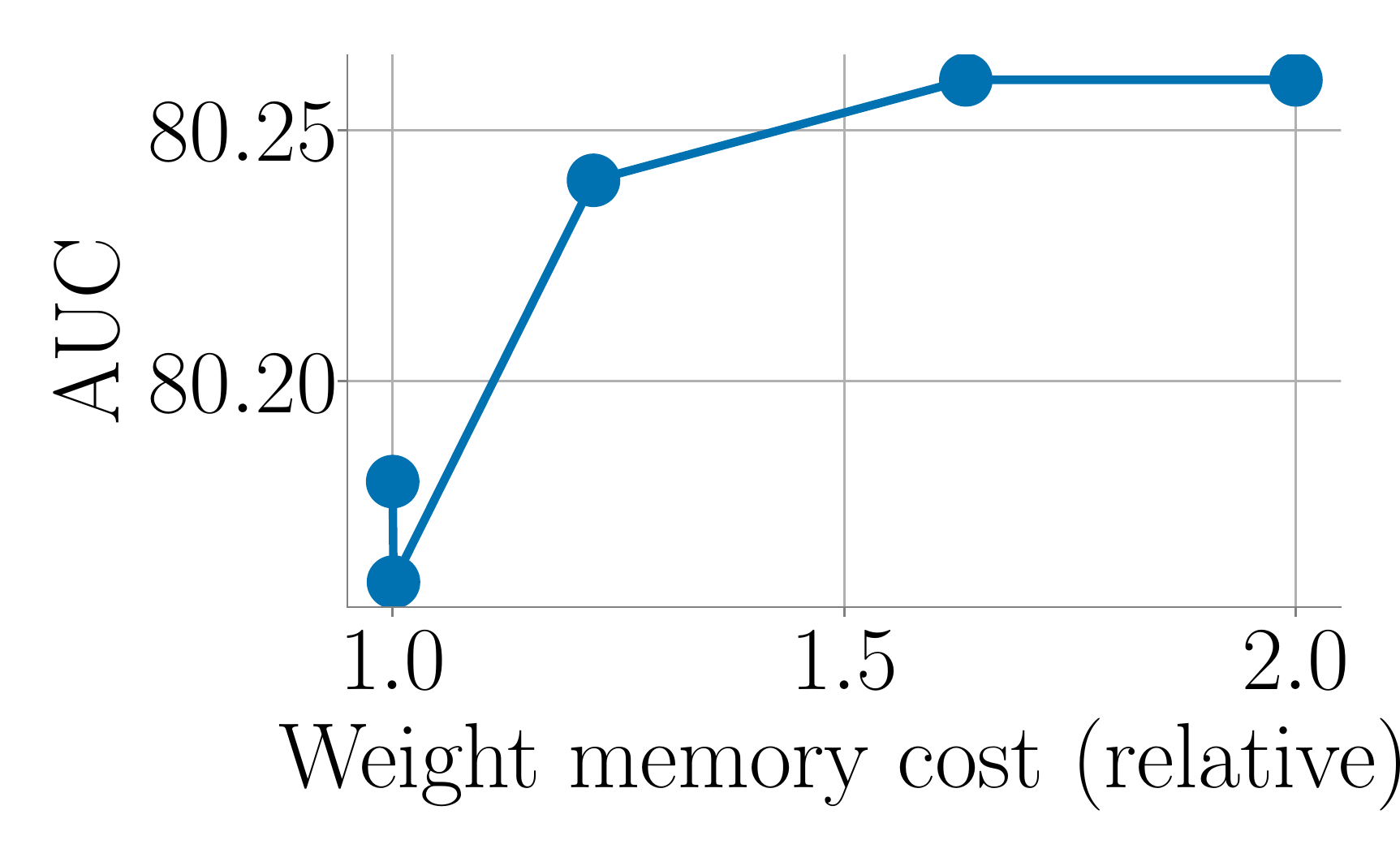}
				\vspace{-2em}
  \caption{\textbf{Efficiency and accuracy trade-off.} 
    With stochastic rounding and Kahan summation on different parts of the DLRM-Kaggle, it attains higher model accuracy at the cost of more weight memory.
  }
  \label{fig:tradeoff}
\end{figure}
			\end{minipage}
		\end{wrapfigure}
\fi
There is a plethora of research work on low-precision training for deep learning models. On certain specific models such as convolutional or recurrent neural networks, training with fixed point or floating point precisions lower than 16-bit has been shown to be feasible when using customized techniques~\citep{wang2018training,zhou2016dorefa,hubara2017quantized,courbariaux2014training,ott2016recurrent,sun2019hybrid}. 
\rebuttal{Instead of proposing new techniques for specific model types using lower than 16-bit precision,
	we focus on identifying the minimal set of simple techniques from numerical analysis literatures which are necessary for future deep learning training accelerators requiring only modern 16-bit FPUs.
	Such emerging accelerators have the potential to unlock substantially improved hardware efficiency compare to those still requiring 32-bit floating-point units.
		}
		
\ifarxiv
\begin{wrapfigure}[16]{r}{0.45\textwidth}
			\begin{minipage}{0.45\textwidth}
				\vspace{-1.25em}
\begin{figure}[H]
	\centering
	\includegraphics[width=0.9\linewidth]{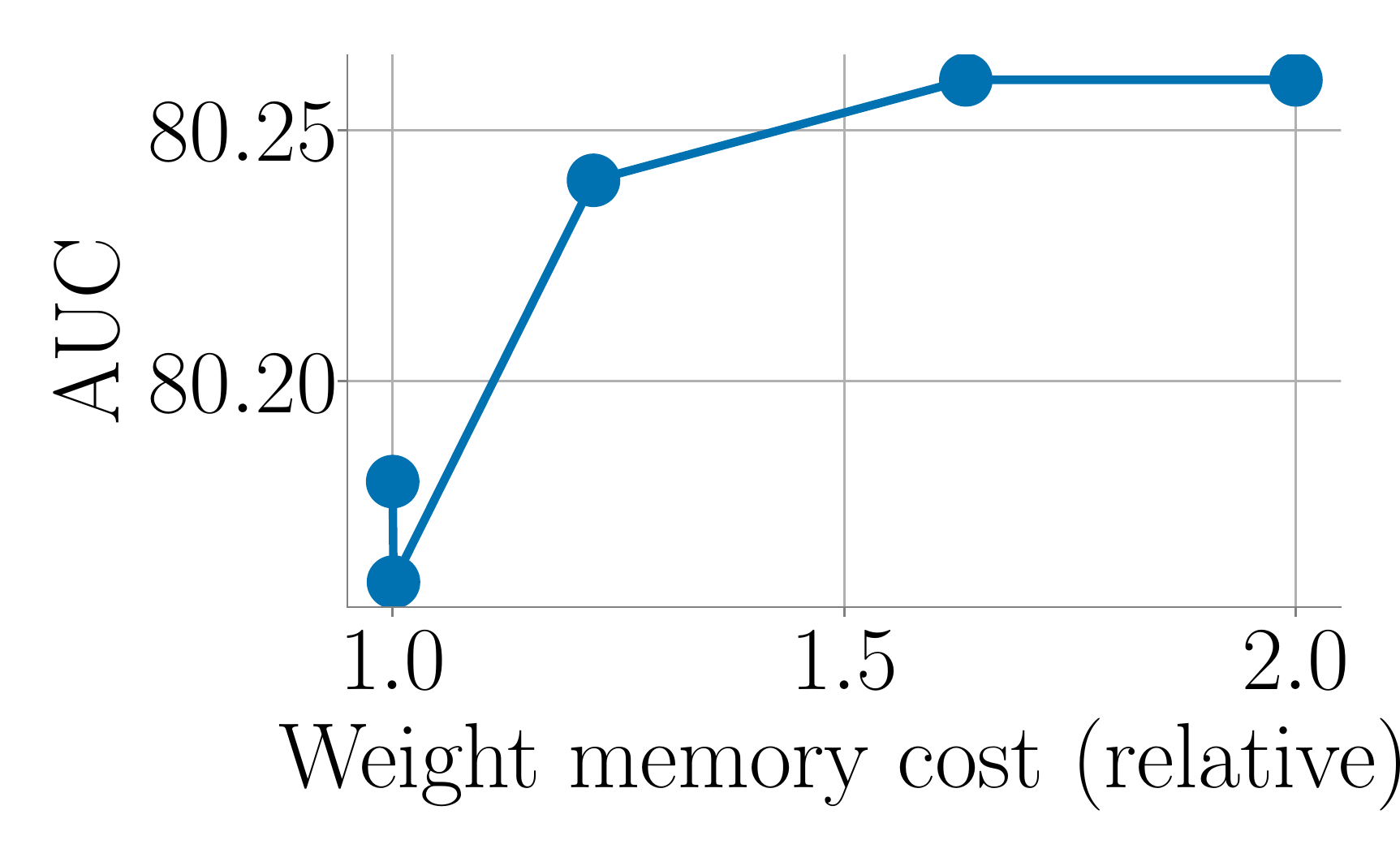}
  \caption{\textbf{Efficiency and accuracy trade-off.} 
    With stochastic rounding and Kahan summation on different parts of the DLRM-Kaggle, it attains higher model accuracy at the cost of more weight memory.
  }
  \label{fig:tradeoff}
\end{figure}
			\end{minipage}
		\end{wrapfigure}
\else
\fi
\rebuttal{Recent work shows that low precision stochastic gradient descent with stochastic rounding for model weight updates only degrades worst-case upper bounds on convergence minimally; this shows that stochastic rounding is an effective technique to attain strong model accuracy in low precision training~\citep{li2017training,hou2018analysis}. Our analysis is complementary to these upper bounds. Specifically, we prove a lower bound on convergence when using standard nearest rounding for model weight updates. This lower bound shows that nearest rounding for weight updates can substantially degrade the convergence no matter how learning rates are tuned. We use this lower bound with the previous upper bounds to inform future accelerator designers that only supporting nearest rounding is not enough to maximize model accuracy; to alleviate this problem, stochastic rounding for model weight updates is one of the minimal supports required in future training accelerators.}

%% file: conclusion.tex
\vspace{-0.25em}
\section{Conclusion}
\vspace{-0.25em}
\label{sec:conclusion}
In this paper we study 16-bit-FPU training algorithms that require only 16-bit floating-point units.
We show that nearest rounding on model weight updates is the primary cause of convergence and model accuracy degradation in standard 16-bit-FPU training. To alleviate this issue, 
we apply two techniques well-known in numerical analysis: stochastic rounding and Kahan summation. With these techniques, we demonstrate that 16-bit-FPU training can match the model accuracy of 32-bit training across many deep learning models.
Our study suggests that it is feasible to design high-accuracy deep learning accelerators using only 16-bit FPUs if stochastic rounding and Kahan summation are supported.

%% file: app_theory.tex
\section{Theory proof}
\label{app:theory}
In this section, we first present the proof for~\Cref{prop:high_prob}. We then discuss~\Cref{theo:covergence_other} to reveal the impact of nearest rounding in the forward and backward pass in the backpropagation compute.
\subsection{Proof of~\Cref{prop:high_prob}}

\begin{proof}
We first prove that when 
\[
    \| \vw - \vw^* \|
    \le
        \frac{\epsilon}{\alpha L + \epsilon} \cdot \min_j \left| w^*_j \right|,
\]  
the model updates across all the dimensions will be canceled in a \emph{single stochastic gradient descent step} on the least-squares regression model. We then prove the lower bound for $\| \vw_t - \vw^* \|$ after using \emph{stochastic gradient descent at iterations $t$}.

We assume that floating-point representation has the property that for any representable number $u\in \R$, all other representable numbers $v\in \R$ have $|\mQ(u) - u| \le \epsilon |u|$.
Also in our Lipschitz continuous gradient assumption (\textbf{A2}), we know that the data magnitude is bounded such that $\|\nabla f_{\sigma (t)}(\vw) - \nabla f_{\sigma (t)}(\vv) \| \leq L \|\vw - \vv\|, \forall \vv, \vw \in \mathbb{R}^d$. Additionally we also assume the overparameterization setting which implies $\nabla f_{\sigma (t)}(\vw^{*})=\bm{0}$ (\textbf{A1}) with $\vw^{\star} =\argmin_{\vw\in \R^d} f(\vw)$.

\paragraph{On the condition to cancel updates for all the dimensions} We consider a model which defines a loss function with Lipschitz continuous gradient. Assume we consider the batch size $1$ setting, the sample gradient at model weight $\vw$ is $ \nabla f_{\sigma(t)}(\vw)$ where the sampled index subset $\sigma(t)\subset \{1,2,.., n \}$ contains a single index.
When the model weight update gets canceled by nearest rounding $\mQ$, we have
\begin{equation}
    \mQ\left(\vw - \alpha \nabla f_i(\vw)\right) = \vw.
\label{equ:prop_proof_0}
\end{equation}
To make this hold, it suffices to show that, for all model weight dimensions $j \in \{1, \ldots, d\}$,
\[
    \left| \left[ \vw - \alpha \nabla f_{\sigma(t)}(\vw) \right]_j - w_j \right| \le \epsilon | w_j |.
\]
with $w_j$ being the $j$-th dimension of $\vw$.
This is equivalent to show that
\begin{equation}
    \alpha \left| \left[ \nabla f_{\sigma(t)}(\vw) - \nabla f_{\sigma(t)}(\vw^{*}) \right]_j \right| \le \epsilon | w_j |.
\label{equ:prop_proof_1}
\end{equation}
because we have $\nabla f_{\sigma (t)}(\vw^{*})=\bm{0}$ from the assumption \textbf{A1}.

In order to prove~\Cref{equ:prop_proof_1}, we first observe that
\begin{equation}
\begin{aligned}
     & \alpha \left| \left[ \nabla f_{\sigma(t)}(\vw) - \nabla f_{\sigma(t)}(\vw^{*}) \right]_j \right| \\
    \le \ &
    \alpha \| \nabla f_{\sigma(t)}(\vw) - \nabla f_{\sigma(t)}(\vw^{*})  \|\\
    \le \ &
    \alpha L \cdot \| \vw - \vw^* \|.
\end{aligned}
\label{equ:prop_proof_2}
\end{equation}
where the last inequality is due to the assumption \textbf{A2} with Lipschitz continuous gradients.

We also observe that
\begin{equation}
\begin{aligned}
    | w_j |
    &=
    \left| \left[\vw - \vw^*\right]_j + w^*_j \right|
    \\&\ge
    \left| w^*_j \right| - \left| \left[\vw - \vw^*\right]_j \right|
    \\&\ge
    \min_j \left| w^*_j \right| - \| \vw - \vw^* \|.
\label{equ:prop_proof_3}
\end{aligned}
\end{equation} with $w^*_j$ being the $j$-th dimension of $\vw^*$.

\Cref{equ:prop_proof_2} and~\Cref{equ:prop_proof_3} shows that, to cancel weight updates, namely satisfying~\Cref{equ:prop_proof_0}, it suffices to prove that
\begin{equation}
    \alpha L \cdot \| \vw - \vw^* \|
    \le
    \epsilon \min_j \left| w^*_j \right| - \epsilon \| \vw - \vw^* \|.
    \label{equ:prop_proof_4}
\end{equation}
When we have
\[
    \| \vw - \vw^* \|
    \le
    \frac{\epsilon}{ \alpha L + \epsilon} \cdot \min_j \left| w^*_j \right|,
\]
\Cref{equ:prop_proof_4} is satisfied. This proves that the weight updates will be canceled with 
\[\mQ\left(\vw - \alpha \nabla f_i(\vw)\right) = \vw.\]

\paragraph{Lower bound on convergence}

\emph{Case I:} the model is initialized far away from the optimum $\vw^*$ with $ \| \vw_0 - \vw^* \|
    \ge
    \frac{\epsilon}{ \alpha L + \epsilon} \cdot \min_j \left| w^*_j \right|$.
    
    To prove the lower bound 
\begin{equation}
	    \| \vw_t - \vw^* \|
    \ge
    \frac{\epsilon\left(1 - \alpha L \right)}{ \alpha L + \epsilon} \cdot \min_j \left| w^*_j \right|.
    \label{equ:theo_1_proof_1}
\end{equation} for any $t\in\{1,2,... \}$, we discuss two possible situations.

In the first situation, if for any $t$, $\vw_t$ never steps into the quantization noise ball with center $\vw^*$ and radius $\frac{\epsilon}{ \alpha L + \epsilon} \cdot \min_j \left| w_j^*\right|$. 
Then the the bound in~\Cref{equ:theo_1_proof_1} directly holds.

In the second situation, we assume model weights step into the quantization noise ball at a certain iteration $t + 1$ for the first time. Thus at iteration $t$ we have that
\begin{align*}
     &\| \vw_{t+1} - \vw^* \| \\
    =\ &
    \| \vw_t - \vw^* - \alpha \nabla f_{\sigma(t)}(\vw_t) \|
    \\=\ &
    \| \vw_t  -  \vw^* - \alpha \left( \nabla f_{\sigma(t)}(\vw_t) - \nabla f_{\sigma(t)}(\vw^{*}) \right) \|
    \\= \ &
    \left| \| \vw_t - \vw^* \| - \alpha \| \nabla f_{\sigma(t)}(\vw_t) - \nabla f_{\sigma(t)}(\vw^{*})  \| \right|
    \\ \ge\ &
    \| \vw_t - \vw^* \| - \alpha L  \| \vw_t - \vw^* \|
    \\ \ge\ &
    (1 - \alpha L) \| \vw_t - \vw^* \|.
\end{align*}
where the second last inequality is because of the facts that $\nabla f_{\sigma(t)}(\vw)$ is $L$-Lipschitz continuous and we use a learning rate $\alpha \leq 1/L$.

Because $\vw_t$ is still outside of the quantization noise ball at iteration $t$, we have $\| \vw_t - \vw^* \| \geq \frac{\epsilon}{ \alpha L + \epsilon} \cdot \min_j \left| w_j^*\right|$. Thus we have 
\begin{equation}
	    \| \vw_{t+1} - \vw^* \| \geq \frac{\epsilon\left(1 - \alpha L \right)}{ \alpha L + \epsilon} \cdot \min_j \left| w^*_j \right|.
\end{equation}

\emph{Case II:} the model is initialized in the quantization noise ball with $ \| \vw_0 - \vw^* \|
    \le
    \frac{\epsilon}{ \alpha L + \epsilon} \cdot \min_j \left| w^*_j \right|$. Because  the model weights halt inside the quantization noise ball, we have that 
    
    \begin{equation}
    \begin{aligned}
	    & \| \vw_t - \vw^* \|
    \\\ge \ &
    \min\left(\frac{\epsilon \left(1 - \alpha L \right)}{ \alpha L + \epsilon} \cdot \min_j \left| w^*_j \right|, \|\vw_0 - \vw^* \|\right).
    \label{equ:theo_1_proof_4}
    \end{aligned}
\end{equation}

\end{proof}

\subsection{Proof of~\Cref{theo:covergence_other}}
\label{app:subsec:A2}
\begin{theorem}
Consider running multiple steps of SGD on a least-squares regression model under assumptions \textbf{A1} and \textbf{A2}, using nearest rounding for only forward and backward compute, but exact arithmetic for model weight updates.
Then if the step size is small enough that $\alpha \le 1/L$, the distance of the weights $\vw_t$ at any timestep $t$ from the optimum will be bounded by
\[
    \textstyle
    \mathbf{E}\left[ \| \vw_t - \vw^* \|^2 \right] 
    \le
    \exp\left(- \alpha \mu t \left( 1 - \frac{4 \epsilon L}{\mu}  \right) \right) \cdot \| \vw_0 - \vw^* \|^2,
\]
where $\mu$ is the smallest eigenvalue of the data covariance matrix $\frac{1}{n} \sum_{i=1}^n \vx_i \vx_i^T$.
\label{theo:covergence_other}
\end{theorem}
As $t$ can be made arbitrarily large, this bound guarantees us substantially more accurate solutions than the lower bound attained by using nearest rounding only for model weight updates in~\Cref{prop:high_prob}. This \emph{shows that rounding for the weight updates is the primary source of error}, as even with all other operations quantized, the algorithm is guaranteed to converge closer to the optimum than is even possible with just the weight updates rounded with nearest rounding. Note that the bound in Theorem~\ref{theo:covergence_other} is able to guarantee arbitrarily accurate solutions because we ignore underflow here. In practice, precision would eventually be limited by underflow even in the setting of Theorem~\ref{theo:covergence_other}; however, the underflow threshold for \BFHS is small enough that this represents a level of error that deep learning applications are generally able to tolerate.  We prove~\Cref{theo:covergence_other} as follows.

\begin{proof}
Under the overparameterization assumption $y_i = \vx_i^T \vw^{*}$ (\textbf{A1}) and bounded data assumption $\|\vx_i \|^2 \leq L$ (\textbf{A2}), we first observe that the error from the activation quantization is bounded by
\begin{align*}
    \left| \mQ(\vx_i^T \vw_t - y_i) - (\vx_i^T \vw_t - y_i) \right|
    &\le
    \epsilon \left| \vx_i^T \vw_t - y_i \right|
    \\&=
    \epsilon \left| \vx_i^T (\vw_t - \vw^*) \right|
    \\&\le
    \epsilon \sqrt{L} \| \vw_t - \vw^* \|.
\end{align*}
Note that here, we are assuming no numerical errors happen inside the dot product because we suppose it is computed by a single FMAC with a higher-precision accumulator.
It follows that
\begin{equation}
\begin{aligned}
    &\left\| \mQ(\vx_i^T \vw_t - y_i) \vx_i - (\vx_i^T \vw_t - y_i) \vx_i \right\|
    \\\le \ &
    \| \vx_i \| \cdot 
    \left| \mQ(\vx_i^T \vw_t - y_i) - (\vx_i^T \vw_t - y_i) \right|
    \\\le \ &
    \epsilon L \| \vw_t - \vw^* \|.
\end{aligned}
\label{equ:app:theo2_1}
\end{equation}
The second quantization is redundant in the least-squares regression case because the already quantized activation will be leveraged as the activation gradient. And the additional activation gradient quantization would not take any effect to introduce new numerical error.
For the third quantization, we observe that its error in the $j$-th coordinate will be bounded by
\begin{align*}
    &\left| \mQ(\mQ(\vx_i^T \vw_t - y_i) \vx_i)_j - (\mQ(\vx_i^T \vw_t - y_i) \vx_i)_j \right|
    \\\le \ &
    \epsilon \left| (\mQ(\vx_i^T \vw_t - y_i) \vx_i)_j \right|,
\end{align*}
which implies that
\begin{align*}
    &\left\| \mQ(\mQ(\vx_i^T \vw_t - y_i) \vx_i) - (\mQ(\vx_i^T \vw_t - y_i) \vx_i) \right\|
    \\= \ &
    \sqrt{\sum_{j=1}^n \left| \mQ(\mQ(\vx_i^T \vw_t - y_i) \vx_i)_j - (\mQ(\vx_i^T \vw_t - y_i) \vx_i)_j \right|^2}
    \\\le \ &
    \sqrt{\sum_{j=1}^n \epsilon^2 \left| (\mQ(\vx_i^T \vw_t - y_i) \vx_i)_j \right|^2}
    \\= \ &
    \epsilon \left\| \mQ(\vx_i^T \vw_t - y_i) \vx_i \right\|
    \\\le \ &
    \epsilon \left\| \vx_i \vx_i^T (\vw_t - \vw^*) \right\| + \mathcal{O}(\epsilon^2)
    \\\le \ &
    \epsilon L \| \vw_t - \vw^* \| + \mathcal{O}(\epsilon^2).
\end{align*}
where the second last inequality is a consequence of~\Cref{equ:app:theo2_1}. If we disregard the $\epsilon^2$ term, we have that
\[
\begin{aligned}
    &\left\| \mQ(\mQ(\vx_i^T \vw_t - y_i) \vx_i) - (\vx_i^T \vw_t - y_i) \vx_i \right\|
    \\\le \ &
    2 \epsilon L \| \vw_t - \vw^* \|.
\end{aligned}
\]
This means that we can write our gradient update step as
\[
    \vw_{t+1} = \vw_t - \alpha \vx_{\sigma(t)} \vx_{\sigma(t)}^T (\vw_t - \vw^*) + \bm{\eta}_t,
\]
where $\sigma(t)$ is an index from the dataset sampled uniformly at random, and $\bm{\eta_t}$ is an error term bounded by $\| \bm{\eta}_t \| \le 2 \alpha \epsilon L \| \vw_t - \vw^* \|$.
This gives
\[
    \vw_{t+1} - \vw^* = \left( \mI - \alpha \vx_{\sigma(t)} \vx_{\sigma(t)}^T \right) (\vw_t - \vw^*) + \bm{\eta}_t,
\]
Taking the norm and the expected value gives
\begin{align*}
    &\mathbf{E} \| \vw_{t+1} - \vw^* \|^2 
    \\= \ &
    \mathbf{E}\Big[ (\vw_t - \vw^*)^T \left( \mI - \alpha \vx_{\sigma(t)} \vx_{\sigma(t)}^T \right)^2 (\vw_t - \vw^*) 
    \\+ \ & 
    2 \bm{\eta}_t^T \left( \mI - \alpha \vx_{\sigma(t)} \vx_{\sigma(t)}^T \right) (\vw_t - \vw^*)
    +
    \| \bm{\eta}_t \|^2 \Big].
\end{align*}
If we let $\Sigma$ denote the expected value
\[
    \bm{\Sigma} = \frac{1}{n} \sum_{i=1}^n \vx_i \vx_i^T,
\]
then
\[
\begin{aligned}
    &\mathbf{E}\left[ \left(\vx_{i_t} \vx_{i_t}^T \right)\left(\vx_{i_t} \vx_{i_t}^T \right)^T \right]
    \\= \ &
    \mathbf{E}\left[ \|\vx_{\sigma(t)}\|^2 \vx_{\sigma(t)} \vx_{\sigma(t)}^T \right]
    \preceq
    L \Sigma.
\end{aligned}
\]
Because of this, we have that
\begin{align*}
    &\mathbf{E} \| \vw_{t+1} - \vw^* \|^2 
    \\\le\ &
    \mathbf{E}\Big[ (\vw_t - \vw^*)^T \left( \mI - 2 \alpha \bm{\Sigma} + \alpha^2 L \bm{\Sigma} \right) (\vw_t - \vw^*) 
    \\+\ & 
    2 \bm{\eta_t}^T \left( \mI - \alpha \bm{\Sigma} \right) (\vw_t - \vw^*)
    +
    \| \bm{\eta_t} \|^2 \Big].
\end{align*}
Since we assumed $\alpha L \le 1$, with an additional application of Cauchy-Schwarz, we can simplify this to
\[
\begin{aligned}
    &\mathbf{E} \| \vw_{t+1} - \vw^* \|^2 
    \\\le\ &
    \mathbf{E}\Big[ (\vw_t - \vw^*)^T \left( \mI - \alpha \bm{\Sigma} \right) (\vw_t - \vw^*) 
    \\ +\ & 
    2 \| \bm{\eta_t} \| \| \vw_t - \vw^* \|
    +
    \| \bm{\eta_t} \|^2 \Big].
\end{aligned}
\]
If we let $\mu$ denote the smallest eigenvalue of $\bm{\Sigma}$, then this can be bounded by
\[
\begin{aligned}
    &\mathbf{E} \| \vw_{t+1} - \vw^* \|^2 
    \\\le\ &
    \mathbf{E}\Big[ (1 - \alpha \mu) \| \vw_t - \vw^* \|^2
    \\+\ & 
    2 \| \bm{\eta_t} \| \| \vw_t - \vw^* \|
    +
    \| \bm{\eta_t} \|^2 \Big].
\end{aligned}
\]
Substituting our bound on $\bm{\eta}$ gives
\[
\begin{aligned}
    &\mathbf{E} \| \vw_{t+1} - \vw^* \|^2 
    \\\le\ &
    \mathbf{E}\Big[ (1 - \alpha \mu) \| \vw_t - \vw^* \|^2
    + 
    4 \alpha \epsilon L \| \vw_t - \vw^* \|^2
    \\+\ &
    4 \alpha^2 \epsilon^2 L^2 \| \vw_t - \vw^* \|^2 \Big].
\end{aligned}
\]
Disregarding the $\epsilon^2$ term, we have
\[
\begin{aligned}
    &\mathbf{E} \| \vw_{t+1} - \vw^* \|^2 
    \\\le\ &
    \mathbf{E}\Big[ (1 - \alpha \mu) \| \vw_t - \vw^* \|^2
    + 
    4 \alpha \epsilon L \| \vw_t - \vw^* \|^2 \Big],
\end{aligned}
\]
which simplifies to
\[
    \mathbf{E} \| \vw_{t+1} - \vw^* \|^2 
    \le
    (1 - \alpha \mu + 4 \alpha \epsilon L) \mathbf{E}\left[ \| \vw_t - \vw^* \|^2 \right].
\]
In other words, as long as $\epsilon L \ll \mu$ (i.e. $\epsilon$ is small relative to the inverse of the condition number $\kappa = L/\mu$ of the problem, we will still converge at a linear rate.
More explicitly,
\[
    \mathbf{E}\left[ \| \vw_t - \vw^* \|^2 \right] 
    \le
    (1 - \alpha \mu + 4 \alpha \epsilon L)^t \cdot \| \vw_0 - \vw^* \|^2.
\]
Or,
\[
    \mathbf{E}\left[ \| \vw_t - \vw^* \|^2 \right] 
    \le
    \exp\left(- \alpha \mu t \left( 1 - 4 \epsilon \kappa \right) \right) \cdot \| \vw_0 - \vw^* \|^2.
\]
	
\end{proof}

%% file: algo.tex
\section{High-accuracy 16-bit-FPU Training}
\label{app:algo}
In this section, we present the algorithm for SGD and AdamW~\citep{loshchilov2017decoupled} when combined with stochastic rounding or Kahan summation on the model weight updates in~\Cref{alg:stoc_sgd,alg:kahan_sgd_app,alg:stoc_adam,alg:kahan_adam}; these four algorithms describe the optimizers in our experiments. 
\rebuttal{In these algorithms, all the scalars and tensors such as model weights and optimizer states are in \BFHS precision.} 
On top of these values, all floating point arithmetic operators take \BFHS inputs and use nearest rounding for the output unless noticerd otherwise.
For SGD and AdamW with stochastic rounding on the model weight updates, we define operator $\ominus$ as a subtraction with \BFHS inputs and stochastic rounding for the output.
Across our experiments using these optimizers, to calculate minibatch gradient $\nabla f_{\sigma(t)}(\vw_t)$, forward and backward compute operators consume \BFHS input and perform nearest rounding for the output. 

\rebuttal{To demonstrate the hardware efficiency of 16-bit-FPU training algorithms in~\Cref{alg:stoc_sgd,alg:kahan_sgd_app,alg:stoc_adam,alg:kahan_adam}, we discuss the hardware overhead of using stochastic rounding and Kahan summation for model weight updates in~\Cref{app:algo:subsect:efficiency_stoc,app:algo:subsect:efficiency_kahan} respectively.}

\subsection{System Efficiency of Stochastic Rounding}
\label{app:algo:subsect:efficiency_stoc}
\rebuttal{In modern hardware, training with stochastic rounding for model weight updates can be implemented efficiently. It has been demonstrated that stochastic rounding can be implemented without any expensive multiplication or division arithmetic~\cite{de2017understanding}. Instead, it only requires 1) generating random bit sequences with a shift register, 2) adding random sequence to the lower mantissa bits and 3) truncating; these operations are inexpensive relative to other optimizer operations regardless of the model size. Secondly, we note that the minimal technique in our study only requires stochastic rounding for model weight updates while even cheaper nearest rounding remains for forward, backward and optimizer operations other than weight updates. Because of the above two reasons, using stochastic rounding for model weight updates only adds minimal overhead compared to standard 16-bit training fully with nearest rounding.}

\subsection{System Efficiency of Kahan Summation}
\label{app:algo:subsect:efficiency_kahan}
\rebuttal{Despite the fact that 16-bit Kahan accumulation requires an additional 16-bit auxiliary value, it can still demonstates advantages over 32-bit and mixed precision training in terms of throughput and memory consumption. In more details, optimizers in both 32-bit and mixed precision training operate on 32-bit weights (for mixed precision, the weights here refer to the master copy) and 32-bit optimizer states such as momentum. On the other hand, in 16-bit training with Kahan summation, the optimizers leverage fully 16-bit weights, optimizer states and auxiliary variables.}

\rebuttal{\emph{Regarding the throughput}, optimizers in 32-bit and mixed precision training require floating-point units with 32-bit multiply, while optimizers in 16-bit training with Kahan summation can leverage units with 16-bit multiply. Given that 2) modern MAC units with 16-bit multiply can be implemented with 1.5X throughput compared to those with 32-bit multiply [10] and 3) Kahan summation only introduce 3 additional add/subtract operations which are inexpensive relative to multiply (e.g. Adam has 9 major multiply operations), optimizers in 16-bit training with Kahan summation can demonstrate meaningful speedup. }

\rebuttal{\emph{Regarding the memory efficiency}, using Adam optimizer as an example, 16-bit training with Kahan summation saves $33\%$ and $43\%$ memory for weights plus optimizer states respectively over 32-bit training and mixed precision training (mixed precision training has both 16-bit and 32-bit weights in memory). We note that these benefits here are especially pronounced for training large SOTA models with billions of model weights using optimizers with sophisticated optimizer states~\citep{shoeybi2019megatron,rajbhandari2019zero}.}

\begin{algorithm}[t]
	\caption{SGD with Stochastic Rounding}
	\label{alg:stoc_sgd}
	\begin{algorithmic}[1]
	     \State \textbf{INPUT:} learning rate $\eta_t$, momentum $\mu$, weight decay $d$, gradient $\nabla f_{\sigma(t)}(
        \vw_t)$ at iteration t
		\While {$t < T$}
        \State $\vg_{t+1} \leftarrow \nabla f_{\sigma(t)}(
        \vw_t) + d * \vw_t$
		\State $ \vm_{t+1} \leftarrow \mu * \vm_{t} + \vg_{t+1}$
		\State $ \vw_{t+1} \leftarrow \vw_{t} \ominus \left( \eta_t * \vm_{t+1}\right)$
		\State $t \leftarrow t + 1$
        \EndWhile
		\State \textbf{RETURN:} $\vw_{T}$
	\end{algorithmic}
\end{algorithm}

\begin{algorithm}[t]
    \caption{SGD with Kahan Summation}
    \label{alg:kahan_sgd_app}
    \begin{algorithmic}[1]
	     \State \textbf{INPUT:} learning rate $\eta_t$, momentum $\mu$, weight decay $d$, gradient $\nabla f_{\sigma(t)}(
        \vw_t)$ at iteration t
        \State $\vc_{0} \leftarrow 0$ %
        \While {$t < T$}
        \State $\vg_{t+1} \leftarrow \nabla f_{\sigma(t)}(
        \vw_t) + d * \vw_t$
        \State $ \vm_{t+1} \leftarrow \mu * \vm_{t} + \vg_{t+1}$
        \State $\vu_{t+1} \leftarrow -\left( \eta_t * \vm_{t+1}\right)$
        \State $\vy_{t+1} \leftarrow \vu_{t+1} - \vc_t$
        \State $\vs_{t+1} \leftarrow \vw_{t} + \vy_{t+1}$
        \State $\vc_{t+1} \leftarrow \left(\vs_{t+1} - \vw_t\right) - \vy_{t+1}$ %
        \State $ \vw_{t+1} \leftarrow \vs_{t+1}$ 
        \State $t \leftarrow t + 1$ 
        \EndWhile
        \State \textbf{RETURN:} $\vw_{T}$
    \end{algorithmic}
\end{algorithm}

\begin{algorithm}[t]
	\caption{AdamW with Stochastic Rounding}
	\label{alg:stoc_adam}
	\begin{algorithmic}[1]
	     \State \textbf{INPUT:} learning rate $\eta_t$, beta $\left(\beta_1, \beta_2\right)$, weight decay $d$, gradient $\nabla f_{\sigma(t)}(
        \vw_t)$ at iteration t
		\State $c_{1, 0} \leftarrow 1, c_{2, 0} \leftarrow 1$
		\While {$t < T$}
        \State $\vg_{t+1} \leftarrow \nabla f(
        \vw_t, \vx_t)$
		\State $ \vm_{t+1} \leftarrow \beta_1 * \vm_{t} + (1 - \beta_1) * \vg_{t}$
		\State $ \vv_{t+1} \leftarrow \beta_2 * \vv_{t} + \left(1-\beta_2\right) * \vg_t^2$
		\State $ c_{1, t + 1} \leftarrow c_{1, t} * \beta_1$
		\State $ c_{2, t + 1} \leftarrow c_{2, t} * \beta_2$
		\State $\hat{\vm}_{t + 1} \leftarrow \vm_{t+1} / (1-c_{1, t+1})$
		\State $\hat{\vv}_{t + 1} \leftarrow \sqrt{\vv_{t+1} / (1-c_{2, t+1})}$
		\State $\vw_{t + 1}$ \par\hskip\algorithmicindent
		$\leftarrow \vw_{t} \ominus \left(\eta_t * \hat{\vm}_{t+1} / \left(\hat{\vv}_{t+1} + \epsilon\right) + \eta_t * d * \vw_t \right)$
		\State $t \leftarrow t + 1$ 
        \EndWhile
		\State \textbf{RETURN:} $\vw_{T}$
	\end{algorithmic}
\end{algorithm}

\begin{algorithm}[t]
	\caption{AdamW with Kahan Summation}
	\label{alg:kahan_adam}
	\begin{algorithmic}[1]
	     \State \textbf{INPUT:} learning rate $\eta_t$, betas $\left(\beta_1, \beta_2\right)$, weight decay $d$, gradient $\nabla f_{\sigma(t)}(
        \vw_t)$ at iteration t
        		\State $c_{1, 0} \leftarrow 1, c_{2, 0} \leftarrow 1$
        \State $\vc_{0} \leftarrow 0$
		\While {$t < T$}
        \State $\vg_t \leftarrow \nabla f(
        \vw_t, \vx_t)$
		\State $ \vm_{t+1} \leftarrow \beta_1 * \vm_{t} + (1 - \beta_1) * \vg_{t}$
		\State $ \vv_{t+1} \leftarrow \beta_2 * \vv_{t} + \left(1-\beta_2\right) * \vg_t^2$
		\State $ c_{1, t + 1} \leftarrow c_{1, t} * \beta_1$
		\State $ c_{2, t + 1} \leftarrow c_{2, t} * \beta_2$
		\State $\hat{\vm}_{t + 1} \leftarrow \vm_{t+1} / (1-c_{1, t+1})$
		\State $\hat{\vv}_{t + 1} \leftarrow \sqrt{\vv_{t+1} / (1-c_{2, t+1})}$
		\State $\vu_{t+1} \leftarrow - \left(\eta_t * \hat{\vm}_{t+1} / \left(\hat{\vv}_{t+1} + \epsilon\right) + \eta_t * d * \vw_t \right)$
		\State $\vy_{t+1} \leftarrow \vu_{t+1} - \vc_t$
        \State $\vs_{t+1} \leftarrow \vw_{t} + \vy_{t+1}$
        \State $\vc_{t+1} \leftarrow \left(\vs_{t+1} - \vw_t\right) - \vy_{t+1}$ 
        \State $ \vw_{t+1} \leftarrow \vs_{t+1}$
		\State $t \leftarrow t + 1$ 
        \EndWhile
		\State \textbf{RETURN:} $\vw_{T}$
	\end{algorithmic}
\end{algorithm}

%% file: app_experiment.tex
\section{Experiment Details}
\label{app:experiment}
In this section, we first discuss the details in the experiment setup. We first discuss the hyperparameter values we use in~\Cref{app:exp:subsec:hyp}. We then present the infrastructure details used in our experiments in~\Cref{app:infra}.

\subsection{Hyperparameters}
\label{app:exp:subsec:hyp}
For each model in our expriment, we use the hyperparameter values acquired from the original paper or code repository. In the rest of this section, we discuss the detailed hyperparameter values for all the seven deep learning applications in our experiments.

\paragraph{ResNet-18-CIFAR10\protect\footnote{\url{https://github.com/kuangliu/pytorch-cifar}}}
We train the model for 350 epochs using the SGD optimizer with mini-batch size, weight decay, and momentum values of $128$, $5\times10^{-4}$, and $0.9$ respectively. The learning rate starts from 0.1 and We manually divide it by $10$ at epochs 150 and 250. We summarize these hyperparameters in \Cref{tab:appendix:resnet18}.

\begin{table}[h]
\vspace{0.08in}
\caption{Training hyperparameter for ResNet-18-CIFAR10
}
\small
\centering
\begin{tabular}{c c  }
\toprule
Heyparameter & Value\\
\midrule
Optimizer & SGD\\
Batchsize & 128\\
Training epochs & 350\\
\multirow{3}{*}{Learning rate} &0-150 epochs: 0.1\\
&150-250 epochs: 0.01\\
&250-350 epochs: 0.001\\
weight decay & $5\times10^{-4}$\\
momentum & 0.9\\
\bottomrule
\end{tabular}
\label{tab:appendix:resnet18}
\end{table}

\paragraph{ResNet-50-ImageNet\protect\footnote{\url{https://github.com/pytorch/examples/tree/master/imagenet}}}
We train the model for 90 epochs using the SGD optimizer with mini-batch size, weight decay, and momentum values of $256$, $1\times10^{-4}$, and $0.9$ respectively. The learning rate starts from 0.1 and we manually divide it by $10$ after every 30 epochs. We summarize these hyperparameters in \Cref{tab:appendix:resnet50}.

\begin{table}[h]
\vspace{0.08in}
\caption{Training hyperparameter for ResNet-50-ImageNet
}
\small
\centering
\begin{tabular}{c c }
\toprule
Heyparameter & Value\\
\midrule
Optimizer & SGD\\
Batchsize & 256\\
Training epochs & 90\\
\multirow{3}{*}{Learning rate} & 0-30 epochs: 0.1\\
& 30-60 epochs: 0.01\\
& 60-90 epochs: 0.001\\
Weight decay & $1\times10^{-4}$\\
Momentum & 0.9\\
\bottomrule
\end{tabular}
\label{tab:appendix:resnet50}
\end{table}

\paragraph{BERT-MNLI\protect\footnote{\url{https://github.com/huggingface/transformers}}}
We fine-tune the model for 3 epochs using the AdamW optimizer with mini-batch size and first order momentum values of $256$, $0.9$ respectively. 
The learning rate starts from $2\times10^{-5}$ and linearly decays to 0 during the training. Note in 16-bit-FPU training algorithms, we use $\beta_2 = 0.997$. This is because $0.999$ is considered as $1$ in \BFHS format; we use the $0.997$ which is a \BFHS representable value that is the closest to $0.999$ and is smaller than $1$. We summarize these hyperparameters in \Cref{tab:appendix:mnli}. 

\begin{table}[h]
\vspace{0.08in}
\caption{Training hyperparameter for BERT-MNLI
}
\small
\centering
\begin{tabular}{c c }
\toprule
Heyparameter & Value\\
\midrule
Model type & Base\\
Optimizer & AdamW\\
Batchsize & 64\\
Training epochs & 3\\
Learning rate & $2\times10^{-5}$\\
Decay rate for 1st moment $\beta_1$ (32-bit and 16-bit) & 0.9\\
Decay rate for 2nd moment $\beta_2$ (32-bit)& 0.999\\
Decay rate for 2nd moment $\beta_2$ (16-bit)& 0.997\\
\bottomrule
\end{tabular}
\label{tab:appendix:mnli}
\end{table}

\paragraph{BERT-Wiki103\protect\footnote{\url{https://github.com/pytorch/fairseq/blob/master/examples/roberta/README.pretraining.md}}}
We pre-train the model for 31250 iterations using the AdamW optimizer with mini-batch size, weight decay, first-order momentum, and second-order momentum values of $512$, $0.01$, $0.9$, and $0.98$ respectively; compared to the original training steps, we scale down the training steps proportionally because we subsampled the datasets. The first $8\%$ of training is used for learning rate warm-up and the peak learning rate value is $1\times10^{-4}$. Throughout the rest of the training (remaining $92\%$), learning rate decays to 0 linearly. We summarize these hyperparameters in \Cref{tab:appendix:wiki103}.

\begin{table}[h]
\vspace{0.08in}
\caption{Training hyperparameter for BERT-wiki103
}
\small
\centering
\begin{tabular}{c c }
\toprule
Heyparameter & Value\\
\midrule
Model type & Base\\
Sub-sample rate & 25\%\\
Optimizer & AdamW\\
Batchsize & 512\\
Peak learning rate & $1\times10^{-4}$\\
Total iterations & 31250\\
Warmup iterations & 2500\\
Weight decay & 0.01\\
Decay rate for 1st moment $\beta_1$ & 0.9\\
Decay rate for 2nd moment $\beta_2$ & 0.98\\
\bottomrule
\end{tabular}
\label{tab:appendix:wiki103}
\end{table}

\paragraph{DLRM-Kaggle\protect\footnote{\url{https://github.com/facebookresearch/dlrm}\label{FBDLRM}}}
We fine-tune the model for 1 epoch using the SGD optimizer with mini-batch size of $128$. The learning rate has constant value of 0.1 for the whole training. We summarize these hyperparameters in \Cref{tab:appendix:kaggle}.

\begin{table}[h]
\vspace{0.08in}
\caption{Training hyperparameter for DLRM-Criteo Kaggle
}
\small
\centering
\begin{tabular}{c c }
\toprule
Heyparameter & Value\\
\midrule
Sparse feature size & 16\\
Bottom MLP & 13-512-256-64-16\\
Top MLP & 512-256-1\\
Optimizer & SGD\\
Batchsize & 128\\
Training epochs & 1\\
Learning rate & 0.1\\
Loss function & BCE\\
\bottomrule
\end{tabular}
\label{tab:appendix:kaggle}
\end{table}

\paragraph{DLRM-Terabyte}
For this model, we obtain hyperparameters from NVIDIA\footnote{\url{https://github.com/NVIDIA/DeepLearningExamples/blob/master/PyTorch/Recommendation/DLRM/dlrm/scripts/main.py}} DLRM repository while we use the Facebook\footref{FBDLRM} implementation. The reason is that we want to keep the implementation the same as DLRM-Kaggle experiment; however, the NVIDIA hyperparameters presents the state-of-the-art time-to-accuracy metric. We fine-tune the model for 1 epoch using the SGD optimizer with mini-batch size of $32768$. The first $5\%$ of training is used for learning rate warm-up and the peak learning rate value is $28$. The learning rate decay starts at the middle ($50\%$ of iterations) of the training, and throughout the rest of the training (remaining $50\%$), learning rate decays to 0 linearly. We summarize these hyperparameters in \Cref{tab:appendix:terabyte}.

\begin{table}[h]
\vspace{0.08in}
\caption{Training hyperparameter for DLRM-Criteo Terabyte
}
\small
\centering
\begin{tabular}{c c }
\toprule
Heyparameter & Value\\
\midrule
Sparse feature size & 16\\
Bottom MLP & 512-256-128\\
Top MLP & 1024-1024-512-256-2\\
Optimizer & SGD\\
Batchsize & 32768\\
Training epochs & 1\\
Learning rate & 28\\
Warmup iterations & 6400\\
Learning rate decay start Step & 64000\\
Learning rate decay steps & 80000\\
Loss function & BCE\\
\bottomrule
\end{tabular}
\label{tab:appendix:terabyte}
\end{table}

\paragraph{DLRM-Kaggle\protect\footnote{\url{https://github.com/SeanNaren/deepspeech.pytorch/releases/tag/v2.0}}}
We train the model for 60 epochs using the SGD optimizer with mini-batch size, weight decay, and first order momentum values of $64$, $1\times10^{-5}$, and $0.9$ respectively. The learning rate starts from $3\times10^{-4}$ and is reduced by $1\%$ after each epoch. We summarize these hyperparameters in \Cref{tab:appendix:deepspeech}.

\begin{table}[h]
\vspace{0.08in}
\caption{Training hyperparameter for Deepspeech2-Librispeech
}
\small
\centering
\begin{tabular}{c c }
\toprule
Heyparameter & Value\\
\midrule
RNN type & LSTM\\
Sub-sampled rate & $10\%$\\
Optimizer & SGD\\
Batchsize & 64\\
Training epochs & 60\\
Learning rate & $3\times10^{-4}$\\
Momentum & 0.9\\
Weight decay & $1\times10^{-5}$\\
Max-norm & 400\\
\bottomrule
\end{tabular}
\label{tab:appendix:deepspeech}
\end{table}

\subsection{Infrastructure Details}
\label{app:infra}
We use AWS p3.16xlarge instances for the large applications including BERT-Wiki103, DeepSpeech2-LibriSpeech and ResNet-50-ImageNet. For the rest of the applications,  we train on p3.2xlarge. The AWS p3.2xlarge instance has one NVIDIA's Tesla V100 GPU, and p3.16xlarge has 8 of them. For all the experiments, we use Python version 3.8.1, PyTorch version 1.5, and CUDA version 10.2.

\section{Extended Empirical Results}
\label{app:exp:results}
In this section, we discuss additional results complementary to those in~\Cref{sec:exp}. In~\Cref{app:exp:bottleneck}, we discuss the validation accuracy curves in our ablation study in~\Cref{sec:exp} where we validate the model accuracy bottlenecks. In~\Cref{app:exp:acc}, we present the validation accuracy curves for 16-bit-FPU training algorithms with stochastic rounding or Kahan summation for model weight updates. Finally in~\Cref{app:experiment:subsec:other}, we present additional results on the following four aspects: 1) The percentage of model weight updates which are canceled during training for a representative deep learning model; 2) The model accuracy attained with lower than 16-bit precision format when stochastic rounding or Kahan summation is enabled for model weight updates; 3) The model accuracy attained by 16-bit-FPU training when we combine stochastic rounding and Kahan summation on all the model weights; 4) the model accuracy of 16-bit-FPU training using \FPHS instead of \BFHS with stochastic rounding or Kahan summation enabled.

\subsection{The Model Accuracy Bottleneck}
\label{app:exp:bottleneck}
\begin{figure*}
\begin{subfigure}[b]{\textwidth}
\centering
\includegraphics[width=0.9\textwidth]{accurate/legend}
\end{subfigure}
  \begin{subfigure}[b]{0.32\textwidth}
    \includegraphics[width=\textwidth]{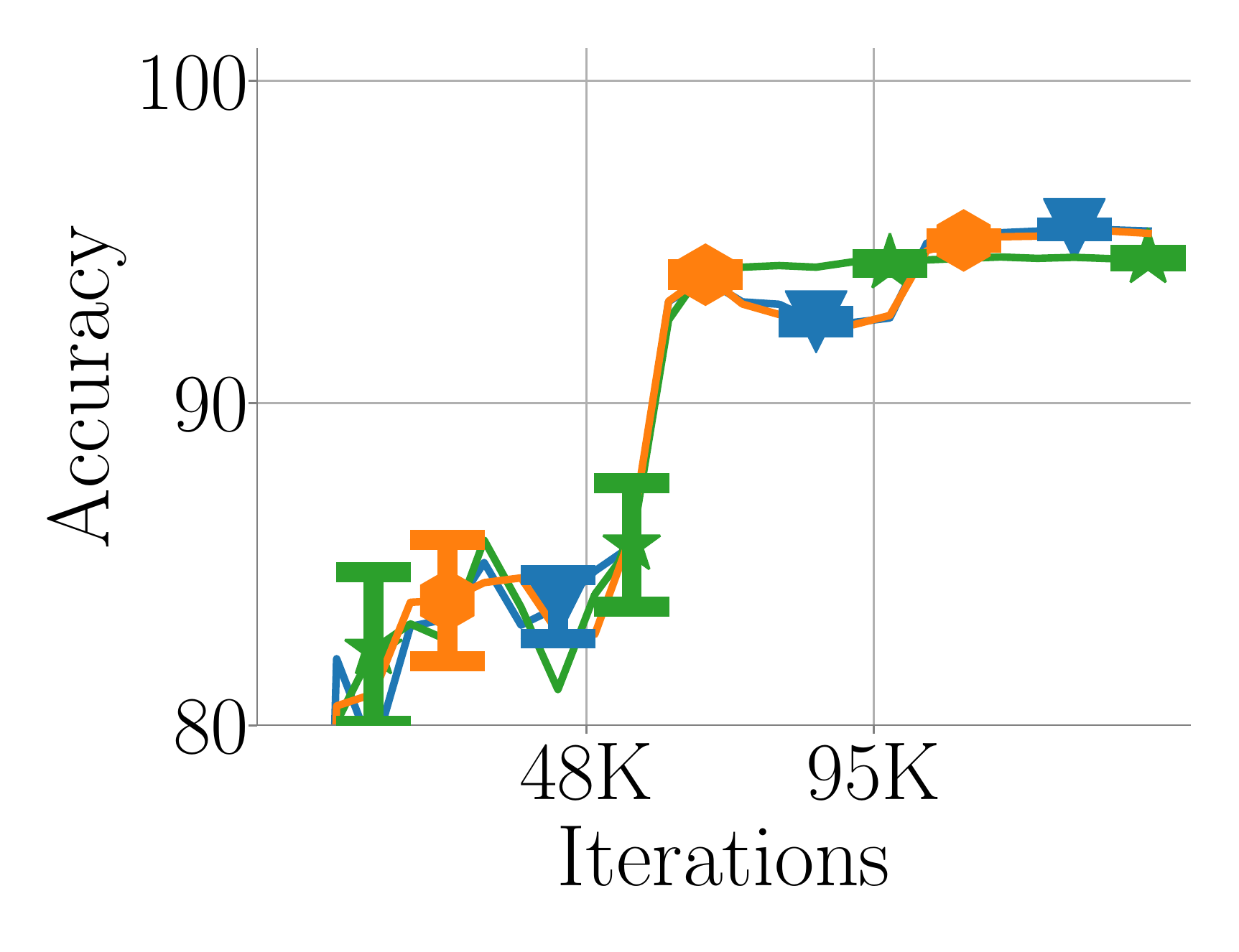}
    \caption{ResNet-18-CIFAR10}
  \end{subfigure}
  \hfill
  \begin{subfigure}[b]{0.32\textwidth}
    \includegraphics[width=\textwidth]{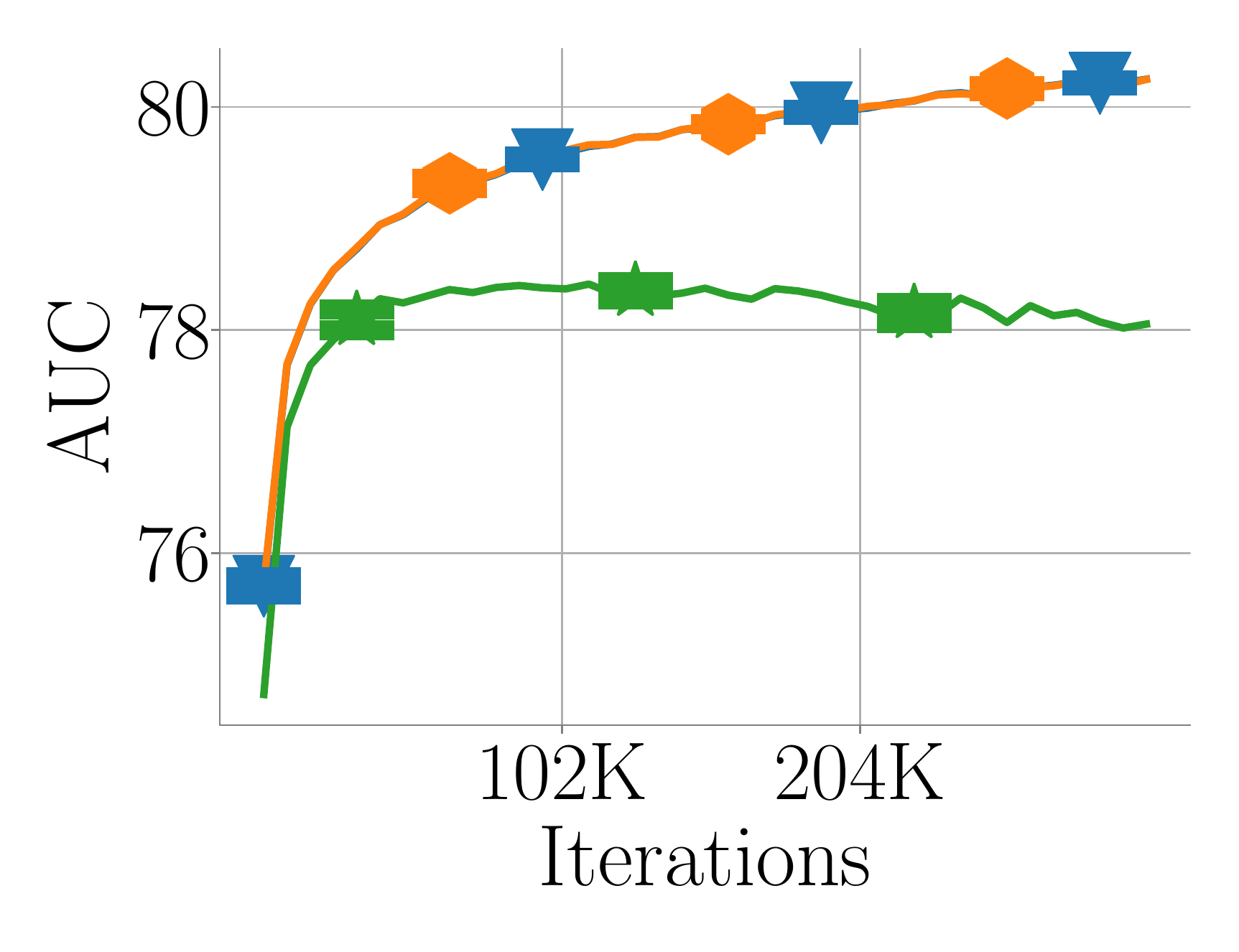}
    \caption{DLRM-Kaggle}
  \end{subfigure}
  \begin{subfigure}[b]{0.32\textwidth}
    \includegraphics[width=\textwidth]{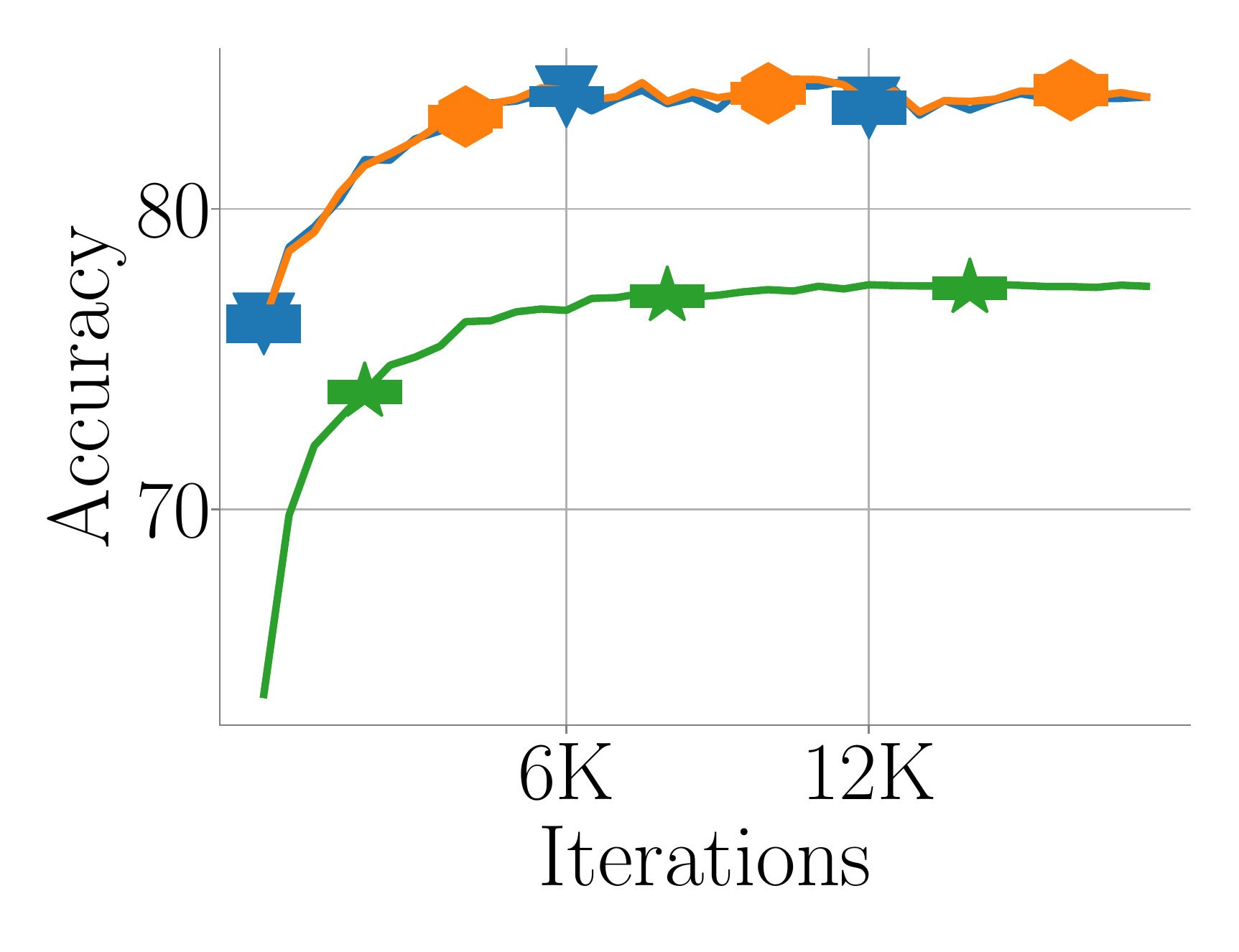}
    \caption{BERT-MNLI}
  \end{subfigure}
  \caption{\textbf{Validation accuracy gap imposed by the standard 16-bit-FPU training.} The standard algorithm fails to match the validation accuracy from 32-bit precision training. By ablating nearest rounding for model weight update from the standard method, we can recover the accuracy gap. This indicates the dominating impact of nearest rounding for model weight updates.}
  \label{fig:app:ablation}
\end{figure*}

In~\Cref{fig:app:ablation}, we present the validation accuracy curves in our ablation study to validate the accuracy bottleneck. We can observe that the standard 16-bit-FPU training algorithm demonstrates lower validation accuracy metric values compared to 32-bit training across the applications. In the ablation, we save the model weights in 32-bit, turn off nearest rounding on model weight updates and keep nearest rounding for all the other compute operators. This ablation isolates the influence of nearest rounding on model weight updates.
    We can see that this ablated algorithm can match the validation accuracy attained by 32-bit precision training. These results together with the ones in~\Cref{sec:exp} shows that nearest rounding on model weight udpates is the primary cause for model accuracy gap in training deep learning models.
    
\begin{figure}
\centering
    \includegraphics[width=0.4\textwidth]{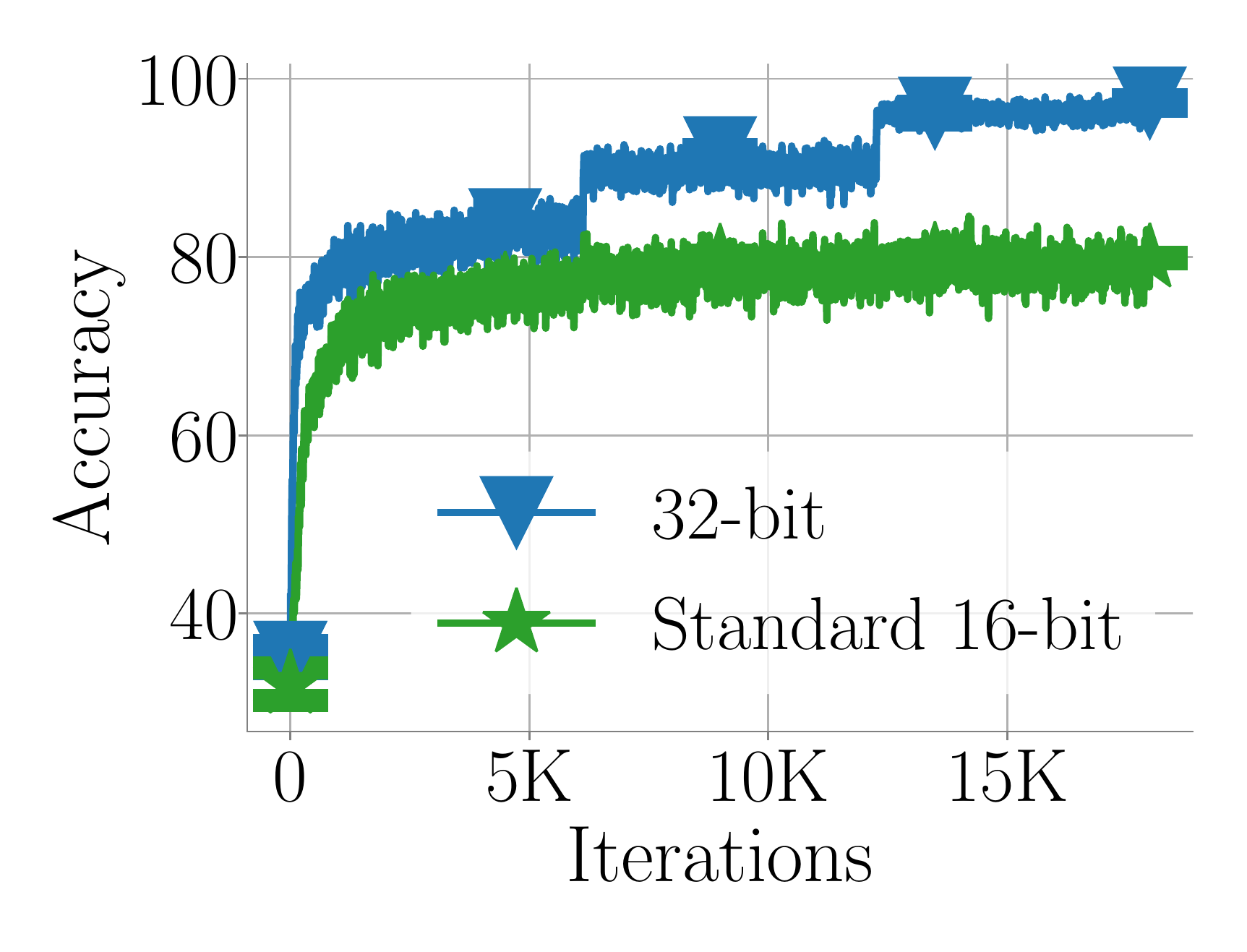}
    \caption{Unsmoothed training accuracy at different number of training iterations for BERT-MNLI.}
	\label{fig:unsmoothed_teaser}
\end{figure}

\rebuttal{In our experiments, 32-bit and standard 16-bit training start from the same initialization with the same model accuracy. For the natural language inference task in~\Cref{fig:teaser}, the gap at the beginning of the curves is due to aggressive smoothing for the clarify in visualization. In~\Cref{fig:unsmoothed_teaser}, we can observe that the unsmoothed curves for 32-bit training and standard 16-bit training indeed start from the same training accuracy level at the beginning.}

\subsection{High-accuracy 16-bit-FPU Training}
\label{app:exp:acc}
\begin{figure*}
\centering
\begin{subfigure}[b]{\textwidth}
\includegraphics[width=\textwidth]{stochastic_and_kahan/legend}
\end{subfigure}
  \begin{subfigure}[b]{0.24\textwidth}
    \includegraphics[width=\textwidth]{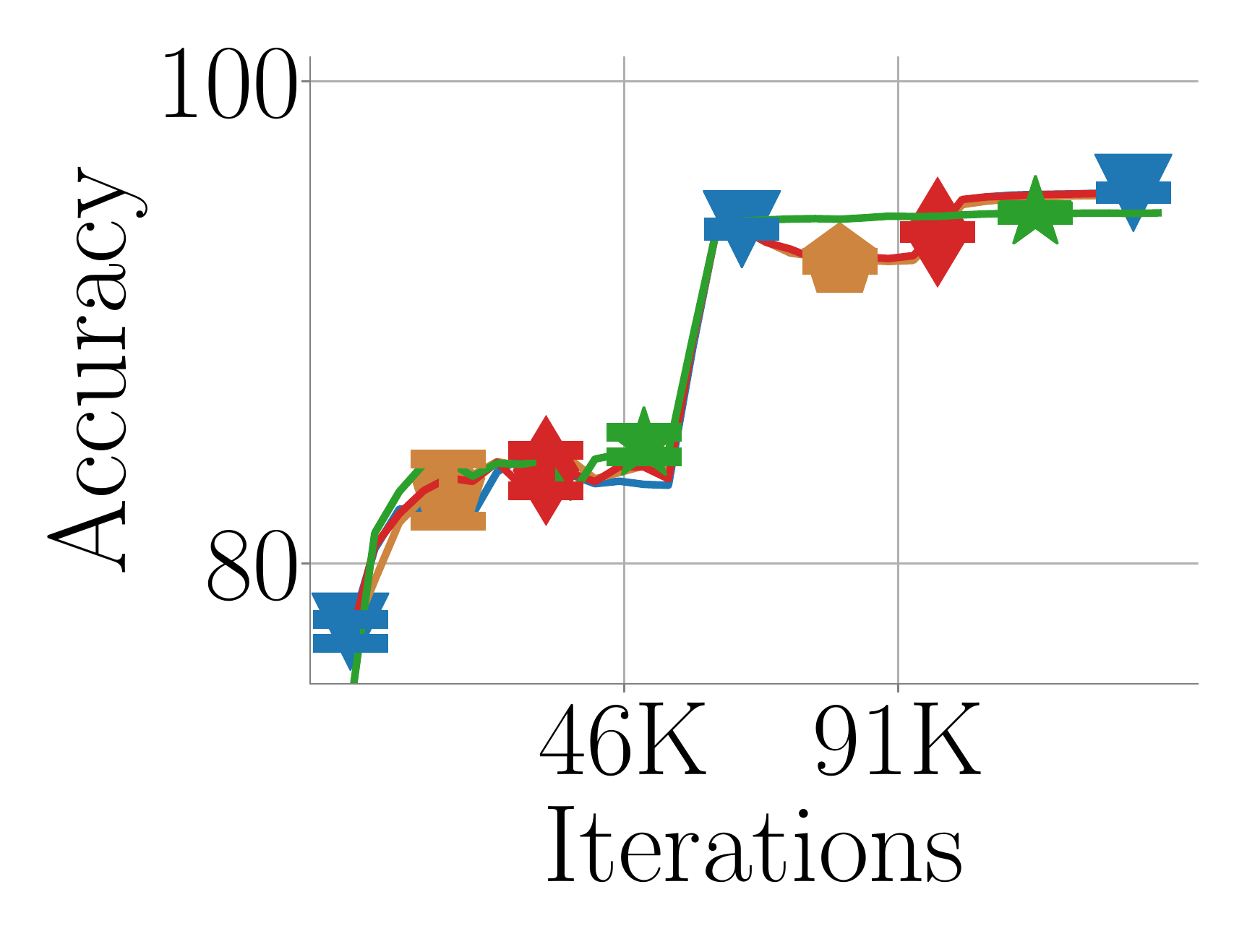}
    \caption{ResNet-18-CIFAR10}
  \end{subfigure}
  \begin{subfigure}[b]{0.24\textwidth}
    \includegraphics[width=\textwidth]{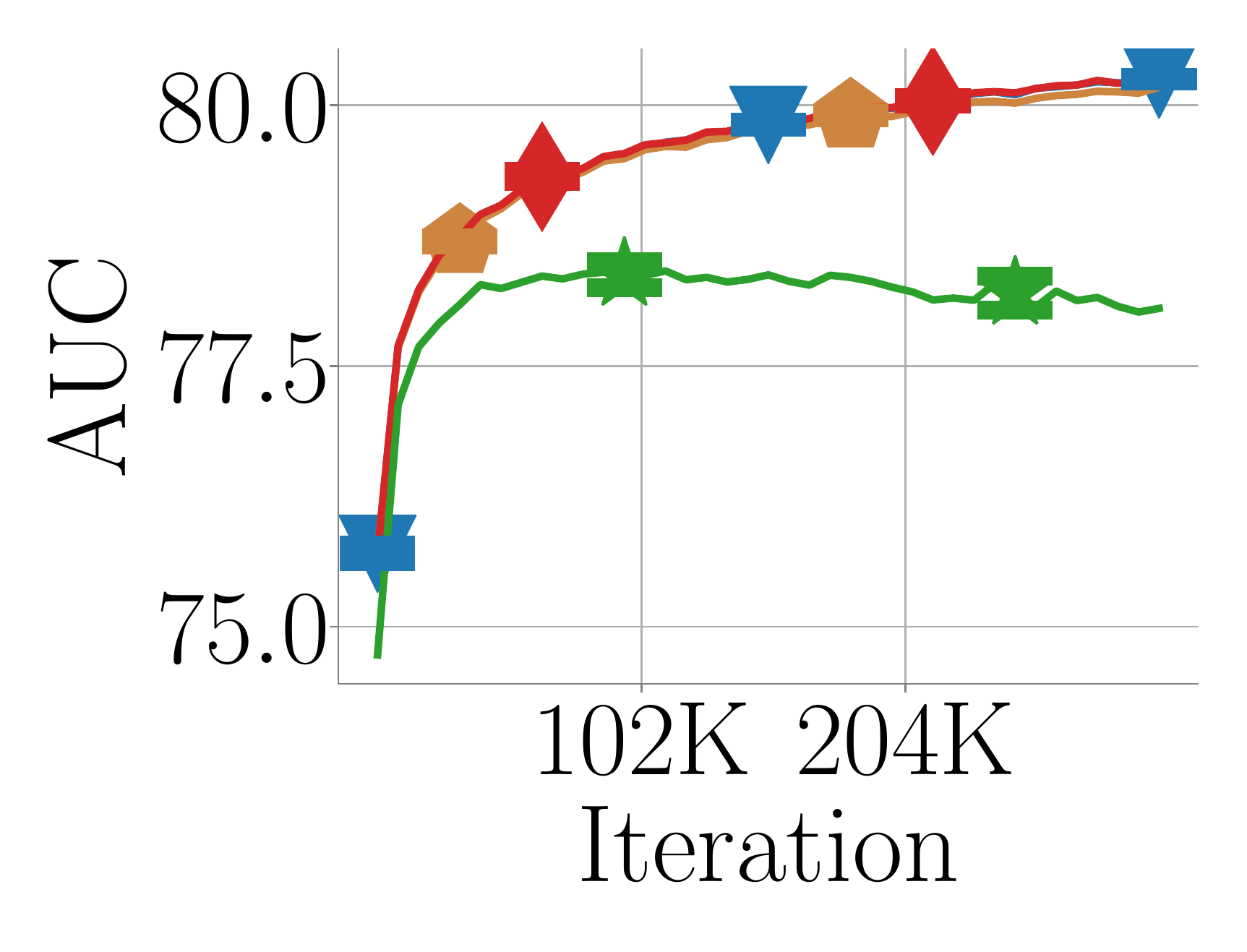}
    \caption{DLRM-Kaggle}
  \end{subfigure}
  \begin{subfigure}[b]{0.24\textwidth}
    \includegraphics[width=\textwidth]{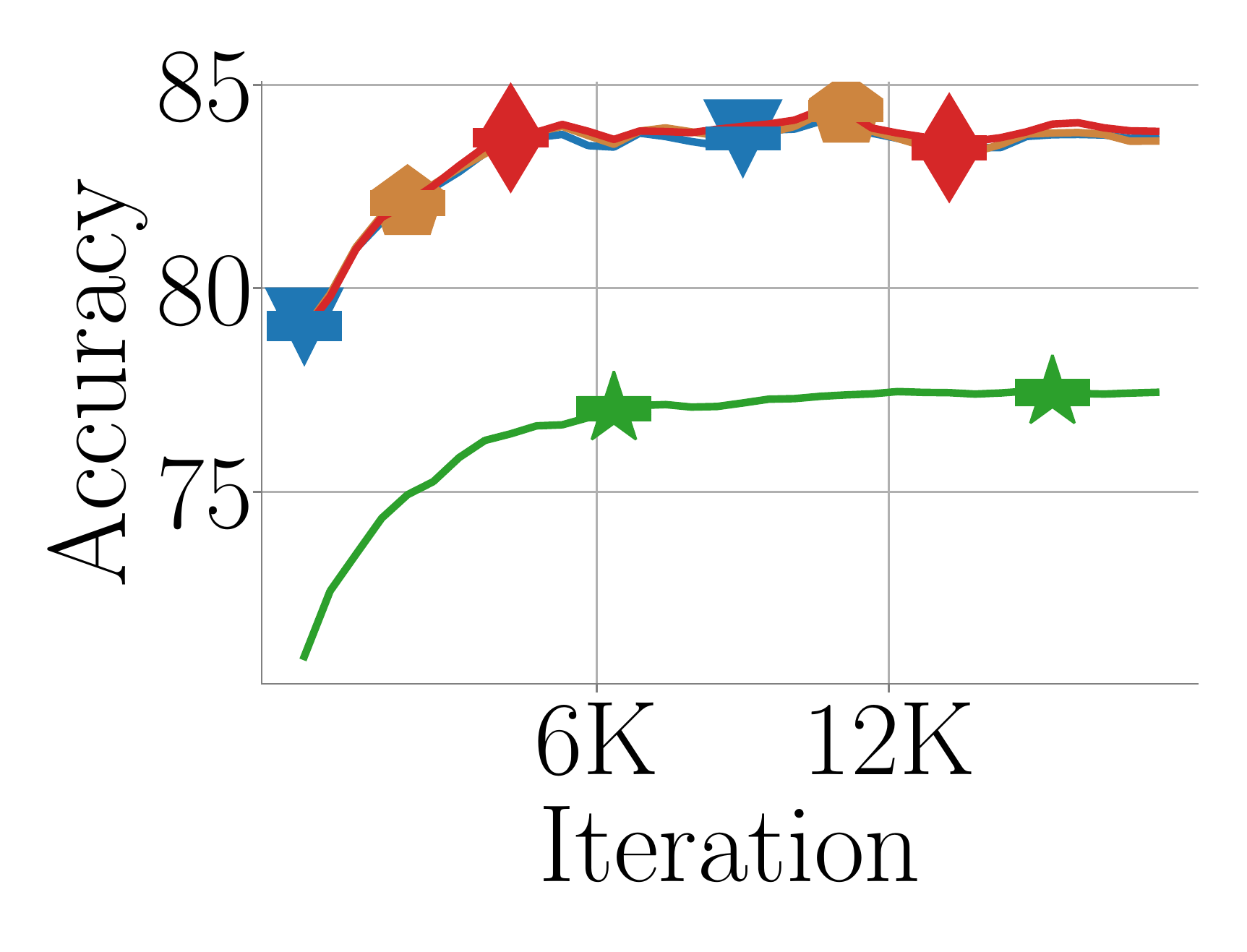}
    \caption{BERT-MNLI}
  \end{subfigure}\\
  \begin{subfigure}[b]{0.24\textwidth}
    \includegraphics[width=\textwidth]{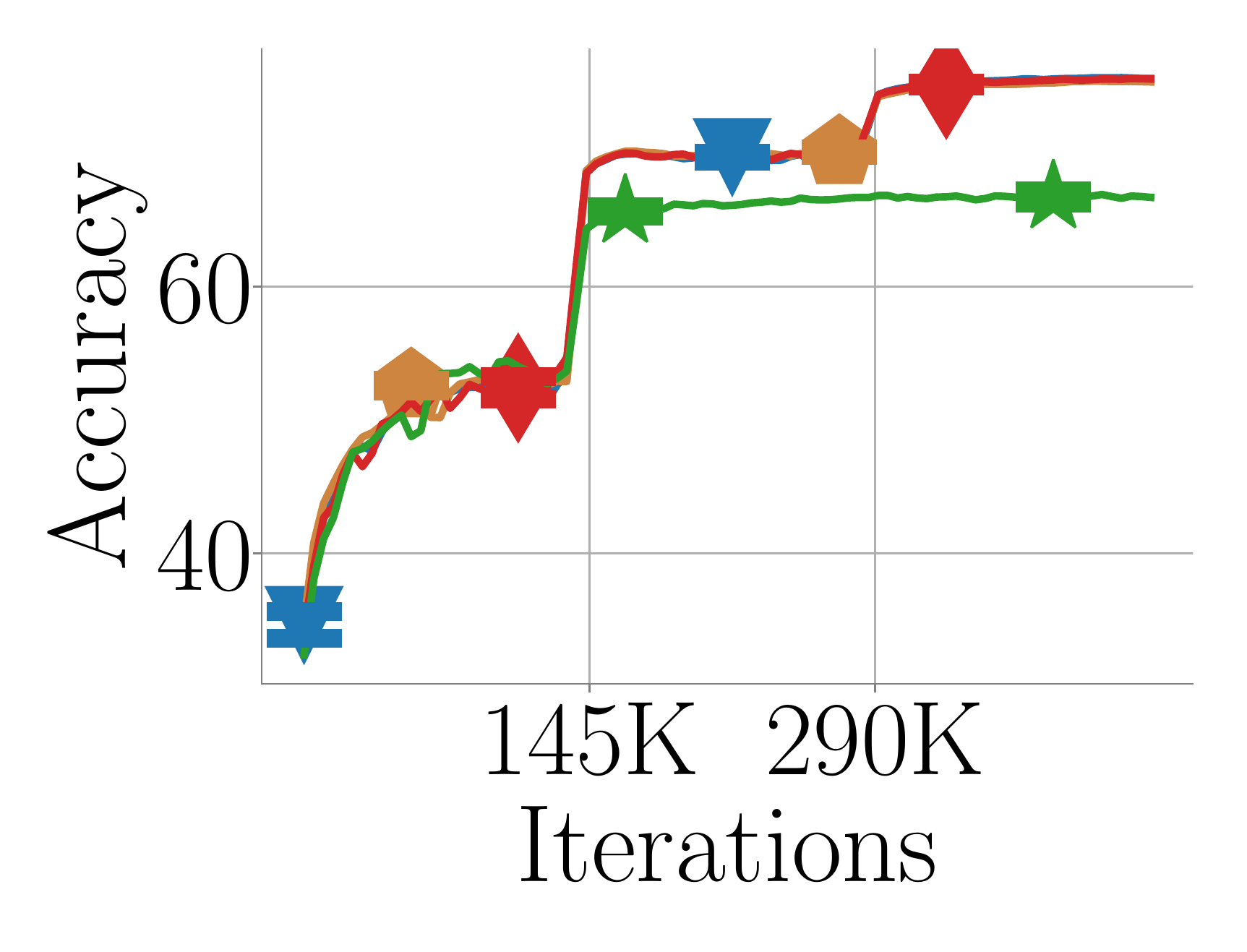}
    \caption{ResNet-50-ImageNet}
  \end{subfigure}
  \hfill
  \begin{subfigure}[b]{0.24\textwidth}
    \includegraphics[width=\textwidth]{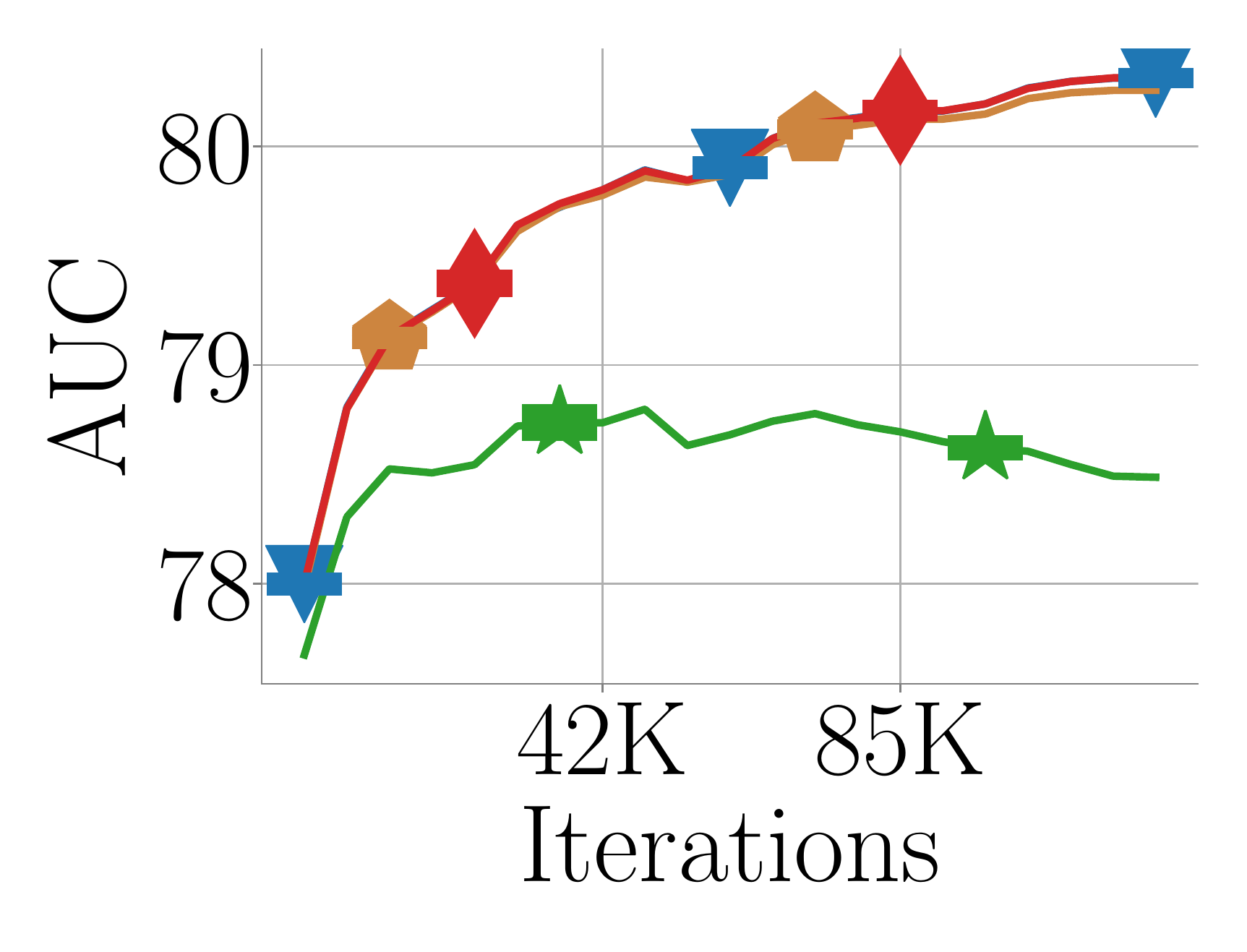}
    \caption{DLRM-Terabyte}
  \end{subfigure}
  \begin{subfigure}[b]{0.24\textwidth}
    \includegraphics[width=\textwidth]{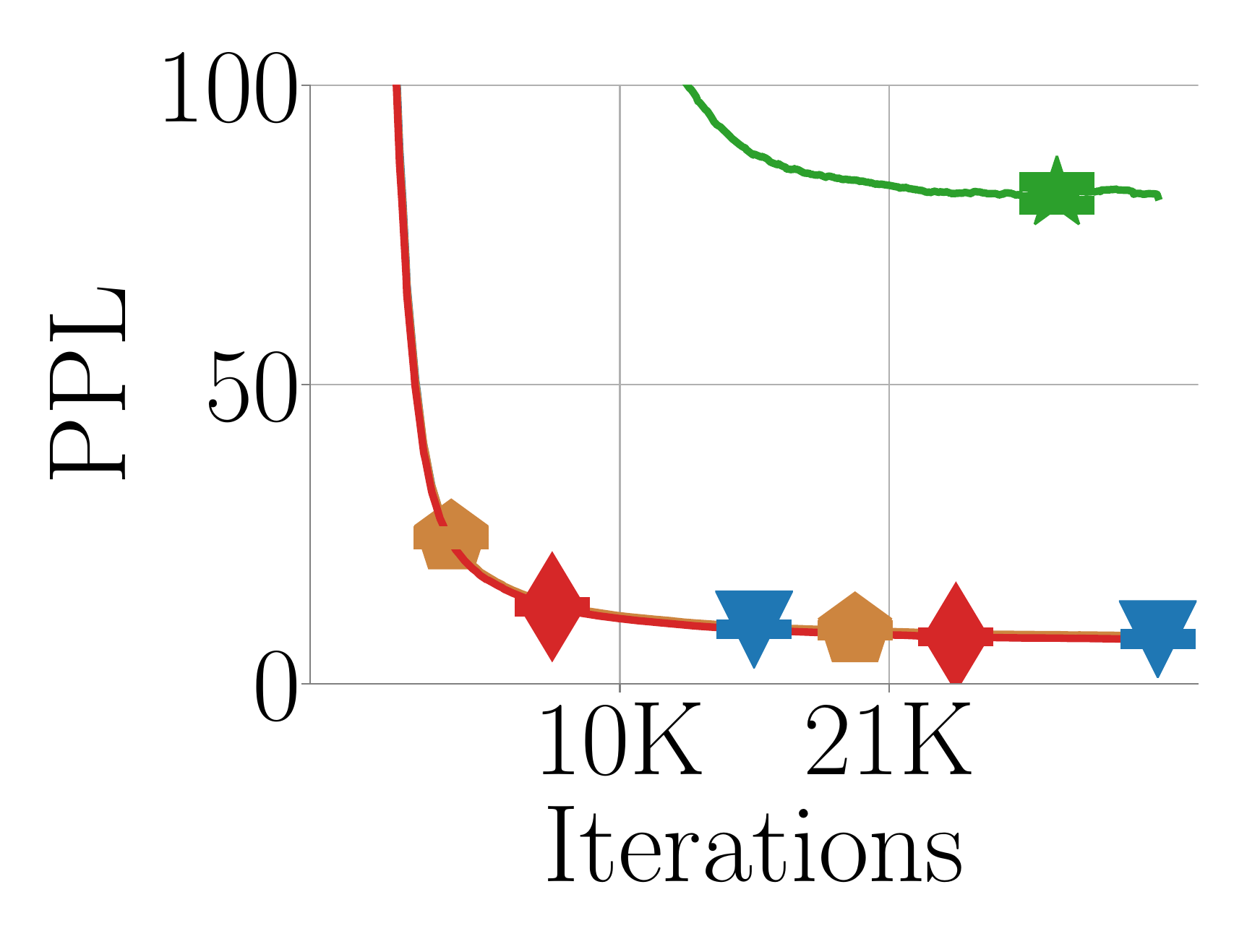}
    \caption{BERT-Wiki103}
  \end{subfigure}
  \begin{subfigure}[b]{0.24\textwidth}
    \includegraphics[width=\textwidth]{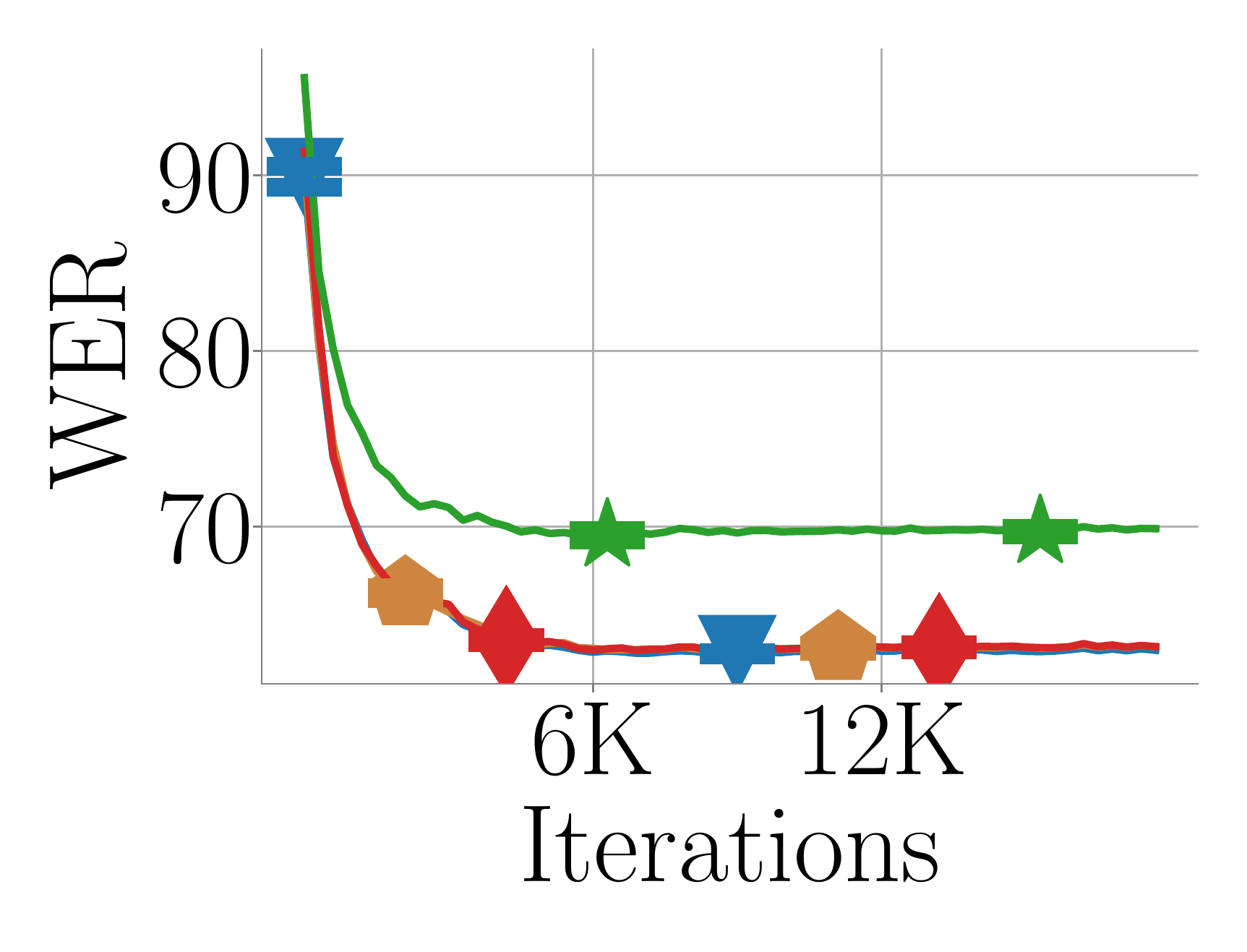}
    \caption{DS2-Librispeech}
    \end{subfigure}
  \caption{\textbf{Validation accuracy for 16-bit-FPU training.} With stochastic rounding or Kahan summation enabled for model weight updates, 16-bit-FPU training matches 32-bit precision training in terms of validation accuracy with negligible differences across the applications in our experiments.}
  \label{fig:app:exp}
\end{figure*}

In~\Cref{fig:app:exp}, we present the validation accuracy curves attained by 16-bit-FPU training with stochastic rounding or Kahan summation for model weight updates. First we observe that with nearest rounding for all compute operations, the standard 16-bit-FPU training algorithm demonstrates validation accuracy gap compared to 32-bit training across training stages.
In the contrast, we can observe that by using stochastic rounding or Kahan summation for model weight updates, 16-bit-FPU training achieves validation accuracy curves matching that of 32-bit training.

\subsection{Other Experiment Results}
\label{app:experiment:subsec:other}

\begin{figure*}
\centering
  \begin{subfigure}[b]{0.30\textwidth}
    \includegraphics[width=\textwidth]{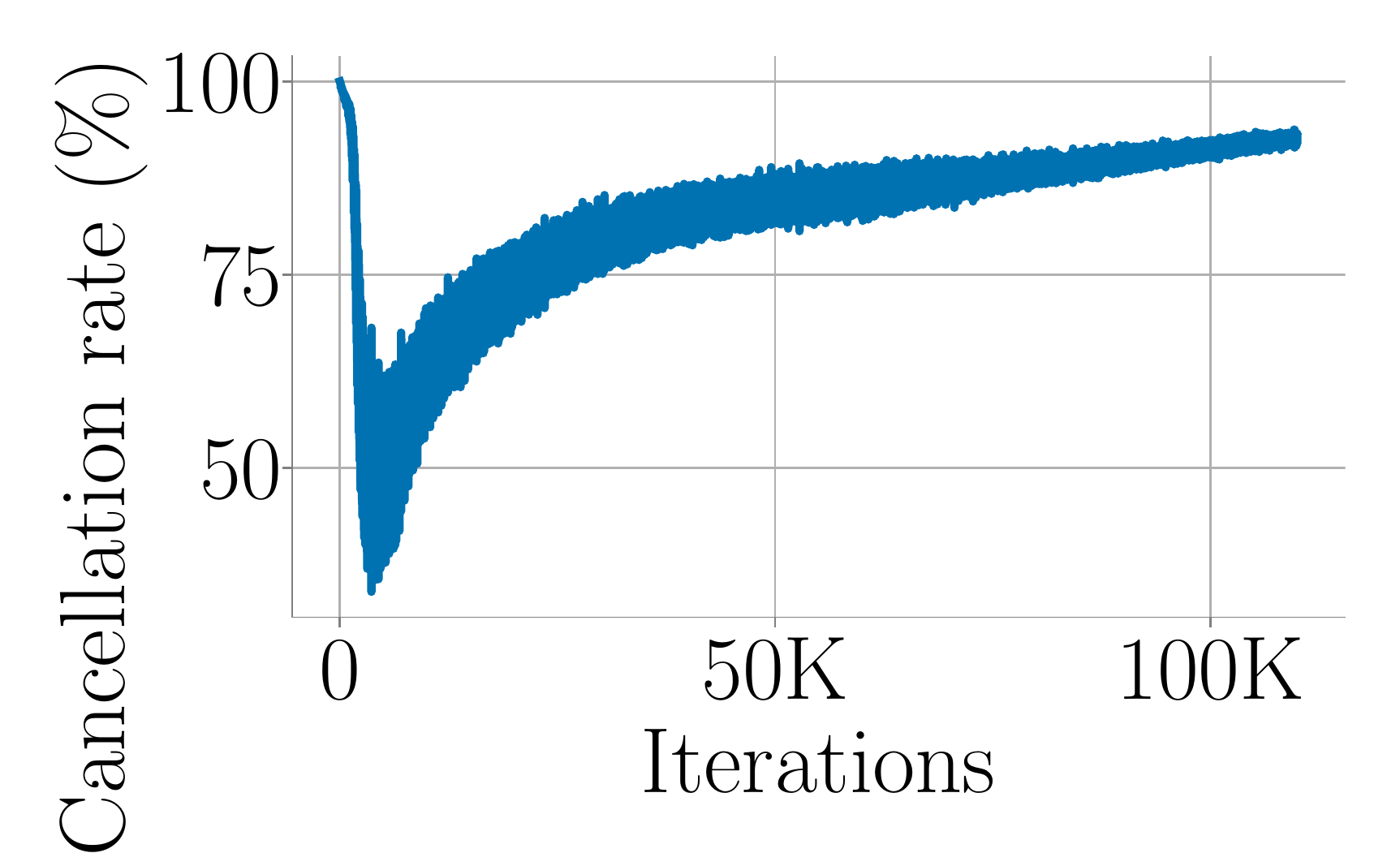}
    \caption{Terabyte-embedding}
    \label{fig:app:validation:embedding}
  \end{subfigure}
  \begin{subfigure}[b]{0.30\textwidth}
    \includegraphics[width=\textwidth]{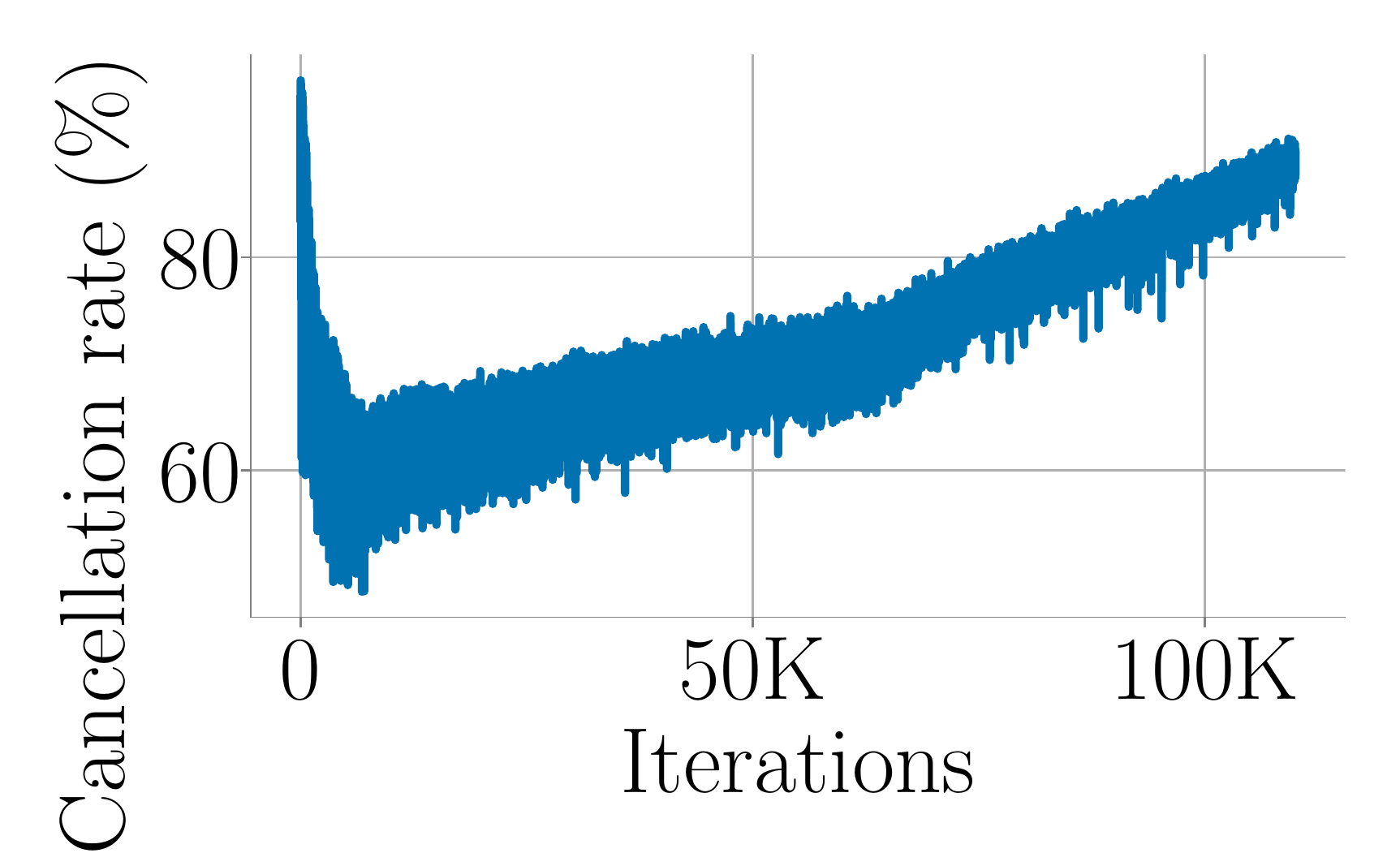}
    \caption{Terabyte-linear}
    \label{fig:app:validation:non-embedding}
  \end{subfigure}\\
    \begin{subfigure}[b]{0.30\textwidth}
    \includegraphics[width=\textwidth]{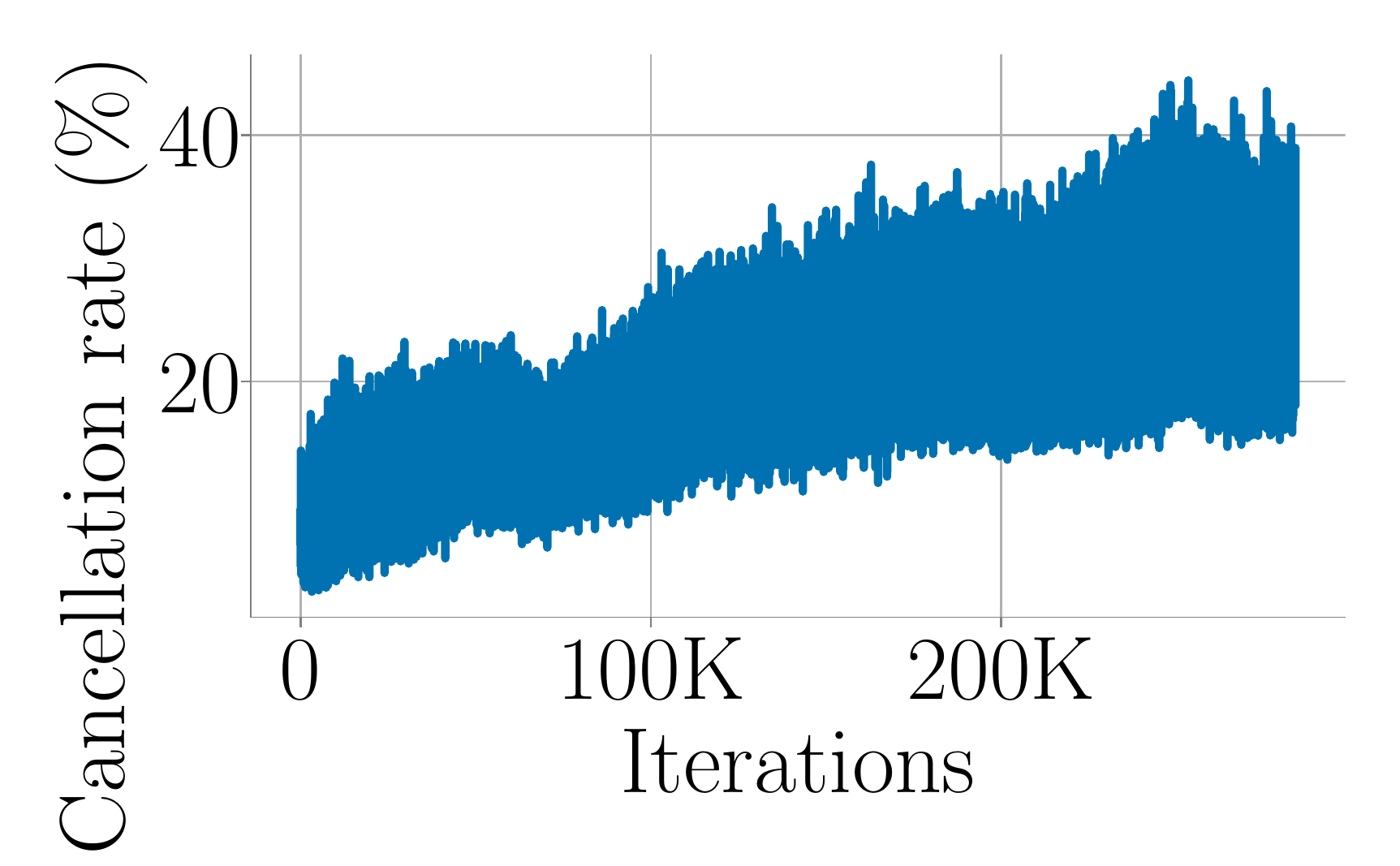}
    \caption{Kaggle-embedding}
    \label{fig:app:validation:non-embedding}
  \end{subfigure}
    \begin{subfigure}[b]{0.30\textwidth}
    \includegraphics[width=\textwidth]{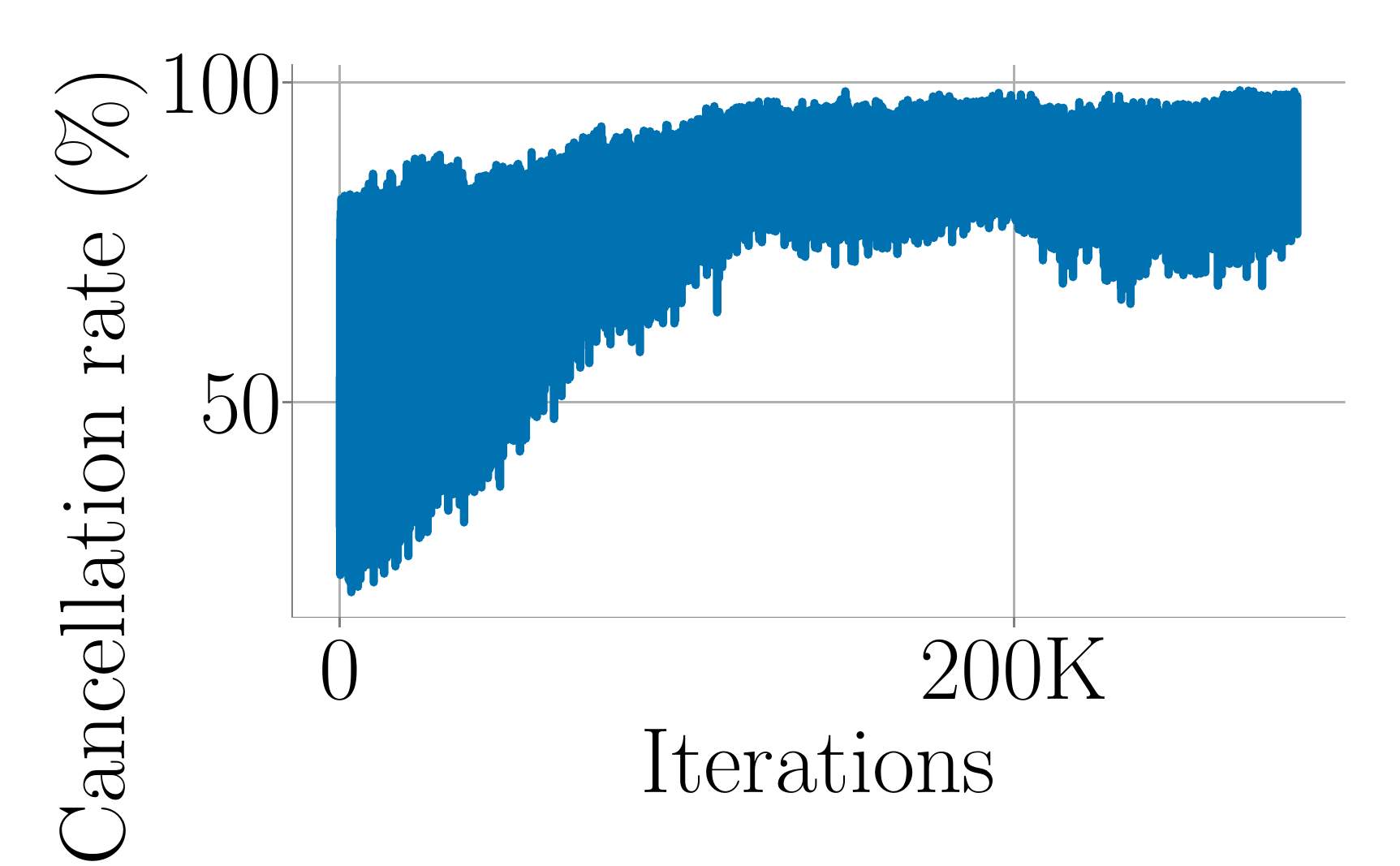}
    \caption{Kaggle-linear}
    \label{fig:app:validation:non-embedding}
  \end{subfigure}
  \caption{\textbf{The percentage of weight updates that are non-zero but get canceled in the standard 16-bit-FPU training.} 
  For both embedding and linear layers, the percentage increases in the mid-to-late training stage for the DLRM model on the Kaggle and Terabyte datasets. The high percentage of cancelation aligns with our insights on the impact of nearest rounding on model weight updates.
  }
  \label{fig:app:validation}
\end{figure*}

\paragraph{Nearest rounding cancels weight updates}
In this experiment, we show that a large portion of the model weight updates can be canceled due to nearest rounding on weight updates for a representative deep learning model; this aligns with our theoretical insights on the impact of weight updates cancellation from the least-squares regression model.
Specifically, we train DLRM model using the standard 16-bit training. In this experiment, we record the percentage of weight updates that are non-zero and get canceled by nearest rounding on model weight updates at each training iteration. Because the DLRM model consists of two different layer types which are embeddings and linear layers in MLPs, we present the results in~\Cref{fig:app:validation} on one embedding layer and one linear layer for each of the dataset we use for the DLRM model; the behavior on these layers is representative for many different embedding layers and linear layers in the DLRM model. 
As it is shown in \Cref{fig:app:validation}, on both the Kaggle and the Terabyte datasets, the percentage of weight updates that are non-zero but get canceled increases as the training progresses. For both the embedding and linear layers, the percentage of canceled weight updates can be larger than $80\%$ in the mid-to-late training stage. For the Kaggle dataset, the increasing percentage of cancellation is due to the shrinking gradient magnitude because we use a constant learning rate on this dataset. On the other hand, we use a decaying learning rate on the Terabyte dataset. Thus the increasing trend on the percentage of weight update cancelation is a consequence of the compound effect from decaying gradient magnitude and decaying learning rates. These observations align with our insights on the least-squares regression model where we observe severe model weight updates cancelation in the mid-to-late training stage.

\begin{figure}[H]
					\centering
					\includegraphics[width=0.35\textwidth]{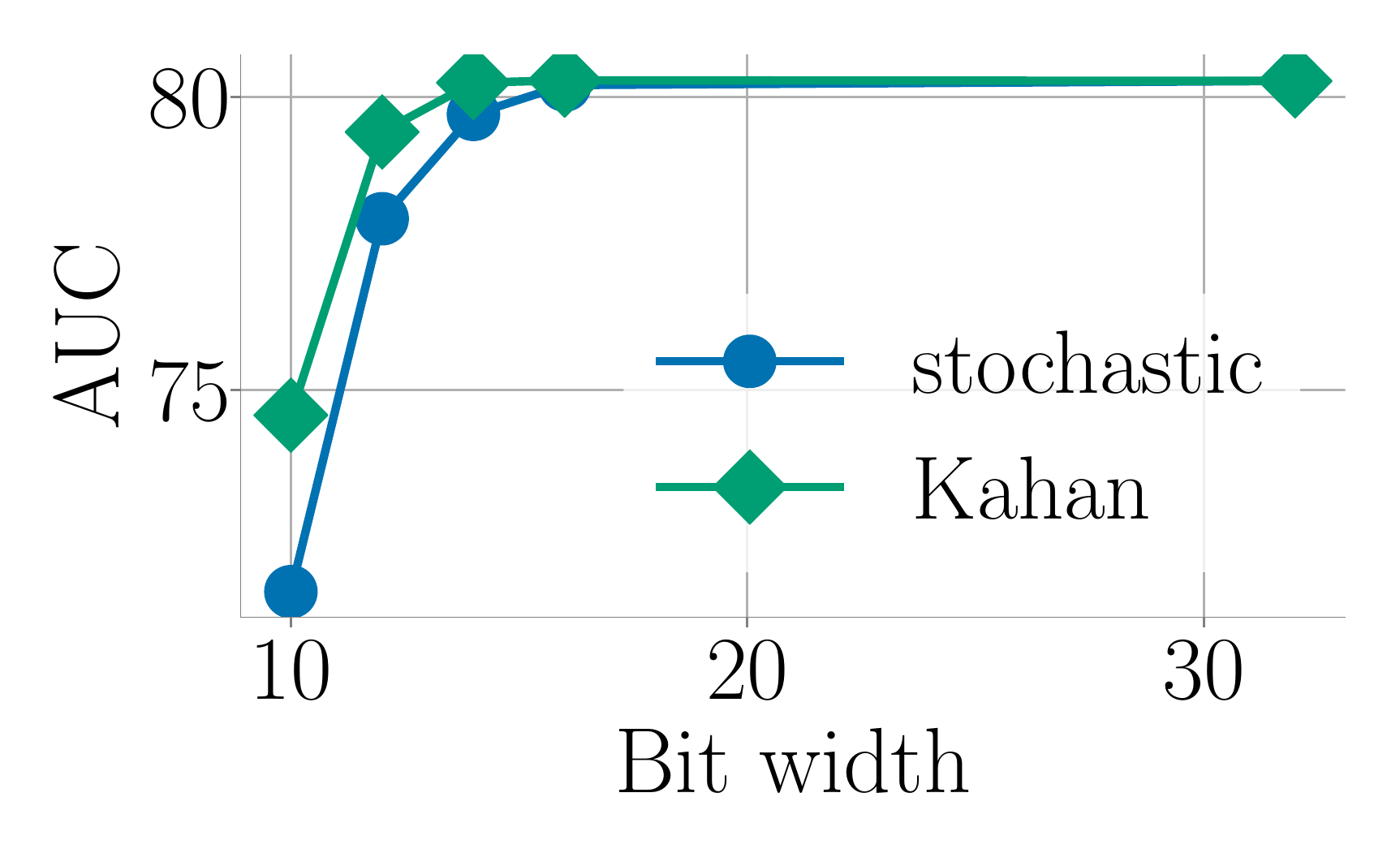}
					\caption{\textbf{Training accuracy with precisions lower than 16-bit.} Besides 14-bit training with Kahan summation for model weight updates, training with lower precision than 16-bit shows lower training accuracy compared to either 16-bit or 32-bit training on DLRM-Kaggle.}
					\label{fig:app:lowprecision}
				\end{figure}

\paragraph{Going below 16-bit}
We train DLRM model on Criteo Kaggle dataset with precisions lower than 16-bit. For the lower precision representations, we keep $8$ exponent bits as \BFHS and only lower the number of mantissa bits to ensure enough number dynamic range for stable training. In~\Cref{fig:app:lowprecision}, we demonstrate the model accuracy attained with 10, 12 and 14-bit precision. Besides using 14-bit precision with Kahan summation for model weight updates, the other configuration all demonstrate lower training AUC value compared to 32-bit training and 16-bit-FPU training with stochastic rounding or Kahan summation. This implies that to go lower than 16-bit precision for training generic deep learning models, additional techniques might be required.

\begin{figure*}
\begin{subfigure}[b]{\textwidth}
\includegraphics[width=\textwidth]{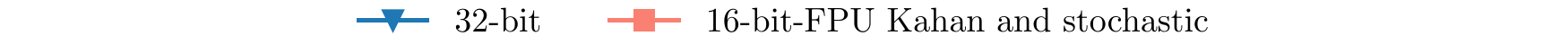}
\end{subfigure}
  \begin{subfigure}[b]{0.32\textwidth}
    \includegraphics[width=\textwidth]{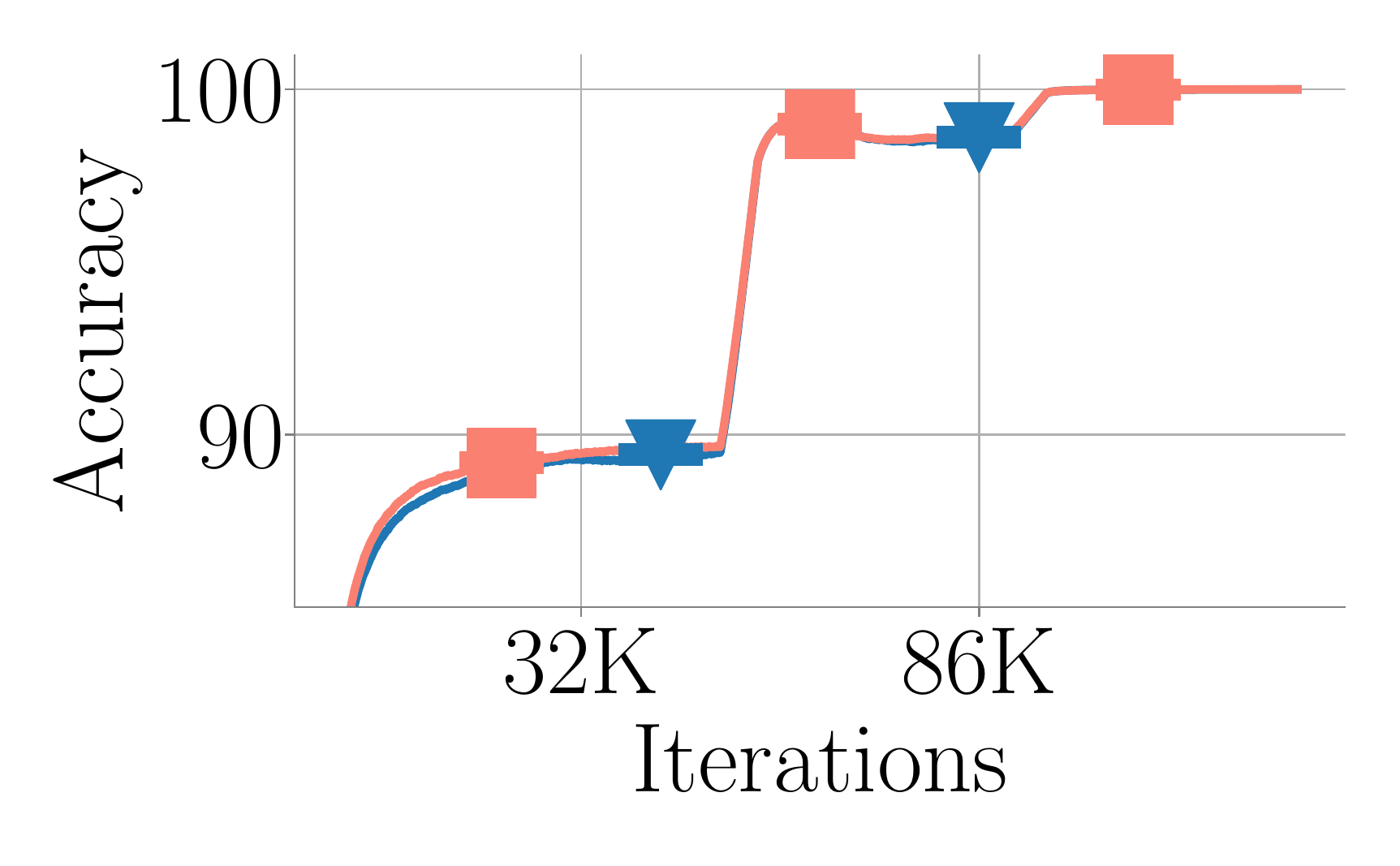}
    \caption{ResNet-18-CIFAR10}
  \end{subfigure}
  \hfill
  \begin{subfigure}[b]{0.32\textwidth}
    \includegraphics[width=\textwidth]{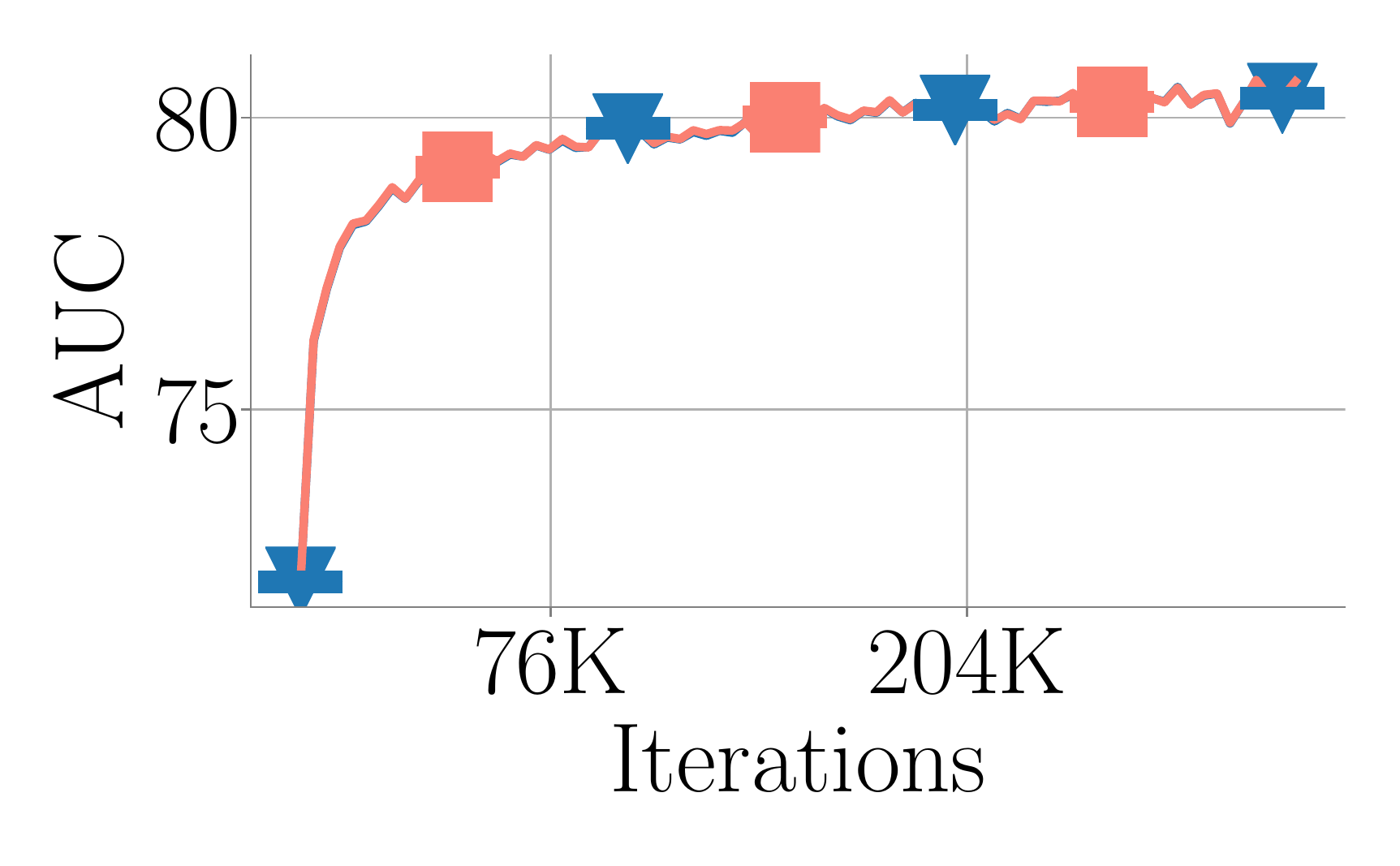}
    \caption{DLRM-Kaggle}
  \end{subfigure}
  \begin{subfigure}[b]{0.32\textwidth}
    \includegraphics[width=\textwidth]{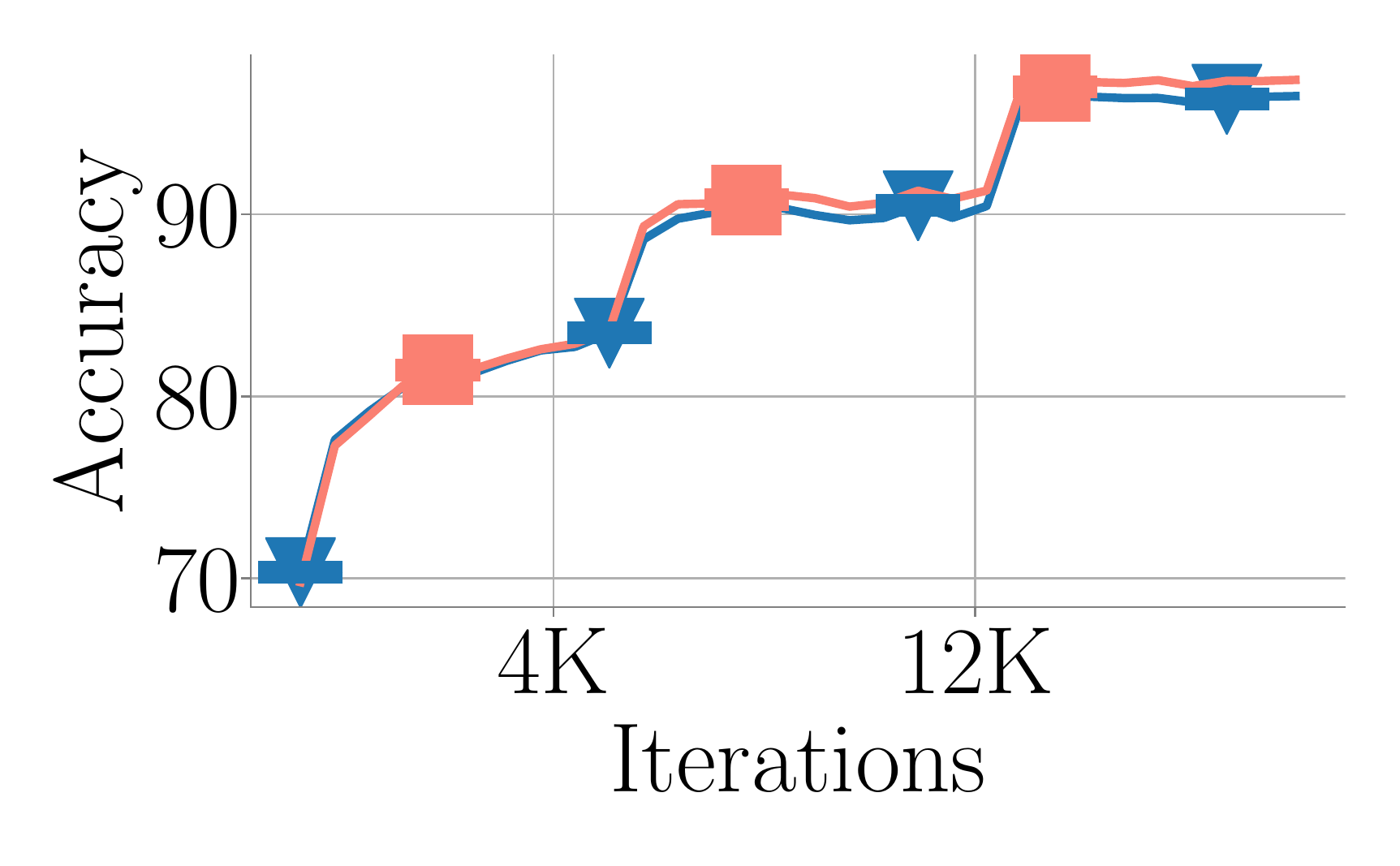}
    \caption{BERT-MNLI}
  \end{subfigure}
  \caption{\textbf{16-bit-FPU training using stochastic rounding and Kahan summation simultaneously.}
By using the Kahan summation and stochastic rounding simultaneously for model weight updates, 16-bit-FPU training can also robustly attain matching model accuracy compared to 32-bit precision training.
  }
  \label{fig:app:kahan_and_stochastic}
\end{figure*}

\paragraph{Combining two techniques}
To test the robustness of the two numerical techniques for 16-bit-FPU training, we discuss the model accuracy attained by applying stochastic rounding and Kahan summation simultaneously on all the model weights. In~\Cref{fig:app:kahan_and_stochastic}, we can observe that stochastic rounding and Kahan summation can work together robustly and attain validation accuracy matching that of 32-bit training. 

\begin{figure*}
\begin{subfigure}[b]{\textwidth}
\includegraphics[width=\textwidth]{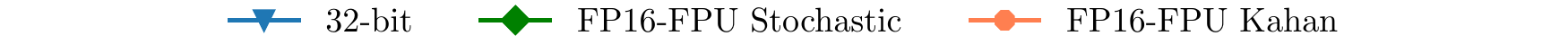}
\end{subfigure}
  \begin{subfigure}[b]{0.32\textwidth}
    \includegraphics[width=\textwidth]{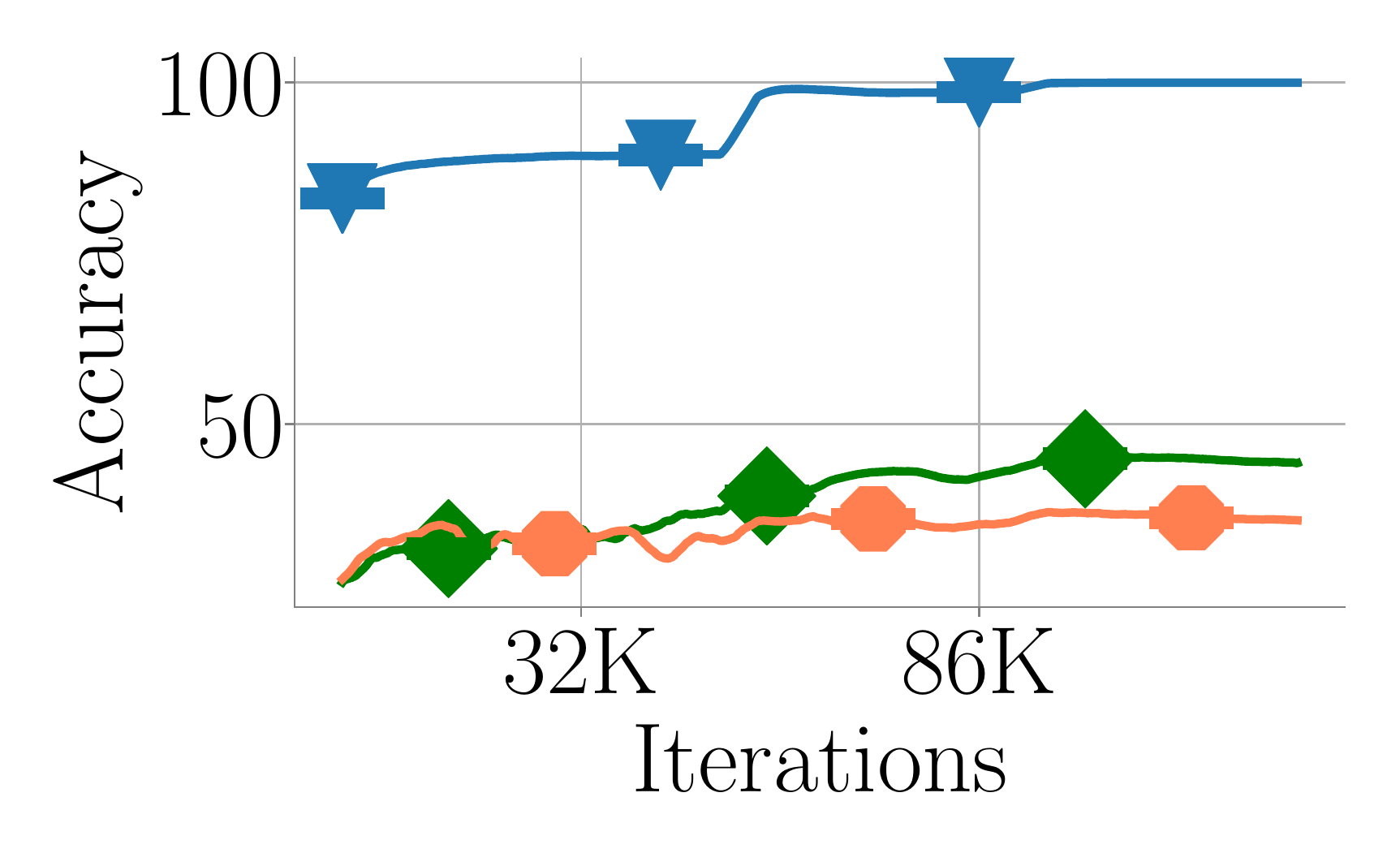}
    \caption{ResNet-18-CIFAR10}
  \end{subfigure}
  \hfill
  \begin{subfigure}[b]{0.32\textwidth}
    \includegraphics[width=\textwidth]{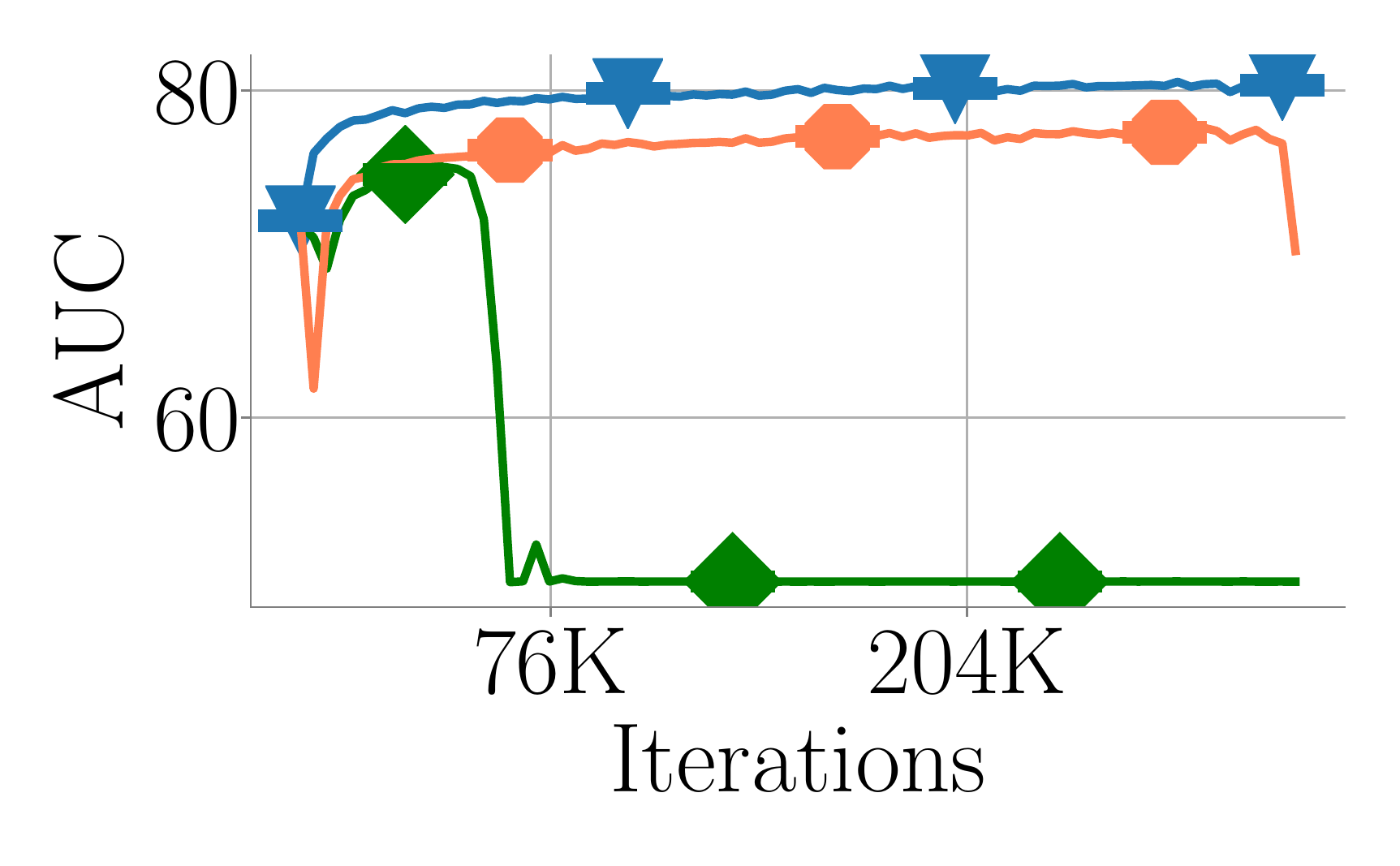}
    \caption{DLRM-Kaggle}
  \end{subfigure}
  \begin{subfigure}[b]{0.32\textwidth}
    \includegraphics[width=\textwidth]{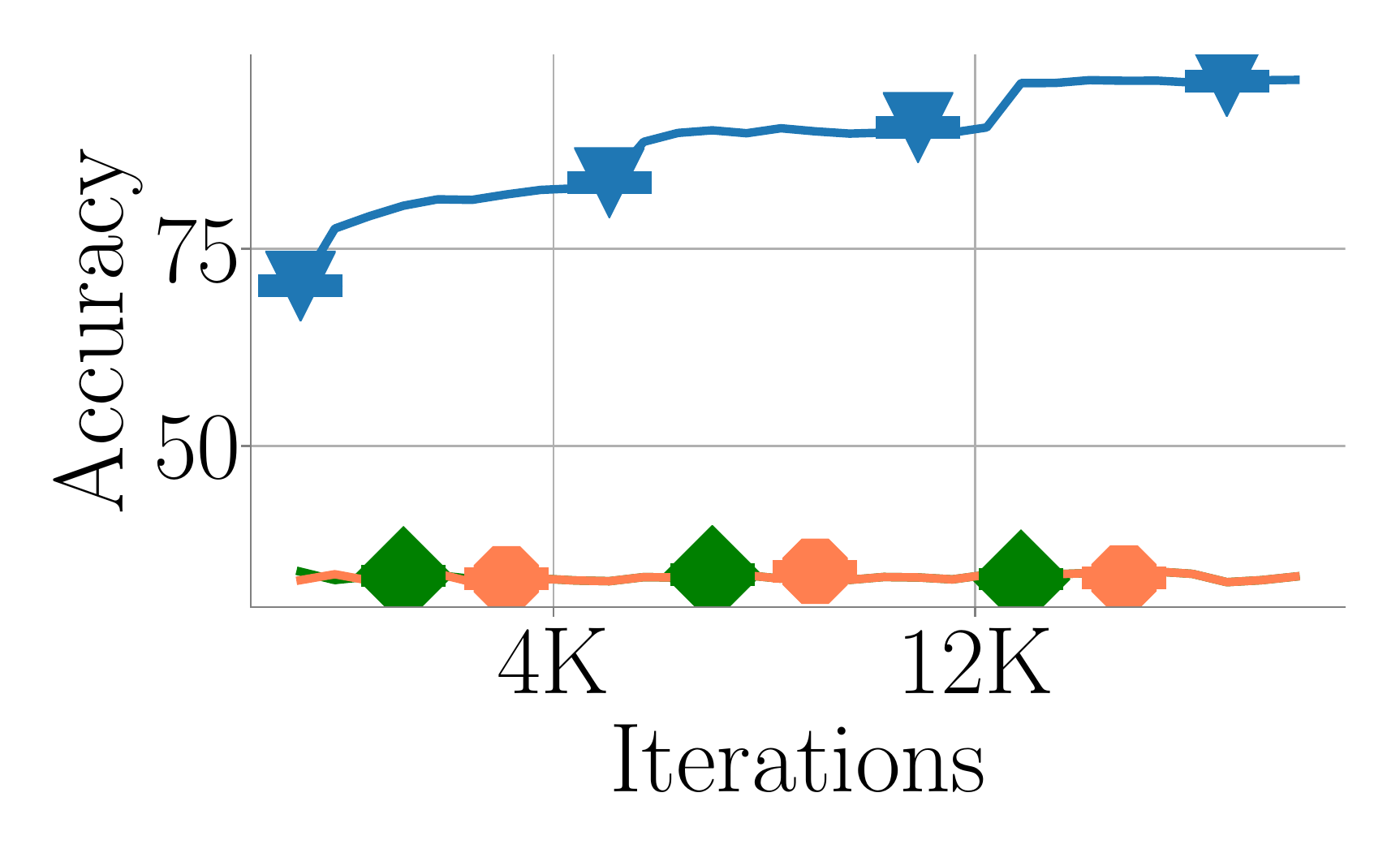}
    \caption{BERT-MNLI}
  \end{subfigure}
  \caption{\textbf{Training accuracy gap imposed by 16-bit-FPU training with \FPHS.} Despite the fact that we enabled stochastic rounding or Kahan summation for the model weight updates, 16-bit-FPU training with \FPHS still demonstrate a significant model accuracy gap compared to 32-bit training. This is mostly due to the small dynamic range of the \FPHS precision.}
  \label{fig:app:fp16}
\end{figure*}

\paragraph{Using \FPHS instead of \BFHS}
We also consider doing 16-bit-FPU training with \FPHS precision. We apply stochastic rounding and Kahan summation for model weights in 16-bit-FPU training with \FPHS precision. We consider the minimal algorithms without any loss scaling techniques as in mixed precision training.
As shown in~\Cref{fig:app:fp16}, we can observe that 16-bit-FPU training with \FPHS precision (even with the stochastic rounding or Kahan summation technique) demonstrates significant model accuracy degradation compared to 32-bit precision training.
Because \FPHS has more mantissa bits but fewer exponent bits than \BFH, the model accuracy gap is mostly due to the small dynamic range of \FPHS.

